\documentclass{article}
\usepackage{iclr2023_conference,times}

\usepackage{microtype}

\usepackage{amsmath}
\usepackage{amssymb}
\usepackage{mathtools}
\usepackage{amsthm}
\usepackage{graphicx}
\usepackage{booktabs}
\usepackage{colortbl}
\usepackage{amsmath}
\usepackage{amssymb}
\usepackage{amsfonts}
\usepackage{multirow}
\usepackage{verbatim}
\usepackage{caption}
\usepackage{longtable}
\usepackage{supertabular}
\usepackage{float}

\usepackage{enumitem}
\usepackage{xcolor}
\usepackage{xspace}
\usepackage{textcomp}
\usepackage{makecell}
\usepackage{multirow}
\usepackage{multicol}
\usepackage{lscape} 
\usepackage{siunitx}
\usepackage{algorithm}
\usepackage{algpseudocode}

\usepackage{amssymb}
\usepackage{pifont}
\newcommand{\cmark}{\ding{51}}%
\newcommand{\xmark}{\ding{55}}%
\usepackage{scrextend}

\usepackage{array}
\usepackage{latexsym}
\usepackage[T1]{fontenc}
\usepackage[utf8]{inputenc}
\usepackage{microtype}
\usepackage{url}            
\usepackage{nicefrac}       
\usepackage{changepage}
\usepackage{xargs}
\usepackage{wrapfig,lipsum,booktabs}
\usepackage{longtable}
\usepackage{subcaption}
\usepackage{pgfplots}
\usetikzlibrary{pgfplots.groupplots}
\pgfplotsset{compat=1.3}
\usepackage{tikz}
\usetikzlibrary{patterns}

\usepackage[most]{tcolorbox}

\usepackage{hyperref}
\usepackage[capitalize,noabbrev]{cleveref}
\crefname{section}{Section}{\S\S}
\crefname{section}{Section}{\S\S}
\crefname{table}{Table}{Tables}
\crefname{figure}{Figure}{Figures}
\crefname{algorithm}{Algorithm}{}
\crefname{equation}{eq.}{}
\crefname{appendix}{Appendix}{}
\crefformat{section}{Section #2#1#3}
\crefformat{footnote}{#2\footnotemark[#1]#3}

\definecolor{mydarkblue}{rgb}{0,0.08,0.45}
\hypersetup{
    colorlinks=true,
    linkcolor=mydarkblue,
    filecolor=blue,      
    urlcolor=mydarkblue,
    citecolor=mydarkblue,
}

\definecolor{battleshipgrey}{rgb}{0.3, 0.3, 0.3}
\definecolor{americanrose}{rgb}{1.0, 0.01, 0.24}
\definecolor{jweigreen}{rgb}{0,0.45,0.24}
\definecolor{bluegray}{rgb}{0.1, 0.1, 0.4}
\definecolor{ao(english)}{rgb}{0.0, 0.5, 0.0}
\definecolor{blanchedalmond}{rgb}{1.0, 0.92, 0.8}
\definecolor{atomictangerine}{rgb}{1.0, 0.6, 0.4}
\definecolor{chocolate(web)}{rgb}{0.82, 0.41, 0.12}
\definecolor{bananayellow}{rgb}{1.0, 0.88, 0.21}
\definecolor{goldenbrown}{rgb}{0.6, 0.4, 0.08}
\definecolor{aliceblue}{rgb}{0.94, 0.97, 1.0}
\definecolor{beige}{rgb}{0.96, 0.96, 0.86}
\definecolor{babyblue}{rgb}{0.54, 0.81, 0.94}
\definecolor{camel}{rgb}{0.76, 0.6, 0.42}
\definecolor{cinnamon}{rgb}{0.82, 0.41, 0.12}
\definecolor{deepskyblue}{rgb}{0.0, 0.75, 1.0}
\definecolor{frenchblue}{rgb}{0.0, 0.45, 0.73}
\definecolor{classicrose}{rgb}{0.98, 0.8, 0.91}
\definecolor{frenchrose}{rgb}{0.96, 0.29, 0.54}
\definecolor{frenchlilac}{rgb}{0.53, 0.38, 0.56}
\definecolor{frenchbeige}{rgb}{0.65, 0.48, 0.36}
\definecolor{darkpastelgreen}{rgb}{0.01, 0.75, 0.24}
\definecolor{fluorescentorange}{rgb}{1.0, 0.75, 0.0}

\definecolor{instructcolorone}{rgb}{0.53, 0.38, 0.56} 
\definecolor{instructcolortwo}{rgb}{1.0, 0.72, 0.77} 
\definecolor{instructcolorthree}{rgb}{0.0, 0.45, 0.73} 
\definecolor{instructcolorfour}{rgb}{0.6, 0.6, 0.6} 
\definecolor{codexcolorone}{rgb}{0.0, 0.75, 1.0}
\definecolor{codexcolortwo}{rgb}{0.0, 0.45, 0.73}
\definecolor{codexcolorthree}{rgb}{0.0, 0.06, 0.54}
\definecolor{palmcolorone}{rgb}{0.77, 0.76, 0.82}
\definecolor{flancolorone}{rgb}{0.0, 0.75, 1.0}
\definecolor{symflancolorone}{rgb}{0.0, 0.45, 0.73}
\definecolor{comparisoncolor}{rgb}{0.44, 0.16, 0.39}
\newcommand{\binaryrandomcolor}[0]{black!50}
\newcommand{\codexshape}[0]{*} 
\newcommand{\basegptshape}[0]{pentagon*} 
\newcommand{\instructgptshape}[0]{triangle*} 
\newcommand{\palmshape}[0]{diamond*} 

\newcommand{\papertitle}[0]{\vspace{-2mm}Symbol tuning improves in-context learning in language models}

\iclrfinalcopy

\usepackage[capitalize,noabbrev]{cleveref}

\theoremstyle{plain}

\theoremstyle{definition}

\theoremstyle{remark}

\usepackage[loose]{minitoc}

\title{\raggedright \papertitle}

\author{
\vspace{4mm}
\hspace{-4.5mm}
    Jerry Wei$^{1, 2, }$\thanks{Work done as a Student Researcher at Google.} \hspace{5mm}
    Le Hou$^{1}$ \hspace{5mm}
    Andrew Lampinen$^{1}$ \hspace{5mm}
    Xiangning Chen$^{1, *}$ \hspace{5mm}
    Da Huang$^{1}$ \hspace{5mm}
    \\
    \textbf{\hspace{-3.2mm}
    Yi Tay$^{1}$ \hspace{3mm}
    Xinyun Chen$^{1}$ \hspace{3mm}
    Yifeng Lu$^{1}$ \hspace{3mm}
    Denny Zhou$^{1}$ \hspace{3mm}
    Tengyu Ma$^{1, 2, }$\thanks{Work done as a Visiting Researcher at Google.} \hspace{3mm}
    Quoc V. Le$^{1}$
    }
    \vspace{4mm}
    \\
    \hspace{-3mm}
    $^1$ Google
    \hspace{5mm}
    $^2$ Stanford University
}

\fancypagestyle{firstpage}{
  \lhead{\begin{picture}(0,0)\put(0,-3){\includegraphics[width=0.1\linewidth]{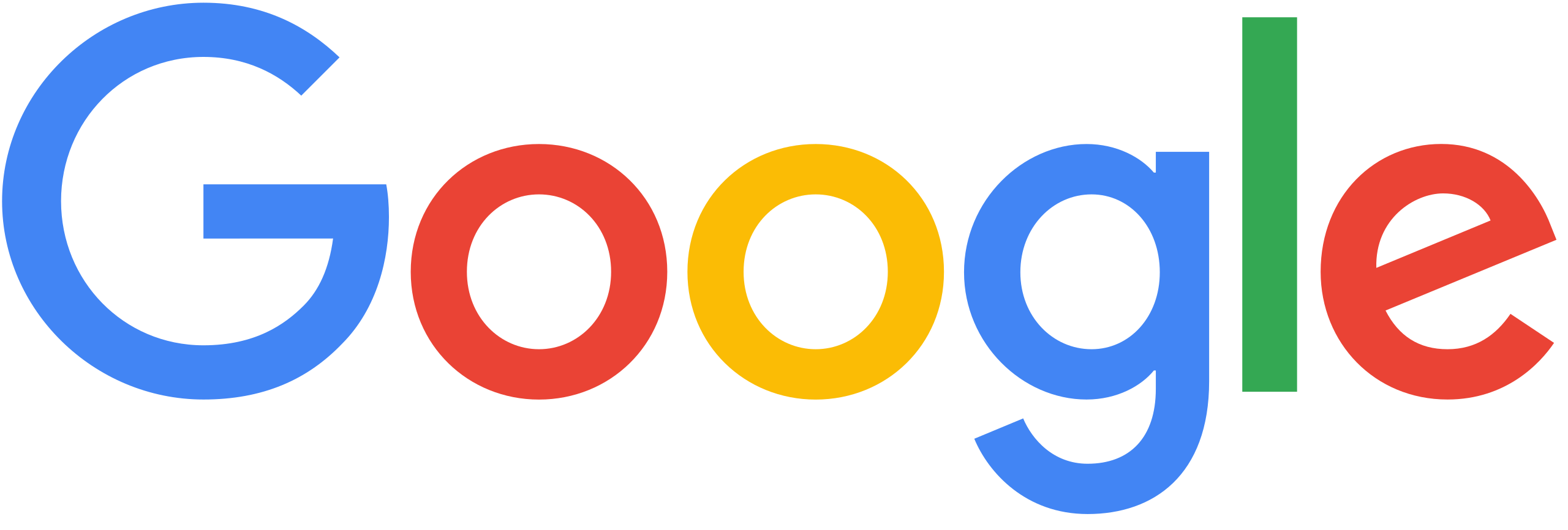}}\end{picture}}
  \rhead{\normalsize{\today}}
}

\begin{document}

\doparttoc 
\faketableofcontents 

\maketitle
\thispagestyle{firstpage}

\vspace{-5mm}
\begin{abstract}
We present \textit{symbol tuning}---finetuning language models on in-context input--label pairs where natural language labels (e.g., ``positive/negative sentiment'') are replaced with arbitrary symbols (e.g., ``foo/bar'').
Symbol tuning leverages the intuition that when a model cannot use instructions or natural language labels to figure out a task, it must instead do so by learning the input--label mappings.

We experiment with symbol tuning across Flan-PaLM models up to 540B parameters and observe benefits across various settings.
First, symbol tuning boosts performance on unseen in-context learning tasks and is much more robust to underspecified prompts, such as those without instructions or without natural language labels.
Second, symbol-tuned models are much stronger at algorithmic reasoning tasks, with up to 18.2\% better performance on the List Functions benchmark and up to 15.3\% better performance on the Simple Turing Concepts benchmark.
Finally, symbol-tuned models show large improvements in following flipped-labels presented in-context, meaning that they are more capable of using in-context information to override prior semantic knowledge.
\end{abstract}
\vspace{5mm}

\begin{figure}[bh]
    \vspace{-5mm}
    \centering
    \includegraphics[width=0.85\linewidth]{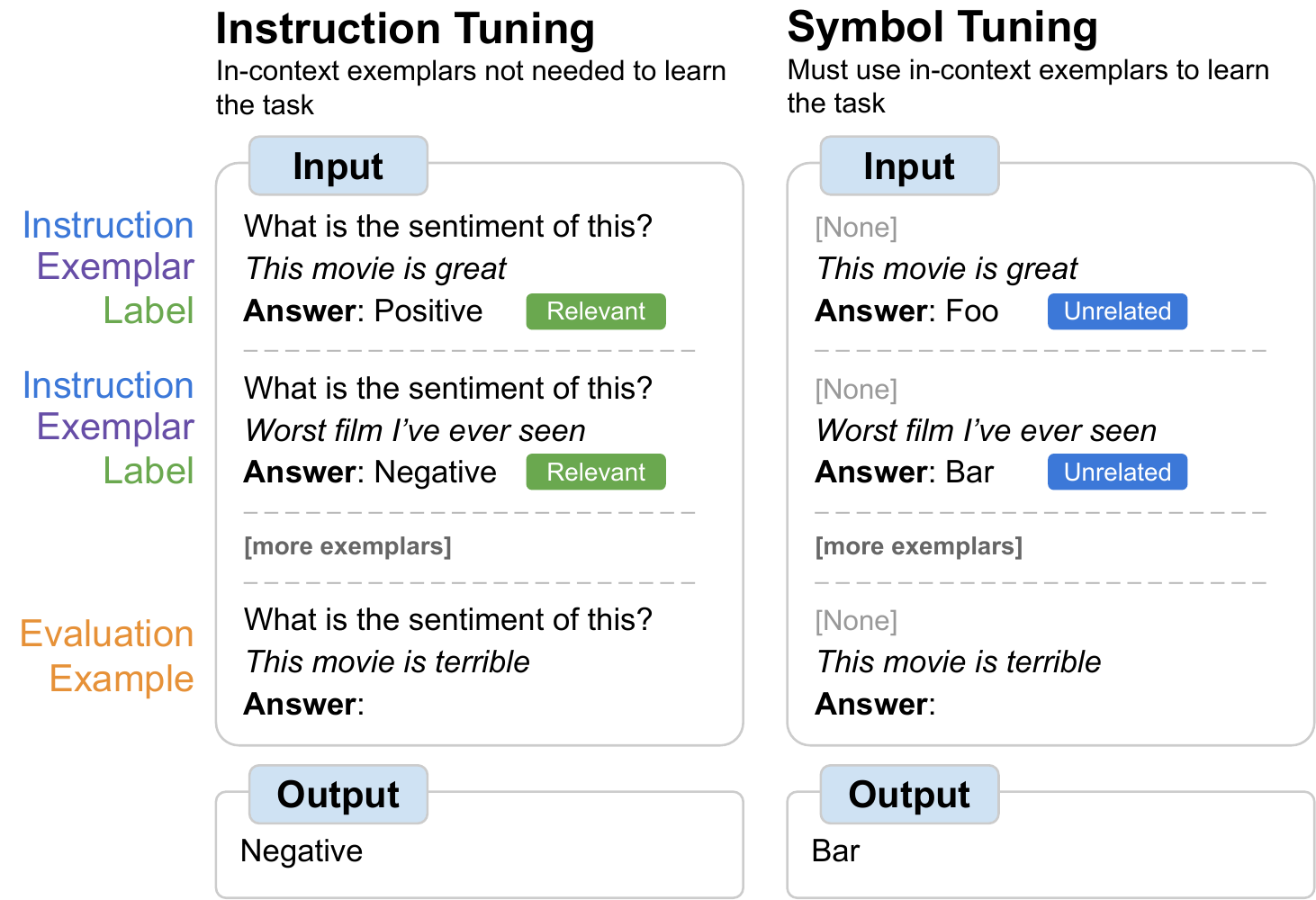}
    \caption{
    We tune models on tasks where natural language labels are replaced with arbitrary symbols (\textit{symbol tuning}).
    Symbol tuning relies on the intuition that when instruction and relevant labels are not available, models must use in-context exemplars to learn the task.
    }
    \label{fig:pull-figure}
\end{figure}

\clearpage
\section{Introduction}
\label{sec:introduction}
A key feature of human intelligence is that humans can learn to perform new tasks by reasoning using only a few examples. 
Scaling up language models has unlocked a range of new applications and paradigms in machine learning, including the ability to perform challenging reasoning tasks via few-shot examples given in-context \citep[][\textit{inter alia}]{brown2020language,chowdhery2022palm,openai2023gpt4}.
Language models, however, are still sensitive to the way that prompts are given, indicating that they are not reasoning in a robust manner.
For instance, language models often require heavy prompt engineering \citep{brown2020language,reynolds2021prompt} or phrasing tasks as instructions \citep[][\textit{inter alia}]{wei2021finetuned,ouyang2022training,sanh2022multitask}, and they exhibit unexpected behaviors such as performance on tasks being unaffected even when shown in-context exemplars with random labels \citep{min2022rethinking} or flipped labels \citep{wei2023larger}.

In this paper, we propose a simple finetuning procedure that we call \textit{symbol tuning}, which significantly improves the ability of language models to reason with and learn from input--label mappings presented in-context.
In the symbol-tuning procedure, we finetune language models on input--label pairs presented in-context where natural language labels are remapped to arbitrary symbols.\footnote{We call our method \textit{symbol} tuning because arbitrary designation is a key property of symbols \citep{Newell1976CompleterSA}, and manipulating symbols is a crucial part of intelligence \citep{Newell1980Physical,Santoro2021Symbolic}.}
The intuition is that when models cannot rely on instructions or relevant natural language labels to figure out a given task, it must instead do so by reasoning with input--label mappings in-context in order to learn the mappings that reveal the task.
We perform symbol tuning using a mixture of 22 NLP datasets with various arbitrary symbols as labels and experiment using several Flan-PaLM models \citep[][8B, 62B, 62B-cont, 540B]{chung2022scaling}.

First, symbol tuning improves performance of baseline models on unseen in-context learning tasks across various settings (with/without instructions, with/without relevant labels), with larger performance gains when instructions or natural language labels are not given in the prompt.
For example, when prompts do not contain instructions or relevant labels, symbol tuning yields a +11.1\% average performance improvement across eleven evaluation tasks for Flan-cont-PaLM-62B.

Second, symbol-tuned models are better at algorithmic reasoning tasks, a striking result since symbol tuning only includes natural language data and did not have any numerical or algorithmic data.
On a set of reasoning evaluation suites for list functions (e.g., remove the last element in a list), symbol-tuned models experience performance improvements of \textbf{+18.2\%} for Flan-PaLM-8B, \textbf{+11.1\%} for Flan-PaLM-62B, and \textbf{+3.6\%} for Flan-PaLM-540B.
On a set of turing concept tasks (e.g., swapping 0s and 1s in a string), symbol-tuned models also improve by \textbf{+15.3\%} for Flan-PaLM-8B and Flan-PaLM-62B and \textbf{+4.7\%} for Flan-PaLM-540B.

Additionally, we experiment on an in-context learning setting where inputs have flipped labels, which forces the model to override its prior knowledge when presented with contradictory information in-context.
Pretrained language models have the ability to somewhat follow flipped labels---this ability is lost during instruction tuning but can be restored via symbol tuning.

Finally, we conduct ablation studies demonstrating that symbol tuning is simple to implement and only requires a relatively-small amount of compute.
Symbol tuning does not require mixing instruction-tuning data or collecting a large number of datasets, and only 1k to 2k steps of tuning are needed to get its benefits.
Overall, we hope that the strong empirical results from symbol tuning encourage further work in allowing language models to reason over arbitrary symbols given in-context.

\section{Symbol tuning}
\label{sec:introduction-to-symbol-tuning}
Despite their ability to perform some reasoning tasks after being shown in-context exemplars \citep{chowdhery2022palm, openai2023gpt4}, language models are still sensitive to the way in which these tasks are presented in prompts \citep{brown2020language,reynolds2021prompt,wei2021finetuned}, suggesting that they are not reasoning in a robust way.
Instruction tuning has been shown to improve performance and allow models to better follow in-context exemplars \citep{Mishra2021Cross,min2021metaicl,wei2021finetuned,Ye2021CrossFit,chung2022scaling}.
One shortcoming, however, is that models are not forced to learn to use the exemplars because the task is redundantly defined in the evaluation example via instructions and natural language labels.
For example, in the left-hand side of \cref{fig:pull-figure}, although the exemplars can help the model understand the task, they are not strictly necessary since the model could ignore the exemplars and just read the instruction.

To make the model better at in-context learning, we propose symbol tuning, in which the model is finetuned on exemplars where the instructions are removed and natural language labels are replaced with semantically-unrelated labels (e.g., ``Foo,'' ``Bar,'' etc.).
In this setup, the task is unclear without looking at the in-context exemplars.
For example, if the prompt from the previous paragraph was changed to ``\textit{<sentence>. Answer: \{Foo, Bar\}}'' (as shown in the right-hand side of \cref{fig:pull-figure}), multiple in-context exemplars would be needed in order to figure out the task.
Because symbol tuning teaches the model to reason over the in-context exemplars, symbol-tuned models should have much better performance on unseen tasks that require reasoning between in-context exemplars and their labels.

\section{Experimental setup}
\label{sec:symbol-tuning-and-experimental-setup}
\subsection{Tuning tasks \& prompt formatting}
\label{sec:tuning-tasks}
\begin{figure}[ht]
    \centering
    \includegraphics[width=0.95\linewidth]{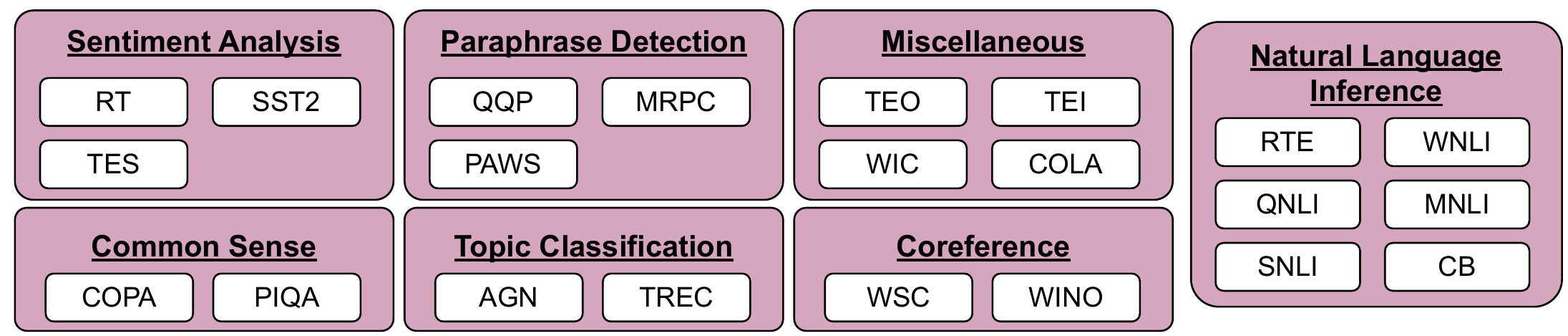}
    \caption{
    Datasets and task types used for symbol tuning.
    See \cref{sec:appendix-symbol-tuning-datasets} for dataset details.
    }
    \label{fig:tuning-datasets}
\end{figure}

\cref{fig:tuning-datasets} shows the 22 publicly-available NLP datasets from HuggingFace \citep{Lhoest2021Huggingface} (see \cref{sec:appendix-symbol-tuning-datasets} for dataset details) that we use for our symbol-tuning procedure (we ablate the number of datasets used for symbol tuning in \cref{sec:number-of-tuning-datasets}).
We selected NLP tasks that have been widely used in the literature \citep{wang2018glue,wang2019superglue}.
Each dataset is categorized into one of seven task types---we only selected classification-type tasks because symbol tuning requires discrete labels.
For each dataset, we use examples from the training split to compose prompts that we use for tuning.
Each prompt uses a randomly-selected input--label format (formats are shown in \cref{sec:appendix-prompt-formatting}) and contains a randomly-selected number between 2 and 10 of in-context exemplars per class.
We remap labels to a randomly-selected label from a set of $\sim$30k labels from three label types as shown in \cref{fig:symbols-used} (we ablate the number of labels in \cref{sec:appendix-does-symbol-tuning-require-using-all-labels} and the label types in \cref{sec:appendix-which-category-of-symbols-is-most-important}).
Examples of generated tuning prompts for each task are shown in \cref{sec:appendix-symbol-tuning-prompts}.

\subsection{Evaluation tasks}
\label{sec:evaluation-tasks}
We want to evaluate a model's ability to perform on unseen tasks, so we cannot evaluate on tasks used in symbol tuning (22 datasets) or used during instruction tuning (1.8k tasks).
Hence, we choose 11 NLP datasets from HuggingFace \citep{Lhoest2021Huggingface} that were not used in either stage of finetuning (details are shown in \cref{sec:appendix-evaluation-datasets}):
\citep[][\textbf{SUBJ}]{conneau2018senteval};
\citep[][\textbf{TEH}]{basile2019semeval};
\citep[][\textbf{TEAB}]{mohammad2016semeval};
\citep[][\textbf{TEAT}]{mohammad2016semeval};
\citep[][\textbf{TEFE}]{mohammad2016semeval};
\citep[][\textbf{TEHI}]{mohammad2016semeval};
\citep[][\textbf{ADEC}]{Alex2021RAFT};
\citep[][\textbf{OR}]{Alex2021RAFT};
\citep[][\textbf{SOT}]{Alex2021RAFT};
\citep[][\textbf{TOS}]{Alex2021RAFT};
and
\citep[][\textbf{TC}]{Alex2021RAFT}.
We use the validation split of each dataset to generate evaluation prompts.
For each dataset, we randomly select a maximum of 100 examples to use during evaluation.
Each evaluation prompt uses a randomly-selected input--label format following \cref{sec:tuning-tasks}, though we fix the number of in-context exemplars per class at $k=4$ (we ablate this parameter in \cref{sec:appendix-do-symbol-tuned-models-require-fewer-in-context-exemplars}).

\begin{figure}[ht]
    \centering
    \includegraphics[width=\linewidth]{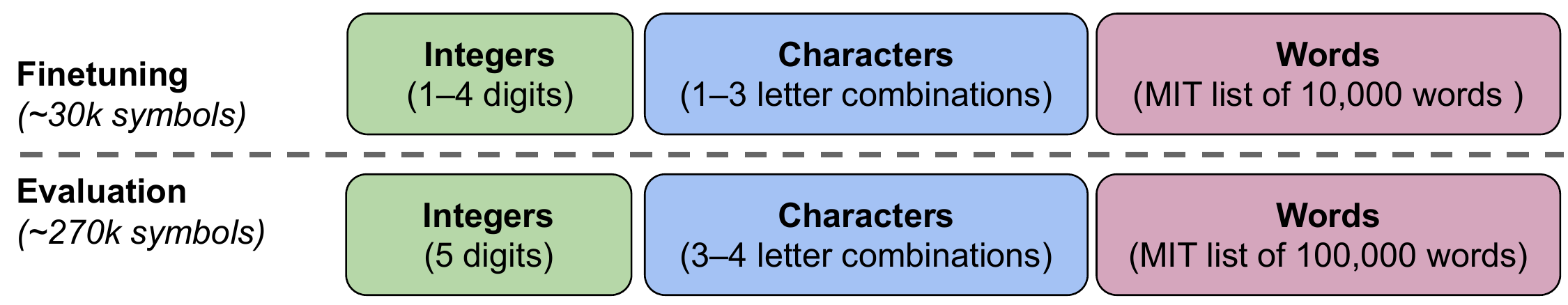}
    \caption{
    We use a set of $\sim$300k arbitrary symbols from three categories (integers, character combinations, and words). 
    $\sim$30k symbols are used during tuning and the rest are held out for evaluation.
    See \cref{sec:appendix-symbol-selection} for more details on the symbols that we used.
    }
    \label{fig:symbols-used}
\end{figure}

We generate prompts for the four different in-context learning (ICL) settings described in \cref{fig:icl-settings}; each setting either contains or does not contain instructions describing the task (see \cref{sec:appendix-evaluation-datasets} for the instructions we use for each task) and does or does not contain relevant natural language labels.
For settings that do not use relevant natural language labels, we remap original labels to a randomly-selected label from a set of approximately 270k semantically-unrelated labels as shown in \cref{fig:symbols-used} (we removed labels that were seen during symbol tuning).
Examples of generated evaluation prompts for each task are shown in \cref{sec:appendix-evaluation-prompts}.

\subsection{Models \& finetuning procedure} 
For our experiments, we tune Flan-PaLM \citep{chung2022scaling}, the instruction-tuned variants of PaLM \citep{chowdhery2022palm}.
We use instruction-tuned variants in order to reduce the number of steps needed for tuning, since symbol tuning an instruction-tuned model does not require relearning the information learned during the original round of instruction tuning.
We use three different sizes of Flan-PaLM models: Flan-PaLM-8B, Flan-PaLM-62B, and Flan-PaLM-540B.
We also tested Flan-cont-PaLM-62B \citep[][PaLM-62B at 1.3T tokens instead of 780B tokens]{chowdhery2022palm}, which we abbreviate as 62B-c.

Our symbol-tuning pipeline mixes all datasets and randomly samples from each dataset.
To ensure that the dataset sizes are balanced (i.e., no dataset gets completely overshadowed), we limit the number of training examples per dataset to a maximum of 25k randomly-selected examples.
Training examples are combined into a single sequence using packing \citep{Raffel2020Exploring}, and inputs are separated from labels using an end-of-sequence (EOS) token.
We tune all models using a batch size of 32 and the Adafactor optimizer \citep{Shazeer2018Adafactor}.
For 8B and 62B models, we tune with a learning rate of $3 \times 10^{-3}$, and we tune Flan-PaLM-540B with a learning rate of $1 \times 10^{-3}$.
We use 2048 and 512, respectively, as the input and target sequence lengths during tuning. 

Symbol tuning for 1k steps on a TPUv4 \citep{jouppi2023tpu} requires approximately 16 minutes with 64 chips for Flan-PaLM-8B, 70 minutes with 128 chips for Flan-PaLM-62B, and 6 hours with 512 chips for Flan-PaLM-540B.
For 8B and 62B model evaluations, we report results from the checkpoint after tuning for 4k steps, and for 540B model evaluations, we report results from the checkpoint after tuning for 1k steps (we ablate the number of tuning steps in \cref{sec:number-of-tuning-steps}).
See \cref{sec:appendix-tuning-procedure} for the number of finetuning steps, learning rate, batch size, and dropout used for each model.
As a baseline, we compare our symbol-tuned models against the instruction-tuned models from \citet{chung2022scaling}, and we also compare symbol tuning against continued instruction tuning in \cref{sec:appendix-are-there-results-caused-by-additional-tuning-or-the-symbol-tuning-data}.

\section{Symbol-tuned models are better in-context learners}
\label{sec:better-in-context-learning}
In the symbol-tuning procedure, models must learn to reason with in-context exemplars in order to successfully perform tasks because prompts are modified to ensure that tasks cannot simply be learned from natural language labels or instructions.
Symbol-tuned models should thus perform better in settings where tasks are unclear and require reasoning between in-context exemplars and their labels.
Additionally, since symbol tuning is meant to improve the ability to follow in-context exemplars, it should not modify prior knowledge and should thus retain the same performance in settings where exemplars are not as necessary to complete the task.

\begin{figure}[th]
    \centering
    \includegraphics[width=0.95\linewidth]{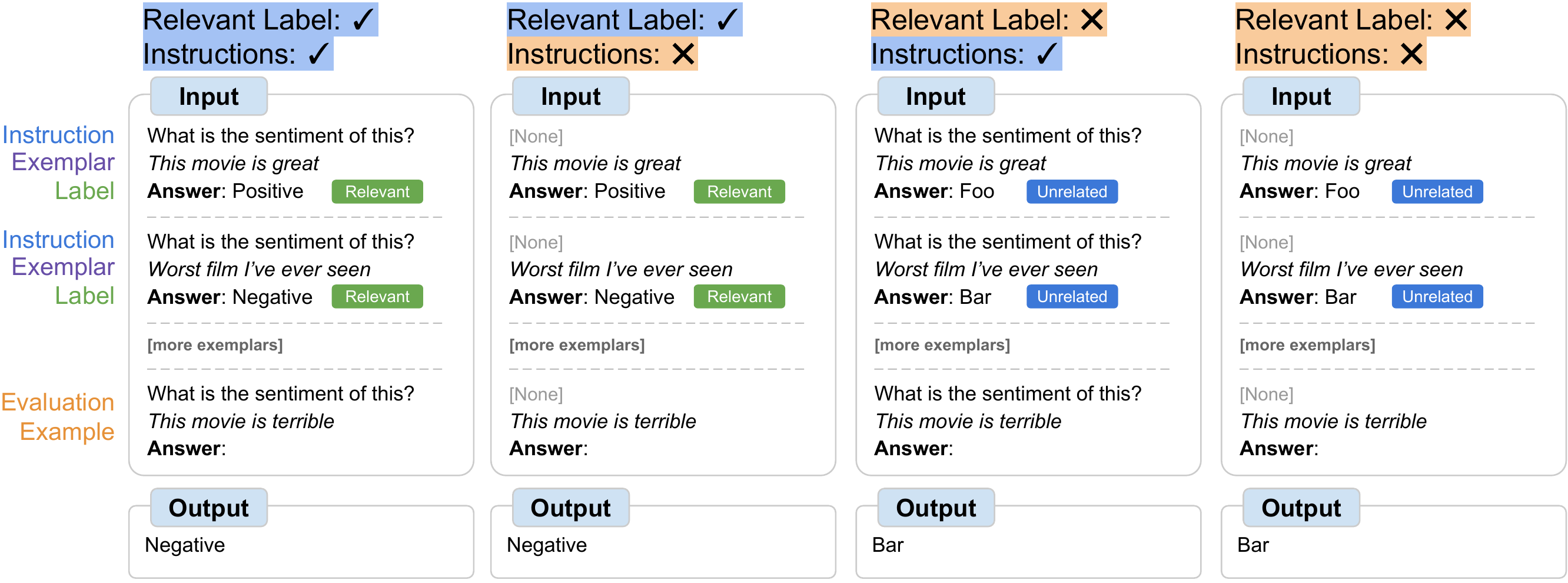}
    \caption{
    Depending on the availability of instructions and relevant natural language labels, models may need to do varying amounts of reasoning with in-context exemplars. 
    When these features are not available, models must reason with the given in-context exemplars in order to successfully perform the task.
    When they are available, reasoning with exemplars can help but is not necessary.
    }
    \label{fig:icl-settings}
\end{figure}

To explore these settings, we define four ICL settings that vary the amount of reasoning required between inputs and labels in order to learn the task (based on the availability of instructions/relevant labels), as shown in \cref{fig:icl-settings}.
The easiest of these settings uses prompts where both instructions and relevant labels are available (as in-context exemplars are not necessary to learn the task), while the hardest setting uses prompts where instructions and relevant labels are both unavailable.

In \cref{tab:icl-main-comparison}, we evaluate model performance before and after symbol tuning in each of these settings.
We find that symbol tuning improves performance across all ICL settings for models 62B and larger, with small improvements in settings with relevant natural language labels (+0.8\% to +4.2\%) and substantial improvements in settings without relevant natural language labels (+5.5\% to +15.5\%).
Strikingly, when relevant labels are unavailable, symbol-tuned Flan-PaLM-8B outperforms Flan-PaLM-62B, and symbol-tuned Flan-PaLM-62B outperforms Flan-PaLM-540B.
This performance difference suggests that symbol tuning can allow much smaller models to perform as well as large models on learning input-label mapping from exemplars (effectively saving $\sim$10x inference compute).

Symbol-tuned models also perform somewhat-comparably in settings with only relevant labels or only instructions, unlike baseline models whose performance in settings with only relevant labels is always better than in settings with only instructions.
Performance in settings with relevant labels actually decreases for Flan-PaLM-8B after symbol-tuning, however, which may suggest that symbol tuning a small model can override its prior knowledge due to overfitting.
Overall, the improvements demonstrate the strong potential of symbol tuning to improve model performance, especially when tasks are not clear and require learning from in-context exemplars.

\begin{table}[hb]
    \centering
    \scriptsize
    \arrayrulecolor{black}
    \resizebox{0.925\columnwidth}{!}{
    \begin{tabular}{l l l l l}
        \toprule
        & \multicolumn{4}{c}{\thead{\scriptsize \textbf{Average performance on eleven tasks}}} \\
        \cmidrule[0.5pt]{2-5}
        \hspace{9mm} \textbf{Relevant labels:} & \colorbox{babyblue}{\textbf{\cmark}} & \colorbox{babyblue}{\textbf{\cmark}} & \colorbox{fluorescentorange}{\textbf{\xmark}} & \colorbox{fluorescentorange}{\textbf{\xmark}} \\
        \hspace{7mm} \textbf{Task instructions:} & \colorbox{babyblue}{\textbf{\cmark}} & \colorbox{fluorescentorange}{\textbf{\xmark}} & \colorbox{babyblue}{\textbf{\cmark}} & \colorbox{fluorescentorange}{\textbf{\xmark}} \\
        \midrule
        Random Guessing \hspace{10mm} & 42.4 & 42.4 & 42.4 & 42.4 \\
        \midrule
        Flan-PaLM-8B & 63.9 & 61.6 & 42.4 & 44.2 \\
        \ \ \ \ \ \ + Symbol tuning (ours) & 57.6 (\textcolor{red}{\textbf{-6.3}}) & 54.3 (\textcolor{red}{\textbf{-7.3}}) & 58.2 (\textcolor{darkpastelgreen}{\textbf{+15.8}}) & 52.8 (\textcolor{darkpastelgreen}{\textbf{+8.6}}) \\
        \\
        Flan-PaLM-62B & 74.3 & 70.0 & 57.0 & 50.5 \\
        \ \ \ \ \ \ + Symbol tuning (ours) & 75.5 (\textcolor{darkpastelgreen}{\textbf{+1.2}}) & 70.8 (\textcolor{darkpastelgreen}{\textbf{+0.8}}) & 71.4 (\textcolor{darkpastelgreen}{\textbf{+14.4}}) & 60.3 (\textcolor{darkpastelgreen}{\textbf{+9.8}}) \\
        \\
        Flan-cont-PaLM-62B & 77.3 & 70.3 & 56.3 & 51.0 \\
        \ \ \ \ \ \ + Symbol tuning (ours) & 78.9 (\textcolor{darkpastelgreen}{\textbf{+1.6}}) & 74.5 (\textcolor{darkpastelgreen}{\textbf{+4.2}}) & 71.8 (\textcolor{darkpastelgreen}{\textbf{+15.5}}) & 62.1 (\textcolor{darkpastelgreen}{\textbf{+11.1}}) \\
        \\
        Flan-PaLM-540B & 82.2 & 77.4 & 70.7 & 58.1 \\
        \ \ \ \ \ \ + Symbol tuning (ours) & 84.4 (\textcolor{darkpastelgreen}{\textbf{+2.2}}) & 78.8 (\textcolor{darkpastelgreen}{\textbf{+1.4}}) & 80.0 (\textcolor{darkpastelgreen}{\textbf{+9.3}}) & 63.6 (\textcolor{darkpastelgreen}{\textbf{+5.5}}) \\
        \bottomrule
    \end{tabular}
    }
    \caption{
    Large-enough symbol-tuned models are better at in-context learning than baselines, especially in settings where relevant labels are not available.
    Performance is shown as average model accuracy (\%) across eleven tasks (per-task results are shown in \cref{sec:appendix-icl-settings}).
    }
    \label{tab:icl-main-comparison}
\end{table}

\clearpage
\section{Symbol tuning improves algorithmic reasoning}
\label{sec:better-reasoning}
Symbol tuning is designed to force the model to learn from input--label mappings in the in-context exemplars because the symbols are unrelated to the task and no instructions are provided (and thus the model cannot rely on any other guidance to determine the task).
For this reason, we posit that symbol tuning should not only improve the model's ability to map natural language inputs to arbitrary symbols, but also its ability to learn other forms of inputs--label mappings such as algorithms.

To test this, we experiment on algorithmic reasoning tasks from BIG-Bench \citep{bigbench}.
We first experiment on a set of list function tasks \citep{Rule2020Child,bigbench} where the model needs to identify a transformation function (e.g., remove the last element in a list) between input and output lists containing non-negative integers.
These tasks were evaluated in a four-shot setting, following our evaluation setup in \cref{sec:evaluation-tasks}.
Additionally, we test models on a set of simple turing concepts \citep{telle2019teaching,bigbench} where models need to reason with binary strings to learn the concept that maps an input to an output (e.g., swapping 0s and 1s in a string).
These tasks have predetermined shots for each evaluation example.
We selected these algorithmic tasks because they test the model's ability to generalize to different task types (the symbol-tuning tasks were classification problems with discrete labels, while these tasks are more open-ended generation problems) and do not require world knowledge (symbol tuning does not increase prior knowledge).

In \cref{fig:algorithmic-reasoning-main-comparison}, we show model performance on the twenty list function tasks with the highest human accuracy baselines\footnote{We do not directly compare with the human baselines because our evaluation format was different.} \citep{rule2020thesis} separated into five categories (category details are described in \cref{sec:appendix-big-bench-list-functions}) and the turing concepts containing 3 or fewer instructions in the AS II subset of the simple turing concepts task.
On the list function tasks, symbol tuning results in an average performance improvement across all tasks of 18.2\% for Flan-PaLM-8B, 11.1\% for Flan-PaLM-62B, 15.5\% for Flan-cont-PaLM-62B, and 3.6\% for Flan-PaLM-540B.
On the turing concept tasks, symbol tuning results in a performance improvement of 15.3\% for Flan-PaLM-8B and Flan-PaLM-62B, 14.1\% for Flan-cont-PaLM-62B, and 4.7\% for Flan-PaLM-540B.
Flan-cont-PaLM-62B with symbol tuning outperforms Flan-PaLM-540B on the list function tasks (in terms of average accuracy across tasks), which is equal to a $\sim$10x reduction in inference compute.
These improvements on an unseen task type suggest that symbol tuning indeed strengthens the model's ability to learn in-context, as the symbol-tuning procedure did not include any algorithmic data and only used natural language data.

\input{Figures/algorithmic-reasoning-main-comparison}

\clearpage
\section{Symbol-tuned models can override priors via flipped labels}
\label{sec:follow-flipped-labels}
\citet{wei2023larger} showed that while pretrained language models (without instruction tuning) could, to some extent, follow flipped labels presented in-context, instruction tuning degraded this ability.
Symbol tuning, on the other hand, forces models to consider the label presented in-context as an arbitrary symbol, which should reduce the model's usage of prior knowledge that contradicts the flipped labels.
For this reason, we expect that symbol tuning would be able to improve and restore the ability to follow flipped labels in-context.

To test this, we flip the labels of both in-context exemplars and the evaluation example for the tasks described in \cref{sec:evaluation-tasks} (we remove tasks with more than two labels from this experiment since it is unclear how to best ``flip'' more than two labels).
For example, for the SST2 dataset, all exemplars that are labeled as having ``positive'' sentiment will now be labeled as having ``negative'' sentiment.
A perfect model that can follow these flipped labels should achieve 100\% accuracy on these tasks if its accuracy on the standard in-context learning setting is also 100\%.

As shown in \cref{fig:flipped-label-main-comparison}, symbol tuning restores the ability to follow flipped labels that was lost during instruction tuning.
We see that there is a similar trend across all model sizes---instruction-tuned models are generally unable to follow flipped labels (as demonstrated by their performance being far below random guessing), but symbol-tuned models are much more capable of doing so.
We found that after symbol tuning, Flan-PaLM-8B sees an average improvement across all datasets of 26.5\%, Flan-PaLM-62B sees an improvement of 33.7\%, and Flan-PaLM-540B sees an improvement of 34.0\%.
For some datasets (e.g., OR, SUBJ, TC), symbol-tuned models can now override priors and follow flipped labels (i.e., achieve much better performance than random guessing), despite instruction-tuned models not being able to do so for any datasets.
Additionally, symbol-tuned models achieve similar or better average performance as pretraining-only models, indicating that symbol tuning has, to some extent, restored the model's original ability to follow flipped labels.

These results further indicate another type of generalized in-context learning capability, as we did not include any flipped labels during symbol tuning.
Although the performance improvement from symbol tuning is large, we note that more work should be done in this area since performance on the flipped-labels settings is, on average, not significantly better than random guessing.

\input{Figures/flipped-label-main-comparison}

\clearpage
\section{Ablation studies}
\label{sec:ablations}

\subsection{Number of tuning steps}
\label{sec:number-of-tuning-steps}
A question that may come to mind is how many steps of finetuning is needed to get the benefits of symbol tuning.
In particular, \citet{chung2022scaling} performed instruction tuning on PaLM models for 40k steps for PaLM-8B and PaLM-62B, 21k steps for PaLM-540B, and 60k steps for cont-PaLM-62B, so it is unclear if symbol tuning would require such extensive tuning.
Intuitively, however, since our symbol-tuning dataset is much smaller than the tuning data from \citet{chung2022scaling}, symbol tuning should require fewer steps for finetuning than instruction tuning does.
To analyze this, we examine model performance in each of the four ICL settings from \cref{fig:icl-settings} with respect to the number of steps tuned.
We train 8B and 62B models for up to 10k steps and 540B models for up to 5k steps, and we evaluate checkpoints every 1k steps on the same evaluation tasks and settings from \cref{sec:better-in-context-learning}.

We show these results in \cref{fig:tuning-steps-main-ablation}.
As expected, we see that symbol tuning does not require many steps of finetuning for any model.
Moreover, the largest changes in performance occur within the first 1k to 2k steps of symbol tuning, after which model performance stays relatively constant.
Flan-PaLM-540B also seems to experience performance drops in all settings after 1k steps, which may indicate that larger models require a more-diverse or larger set of symbol-tuning data.
These results suggest that symbol tuning does not require extensive compute for exhaustive tuning.

\input{Figures/tuning-steps-main-ablation}

\input{Figures/flan-mixture-main-ablation}

\subsection{Mixing instruction-tuning data}
\label{sec:mixing-instruction-tuning-data}
In \cref{sec:better-in-context-learning}, we found that small models may actually overfit to the symbol-tuning data, resulting in performance drops in ICL settings where relevant labels are available.
One potential way of preventing this is to include instruction-tuning data during symbol tuning.
Since instruction-tuning examples contain relevant labels and instructions that match a model's prior knowledge, they may help reinforce prior knowledge and prevent small models from ``forgetting'' their priors.
We create several mixtures of instruction-tuning data and symbol-tuning data to test this idea.
For each mixture, we use varying ratios of instruction-tuning data to symbol-tuning data (e.g., a mixture with 33.3\% symbol-tuning data means that instruction-tuning data is weighted twice as heavily as symbol-tuning data).
Our instruction-tuning data is directly taken from \citet{chung2022scaling} and then mixed with our symbol-tuning data from \cref{sec:tuning-tasks}.

We then tune models on these mixtures and evaluate their performance.\footnote{We exclude Flan-PaLM-540B from this ablation study to reduce computational costs.}
In \cref{fig:flan-mixture-main-ablation}, we show model performance on the ICL settings from \cref{sec:better-in-context-learning}.
We find that even a small mixture of symbol-tuning data (e.g., 16\%) versus instruction-tuning data can significantly change model performance.
\begin{wrapfigure}{r}{5cm}
    \vspace{-1mm}
    \centering
    \begin{tikzpicture}
        \pgfplotsset{footnotesize,samples=10}
        \begin{groupplot}[
            group style = {group size = 4 by 1, horizontal sep = 20pt},
            width = \linewidth, 
            height = \linewidth]
            \nextgroupplot[
                align = center,
                legend style={at={(-0.12,1.4)},anchor=south},
                xmin=-5, xmax=105,
                ymin=15, ymax=75,
                xtick={0, 16, 33, 50, 66, 83, 100},
                xticklabels={0, 16, 33, 50, 66, 83, 100},
                axis x line*=bottom,
                axis y line*=left,
                xlabel={\% Symbol tuning data \\ (rest is instruction tuning data)},,
                ylabel={Accuracy (\%)},
                ytick={20, 30, 40, 50, 60, 70},
                yticklabels={20, 30, 40, 50, 60, 70},
                grid style=dashed,
                x label style={at={(axis description cs:0.5,-0.15)},anchor=north},
                y label style={at={(axis description cs:-0.15,0.5)},anchor=south},
                xtick pos=bottom,
                ytick pos=left,
                legend cell align=left,
                    legend style={
                            at={(-0.05,-0.65)},
                            anchor=west,
                            column sep=1ex,
                            font=\small,
                            draw=none,
                    }
                ]
                \addplot[
                    color=codexcolorone,
                    mark=\palmshape,
                    mark size=1.5pt,
                    line width=1pt,
                    ]
                    coordinates {
                    (0, 26.3)
                    (16, 35.8)
                    (33, 38.2)
                    (50, 43.5)
                    (66, 50.8)
                    (83, 51.8)
                    (100, 53.0)
                    };
                    \addlegendentry{Flan-PaLM-8B}
                \addplot[
                    color=codexcolortwo,
                    mark=\palmshape,
                    mark size=1.5pt,
                    line width=1pt,
                    ]
                    coordinates {
                    (0, 24.2)
                    (16, 35.5)
                    (33, 37.8)
                    (50, 43.7)
                    (66, 49.0)
                    (83, 50.2)
                    (100, 57.5)
                    };
                    \addlegendentry{Flan-PaLM-62B}
                \addplot[
                    color=codexcolorthree,
                    mark=\palmshape,
                    mark size=1.5pt,
                    line width=1pt,
                    ]
                    coordinates {
                    (0, 26.5)
                    (16, 38.7)
                    (33, 55.2)
                    (50, 56.7)
                    (66, 56.3)
                    (83, 61.0)
                    (100, 62.3)
                    };
                    \addlegendentry{Flan-cont-PaLM-62B}
        \end{groupplot}
    \end{tikzpicture}
    \caption{
    Tuning models using mixtures with a higher proportion of symbol-tuning data results in better performance in the flipped label setting.
    Performance is shown using the average accuracy across the six datasets from \cref{sec:follow-flipped-labels}.
    }
    \label{fig:flan-mixture-flipped-label-ablation}
    \vspace{-9mm}
\end{wrapfigure}
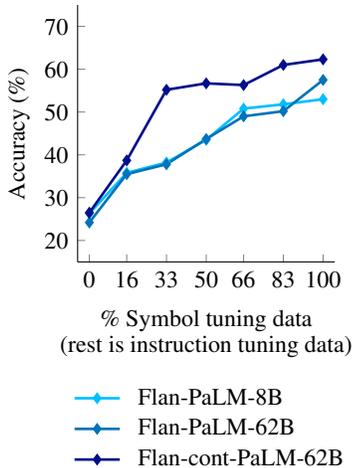
\hspace{-1mm}Furthermore, higher proportions of symbol-tuning data after this initial change generally do not significantly affect model performance.\footnote{Flan-PaLM-8B experiences a performance drop in the settings that include relevant natural language labels, which was also seen in \cref{sec:better-in-context-learning}.}
These results indicate that, in terms of a model's ability to succeed in these ICL settings, the proportion of symbol-tuning data used is not important as long as some non-trivial amount of symbol-tuning data is used.
As shown in \cref{fig:flan-mixture-flipped-label-ablation}, however, the proportion of symbol-tuning data is much more impactful for succeeding in flipped-label settings.
We find that there is a strong correlation between a higher mixture of symbol-tuning data and a model's ability to follow flipped labels, a trend that holds regardless of the size of the model.
Combining this result with the trend shown in \cref{fig:flan-mixture-flipped-label-ablation}, we propose using only symbol-tuning data as a default setting because it does not significantly decrease model performance (for large-enough models) and because a higher percentage of symbol-tuning data significantly improves the model's ability to override prior knowledge with in-context exemplars.

\input{Figures/num-tuning-datasets-main-ablation}

\subsection{Number of tuning datasets}
\label{sec:number-of-tuning-datasets}
The overall goal of symbol tuning is to teach models that any arbitrary label for an input--label mapping should be treated as a symbol to be learned.
The symbol-tuning procedure should thus only be successful if a diverse-enough set of tasks are shown such that the model can learn to generalize its behavior to new tasks.
To test this, we randomly remove a varying number of tasks from the mixture and retune models on these new mixtures.\footnote{We exclude Flan-PaLM-540B from this ablation study to reduce computational costs.}
We then evaluate these models on the ICL settings from \cref{sec:better-in-context-learning}.

We show these results in \cref{fig:num-tuning-datasets-main-ablation}.
First, we see that as a general trend, using more datasets for symbol tuning improves performance.
This effect seems to slightly plateau as more datasets are added, and 62B models benefit more from added datasets than the 8B model does.
Second, we find that symbol tuning with a small number of datasets (e.g., only one or two datasets) can hurt performance in settings where relevant labels are available.
For example, while symbol tuning using just one dataset can significantly improve performance in settings without relevant labels, it simultaneously decreases model performance in settings where relevant labels are available.
These results imply that symbol tuning works best when a large variety of tasks are used, and symbol tuning with only a small number of tasks may result in models that perform worse in settings with relevant labels.
Given these results, we note that future work may be needed to investigate the effects of scaling up the symbol-tuning procedure.

\section{Related work}
\label{sec:related-work}
\subsection{In-context learning via semantic prior knowledge}
Recent studies on in-context learning suggest that prior knowledge plays a significant role in how models learn in-context.
For example, \citet{wei2023larger} showed that some small models and instruction-tuned models cannot follow flipped labels presented in-context, suggesting that these models primarily utilize prior knowledge for in-context learning.
\citet{min2022rethinking} found a similar result that using random ground-truth labels in in-context exemplars does not significantly affect performance, meaning that performance may be driven by other factors such as the label space.

\citet{reynolds2021prompt} also showed that cleverly-constructed prompts in a zero-shot setting could outperform prompts in a few-shot setting, implying that, for some tasks, models can achieve better performance by leveraging their existing knowledge than from attempting to learn the task from in-context exemplars.
Additionally, in chain-of-thought prompting \citep{wei2022chain}, \citet{madaan2022text} and \citet{wang2022towards} showed that performance on multi-step reasoning tasks does not decrease when models are provided with logically-incorrect prompts.
\citet{Raghu2020Rapid} also demonstrated that systems such as MAML can effectively ``memorize'' labels when trained in a way where all labels can be memorized, which further illustrates that, when possible, models may attempt to use prior knowledge rather than adapt to each new task.

Our findings do not dispute the idea that semantic prior knowledge can provide significant benefits to in-context learning.
Indeed, we showed that instruction-tuned models cannot follow flipped labels in-context, which is consistent with the findings from \citet{wei2023larger}.
We instead aim to demonstrate that through symbol tuning, language models can retain the benefits of utilizing prior knowledge while also improving their ability to learn from the input--label pairs shown in the in-context exemplars.

\subsection{In-context learning via in-context exemplars}
At the same time, however, other recent work has suggested that language models can, in fact, learn in-context using the given exemplars.
This ability may be more useful than the ability to use semantic prior knowledge because it would allow models to perform tasks that are not seen in or contradict pretraining data.
\citet{Garg2022What}, for instance, showed that transformers trained from scratch can perform in-context learning on linear-regression tasks at a similar performance level as the least-squares estimator.
This capability was shown to result from transformers implementing standard learning algorithms such as gradient descent \citep{Akyurek2022What,Oswald2022Transformers,Dai2022Why}.
Furthermore, \citet{Webson2021Do} demonstrated that, in a natural language setting, language models can learn at the same rate during finetuning even when given irrelevant or misleading prompts.
On a broader level, \citet{Rajendran2020Meta} and \citet{yin2020metalearning} found that adding noise to, shuffling, or regularizing the label space can make systems better at learning and adapting to new tasks.
In this paper, we attempt to improve the degree to which language models are able to learn tasks via input--label mappings.
Our symbol-tuning method can be seen as a form of label augmentation and is thus similar to the proposed methods from \citet{Rajendran2020Meta} and \citet{yin2020metalearning}, though it differs crucially in that we apply them to tune large language models.
We found that symbol-tuned models saw significant improvements in their ability to learn in-context (e.g., on algorithmic tasks or settings with underspecified prompts).

\subsection{Tuning language models}
Our work presented symbol tuning, a form of finetuning on input--label pairs where labels are remapped to arbitrary symbols.
Symbol tuning relates to a broader body of work showing that finetuning language models can significantly alter their behavior and performance in different settings.
For example, \citet{wei2021finetuned} first presented instruction tuning (finetuning on tasks phrased as instructions) and showed that this finetuning procedure substantially improves model performance in zero-shot settings.
\citet{chung2022scaling} further scaled this procedure by adding more tasks, increasing model sizes, and adding chain-of-thought data, demonstrating that, with these changes, tuned models are significantly better at chain-of-thought reasoning, open-ended generation, and several evaluation benchmarks.
Our experimental findings match these results, though our work differs by not only focusing on settings with in-context exemplars and underspecified prompts, but also by modifying the tuning procedure to make tasks harder to learn and require additional reasoning with exemplars.

\section{Conclusions}
\label{sec:discussion}
In this paper, we presented \textit{symbol tuning}, a new method of tuning models on tasks where natural language labels are remapped to arbitrary symbols.
Symbol tuning is based off of the intuition that when models cannot use instructions or relevant labels to determine a presented task, it must do so by instead learning from in-context exemplars.
We tuned four language models (Flan-PaLM-8B, Flan-PaLM-62B, Flan-cont-PaLM-62B, and Flan-PaLM-540B) using our symbol-tuning procedure, utilizing a tuning mixture of 22 datasets and approximately 30k arbitrary symbols as labels.

Experimentally, we showed that symbol tuning can significantly improve a model's ability to learn from in-context exemplars in not only natural language settings, but also on algorithmic tasks.
First, we showed that symbol tuning improves performance on unseen in-context learning tasks, especially when prompts do not contain instructions or relevant labels.
We also found that symbol-tuned models were much better at algorithmic reasoning tasks, despite the lack of numerical or algorithmic data in the symbol-tuning procedure.
Moreover, in an in-context learning setting where inputs have flipped labels, symbol tuning (for some datasets) reunlocks the ability to follow flipped labels that was lost during instruction tuning.
Finally, we demonstrated that symbol tuning does not require extensive compute or complex implementations in order to achieve these improvements.

Through symbol tuning, we aim to have increased the degree to which models can examine and learn from input--label mappings during in-context learning.
We hope that our results encourage further work towards improving language models' ability to reason over symbols presented in-context.

\clearpage
\bibliography{symbol-tuning}
\bibliographystyle{iclr2023_conference}

\clearpage
\appendix
\addcontentsline{toc}{section}{Appendix} 
\part{Appendix} 
\parttoc
\clearpage

\section{Frequently Asked Questions}
\label{sec:appendix-frequently-asked-questions}
\subsection{Are these results caused by additional tuning or the symbol tuning data?}
\label{sec:appendix-are-there-results-caused-by-additional-tuning-or-the-symbol-tuning-data}
One unanswered question that arises is whether our results come from the symbol-tuning data or whether they come from the additional steps of tuning.
To answer this question, we continue tuning Flan-PaLM models using the same instruction-tuning mixture from \citet{chung2022scaling} for the same number of steps that the model was symbol tuned using (see \cref{sec:appendix-tuning-procedure}).
We then compare these instruction-tuned models with our symbol-tuned models on each reasoning task from \cref{sec:better-reasoning}, the flipped-label setting from \cref{sec:follow-flipped-labels}, and the ICL settings from \cref{sec:better-in-context-learning} in \cref{tab:appendix-instruct-tune-vs-symbol-tune}.\footnote{We exclude comparisons on the ICL settings with relevant natural language labels because, as shown in \cref{sec:better-in-context-learning}, symbol tuning did not significantly improve performance in these settings.}

We find that our symbol-tuned models significantly outperform the models with continued instruction tuning on each of these evaluations.
These results suggest that, indeed, the performance improvements on these tasks were not a result of simply tuning the model for more steps.
Instead, we conclude that the symbol-tuning data itself is the root cause of the results we observed in this paper.

\begin{table}[bh]
    \centering
    \scriptsize
    \arrayrulecolor{black}
    \setlength{\tabcolsep}{3pt}
    \resizebox{\columnwidth}{!}{
    \begin{tabular}{l l l l l l}
        \toprule
            & \multicolumn{2}{c}{\thead{\footnotesize Algorithmic Reasoning}} & \multicolumn{3}{c}{\thead{\footnotesize In-Context Learning}} \\
        \cmidrule[0.5pt](lr){2-3}
        \cmidrule[0.5pt](lr){4-6}
            \multirow{2}{*}{\footnotesize Model} & \thead[l]{\scriptsize Turing \\ \scriptsize Concepts} & \thead[l]{\scriptsize List \\ \scriptsize Functions} & \thead[l]{\scriptsize Flipped \\ \scriptsize Labels} & \thead[l]{\scriptsize No Relevant Target \\ \scriptsize + Instruction} & \thead[l]{\scriptsize No Relevant Target \\ \scriptsize + No Instruction} \\
        \midrule
            Random Guessing \hspace{17.5mm} & 0 \hspace{14mm} & 0 \hspace{14mm} & 50 \hspace{14mm} & 42.4 \hspace{14mm} & 42.4 \hspace{14mm} \\
        \midrule
            \textbf{Flan-PaLM-8B} & 17.6 & 19.2 & 26.5 & 42.4 & 44.2 \\
            \ \ \ \ \ \ + Instruction tuning & 16.5 & 23.1 & 26.3 & 44.4 & 45.6 \\
            \ \ \ \ \ \ + Symbol tuning (ours) & 32.9 (\textcolor{darkpastelgreen}{+16.4}) & 37.4 (\textcolor{darkpastelgreen}{+14.3}) & 53.0 (\textcolor{darkpastelgreen}{+23.7}) & 58.2 (\textcolor{darkpastelgreen}{+13.8}) & 52.8 (\textcolor{darkpastelgreen}{+7.2}) \\
            \\
            \textbf{Flan-PaLM-62B} & 61.2 & 56.1 & 23.8 & 57.0 & 50.5 \\
            \ \ \ \ \ \ + Instruction tuning & 54.1 & 56.3 & 24.2 & 59.9 & 54.3 \\
            \ \ \ \ \ \ + Symbol tuning (ours) & 76.5 (\textcolor{darkpastelgreen}{+22.4}) & 67.2 (\textcolor{darkpastelgreen}{+10.9}) & 57.5 (\textcolor{darkpastelgreen}{+33.3}) & 71.4 (\textcolor{darkpastelgreen}{+11.5}) & 60.3 (\textcolor{darkpastelgreen}{+6.0}) \\
            \\
            \textbf{Flan-cont-PaLM-62B} & 64.7 & 54.7 & 27.3 & 56.3 & 51.0 \\
            \ \ \ \ \ \ + Instruction tuning & 68.2 & 65.0 & 26.5 & 59.0 & 52.4 \\
            \ \ \ \ \ \ + Symbol tuning (ours) & 78.8 (\textcolor{darkpastelgreen}{+10.6}) & 70.2 (\textcolor{darkpastelgreen}{+5.2}) & 62.3 (\textcolor{darkpastelgreen}{+35.8}) & 71.8 (\textcolor{darkpastelgreen}{+12.8}) & 62.1 (\textcolor{darkpastelgreen}{+9.7}) \\
            \\
            \textbf{Flan-PaLM-540B} & 63.5 & 69.5 & 20.7 & 70.7 & 58.1 \\
            \ \ \ \ \ \ + Instruction tuning & 61.2 & 68.9 & 19.2 & 73.6 & 59.5 \\
            \ \ \ \ \ \ + Symbol tuning (ours) & 68.2 (\textcolor{darkpastelgreen}{+7.0}) & 73.1 (\textcolor{darkpastelgreen}{+4.2}) & 54.7 (\textcolor{darkpastelgreen}{+35.5}) & 80.0 (\textcolor{darkpastelgreen}{+6.4}) & 63.6 (\textcolor{darkpastelgreen}{+4.1}) \\
        \bottomrule
    \end{tabular}
    }
    \caption{
    Symbol-tuned models perform better than instruction-tuned models on the turing concept and list function tasks from \cref{sec:better-reasoning}, the flipped-label setting from \cref{sec:follow-flipped-labels}, and the ICL settings without relevant labels from \cref{sec:better-in-context-learning}.
    Performance change is calculated by subtracting the instruction-tuned model's performance from the symbol-tuned model's performance.
    Evaluation setups are the same for each task as they were in the respective section that introduced them; performance is shown as the accuracy (\%) averaged across all subtasks.
    Per-task results for list function tasks from \cref{sec:better-reasoning} are shown in \cref{sec:appendix-big-bench-list-functions}.
    Per-task results for ICL settings from \cref{sec:better-in-context-learning} are shown in \cref{sec:appendix-icl-settings}.
    }
    \label{tab:appendix-instruct-tune-vs-symbol-tune}
\end{table}

\subsection{Does symbol tuning affect performance on benchmarks?}
\label{sec:appendix-does-symbol-tuning-affect-performance-on-benchmarks}
As shown in \cref{sec:better-in-context-learning}, symbol-tuned models see only minor performance improvements in ICL settings with relevant labels, and small models (e.g., Flan-PaLM-8B) experience performance drops on these settings after symbol tuning.
A natural question that follows is whether these differences on our unseen tasks translate to similar differences in well-studied benchmarks, as examples from these benchmarks often contain instructions and relevant labels.
In particular, we examine model performance on the MMLU \citep{Hendrycks2021MMLU} and BIG-Bench Hard \citep{suzgun2022challenging} benchmarks.
For this experiment, we set prompts in a 5-shot setting for MMLU and a 3-shot setting for BIG-Bench Hard, following the settings used in \citet{chung2022scaling}.

\begin{figure}[ht]
    \centering
    \pgfplotsset{
        width=0.45\linewidth, 
        height=0.425\linewidth,
        /pgfplots/ybar legend/.style={
            /pgfplots/legend image code/.code={
                \draw[##1,/tikz/.cd,yshift=-0.25em]
                (0cm,0cm) rectangle (7pt,0.8em);
            },
        },
    }
    \centering
    \begin{tikzpicture}
        \begin{groupplot}[
            group style={
            group name=plot,
            horizontal sep=40pt,
            vertical sep=20pt,
            group size=2 by 1},]
            \nextgroupplot[
                ybar=0pt,
                ymin=0, ymax=105,
                ytick={0, 10, 20, 30, 40, 50, 60, 70, 80, 90, 100},
                major x tick style = transparent,
                bar width=9 pt,
                enlarge x limits=0.25,
                typeset ticklabels with strut,
                ylabel={Accuracy (\%)},
                title={\textbf{MMLU}},
                symbolic x coords={8B, 62B, 62B-c, 540B},  
                xtick=data,  
                axis x line*=bottom,
                axis y line*=none,
                legend cell align=left,
                    legend style={
                            at={(1.1,-0.2)},
                            anchor=north,
                            column sep=1ex,
                            font=\small,
                            draw=none,
                            legend columns=3,
                    },
                ]  
                \addplot[ybar, fill=flancolorone,  postaction={}] coordinates {
                    (8B, 49.5)
                    (62B, 59.8)
                    (62B-c, 65.3)
                    (540B, 73)
                };  
                \addplot[ybar, fill=symflancolorone,  postaction={}] coordinates {
                    (8B, 47.5)
                    (62B, 58.6)
                    (62B-c, 64.9)
                    (540B, 72.8)
                };
                \legend{
                    Flan-PaLM \ \ \ \ \ \ \ \ \ \ ,
                    Flan-PaLM + Symbol tuning (ours)
                }
            \nextgroupplot[
                ybar=0pt,
                ymin=0, ymax=105,
                ytick={0, 10, 20, 30, 40, 50, 60, 70, 80, 90, 100},
                major x tick style = transparent,
                bar width=8 pt,
                enlarge x limits=0.25,
                typeset ticklabels with strut,
                title={\textbf{BIG-Bench Hard}},
                symbolic x coords={8B, 62B, 62B-c, 540B},  
                xtick=data,  
                axis x line*=bottom,
                axis y line*=none,
                legend cell align=left,
                    legend style={
                            at={(0.5,-0.2)},
                            anchor=north,
                            column sep=1ex,
                            font=\small,
                            draw=none,
                            legend columns=1,
                    },
                ]  
                \addplot[ybar, fill=flancolorone,  postaction={}] coordinates {
                    (8B, 36.2)
                    (62B, 47.1)
                    (62B-c, 50.5)
                    (540B, 57.8)
                };
                \addplot[ybar, fill=symflancolorone,  postaction={}] coordinates {
                    (8B, 33.0)
                    (62B, 46.1)
                    (62B-c, 50.1)
                    (540B, 57.9)
                };  
        \end{groupplot}
    \end{tikzpicture}
    \caption{
    Performance on MMLU and BIG-Bench Hard does not significantly change after symbol tuning.
    Accuracy shown is an unweighted average over all tasks for each benchmark (per-task results are shown in \cref{sec:appendix-mmlu} and \cref{sec:appendix-big-bench-hard}).
    }
    \label{fig:appendix-benchmark-performance}
\end{figure}
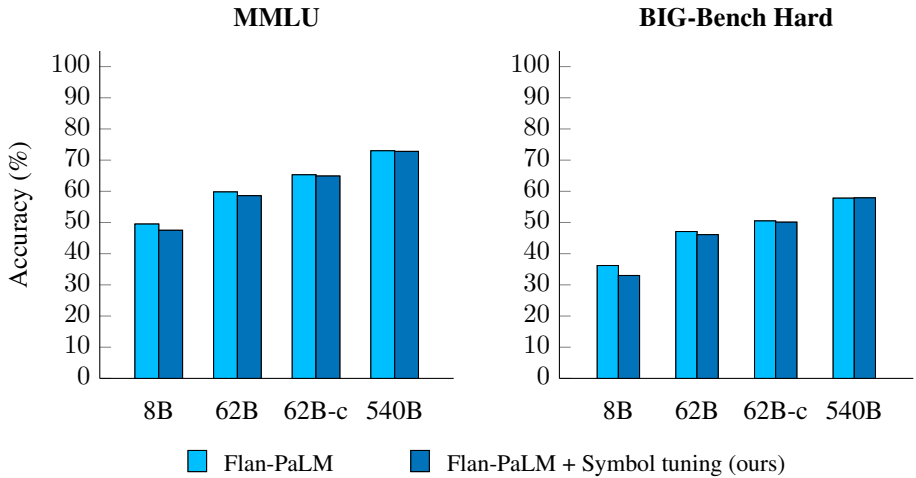

In \cref{fig:appendix-benchmark-performance}, we show model performance on these benchmarks for each symbol-tuned model.
We find that small models (i.e., Flan-PaLM-8B) may experience minor performance drops after symbol tuning.
This aligns with the result shown in \cref{sec:better-in-context-learning} and further bolsters the possibility that, after symbol tuning, small models may tend to use prior knowledge less and purely attempt to learn in-context instead.
For larger models, on the other hand, symbol tuning only results in performance changes within approximately $\pm 1$\%, indicating relatively-consistent performance before and after symbol tuning.
This consistent performance is expected, however, as symbol tuning is meant to improve a model's ability to learn from and reason with in-context exemplars, and models likely do not use in-context exemplars in order to succeed on these benchmarks.\footnote{Instruction-tuned models achieve similar performance in zero-shot settings versus few-shot settings on these benchmarks \citep{chung2022scaling}, suggesting that in-context exemplars are not crucial for completing these tasks.}

\subsection{Can symbol tuning improve chain-of-thought reasoning?}
\label{sec:appendix-can-symbol-tuning-improve-chain-of-thought-reasoning}
One limitation of symbol tuning is that it does not include any data with chain-of-thought (CoT) reasoning \citep{wei2022chain} since it is unclear how to best replace intermediate steps with symbols.
We thus want to examine whether symbol tuning affects chain-of-thought reasoning given its ability to improve in-context learning.
To analyze this, we reformat prompts from the two benchmarks in \cref{sec:appendix-does-symbol-tuning-affect-performance-on-benchmarks} to use chain-of-thought prompting and evaluate all symbol-tuned models.
We use the same chain-of-thought prompts that were used in \citet{chung2022scaling}.

\begin{figure}[t]
    \centering
    \pgfplotsset{
        width=0.45\linewidth, 
        height=0.45\linewidth,
        /pgfplots/ybar legend/.style={
            /pgfplots/legend image code/.code={
                \draw[##1,/tikz/.cd,yshift=-0.25em]
                (0cm,0cm) rectangle (7pt,0.8em);
            },
        },
    }
    \centering
    \begin{tikzpicture}
        \begin{groupplot}[
            group style={
            group name=plot,
            horizontal sep=40pt,
            vertical sep=20pt,
            group size=2 by 1},]
            \nextgroupplot[
                ybar=0pt,
                ymin=0, ymax=105,
                ytick={0, 10, 20, 30, 40, 50, 60, 70, 80, 90, 100},
                major x tick style = transparent,
                bar width=9 pt,
                enlarge x limits=0.25,
                typeset ticklabels with strut,
                ylabel={Accuracy (\%)},
                title={\textbf{MMLU (+CoT)}},
                symbolic x coords={8B, 62B, 62B-c, 540B},  
                xtick=data,  
                axis x line*=bottom,
                axis y line*=none,
                legend cell align=left,
                    legend style={
                            at={(1.1,-0.2)},
                            anchor=north,
                            column sep=1ex,
                            font=\small,
                            draw=none,
                            legend columns=3,
                    },
                ]  
                \addplot[ybar, fill=flancolorone,  postaction={}] coordinates {
                    (8B, 39.7)
                    (62B, 56.2)
                    (62B-c, 62.9)
                    (540B, 69.5)
                };  
                \addplot[ybar, fill=symflancolorone,  postaction={}] coordinates {
                    (8B, 39.6)
                    (62B, 57.4)
                    (62B-c, 62.6)
                    (540B, 68.4)
                };
                \legend{
                    Flan-PaLM \ \ \ \ \ \ \ \ \ \ ,
                    Flan-PaLM + Symbol tuning (ours)
                }
            \nextgroupplot[
                ybar=0pt,
                ymin=0, ymax=105,
                ytick={0, 10, 20, 30, 40, 50, 60, 70, 80, 90, 100},
                major x tick style = transparent,
                bar width=8 pt,
                enlarge x limits=0.25,
                typeset ticklabels with strut,
                title={\textbf{BIG-Bench Hard (+CoT)}},
                symbolic x coords={8B, 62B, 62B-c, 540B},  
                xtick=data,  
                axis x line*=bottom,
                axis y line*=none,
                legend cell align=left,
                    legend style={
                            at={(0.5,-0.2)},
                            anchor=north,
                            column sep=1ex,
                            font=\small,
                            draw=none,
                            legend columns=1,
                    },
                ]  
                \addplot[ybar, fill=flancolorone,  postaction={}] coordinates {
                    (8B, 30.5)
                    (62B, 44.9)
                    (62B-c, 54.4)
                    (540B, 65.8)
                };
                \addplot[ybar, fill=symflancolorone,  postaction={}] coordinates {
                    (8B, 17.7)
                    (62B, 44.9)
                    (62B-c, 53.4)
                    (540B, 67.3)
                };  
        \end{groupplot}
    \end{tikzpicture}
    \caption{
    Performance on MMLU and BIG-Bench Hard when using chain-of-thought prompting \citep{wei2022chain} does not significantly change after symbol tuning, though an outlier occurs where Flan-PaLM-8B experiences a significant decrease in performance on BIG-Bench Hard.
    Accuracy is shown as an unweighted average over all tasks for each benchmark (per-task results are shown in \cref{sec:appendix-mmlu} and \cref{sec:appendix-big-bench-hard}).
    }
    \label{fig:appendix-benchmark-cot-performance}
\end{figure}
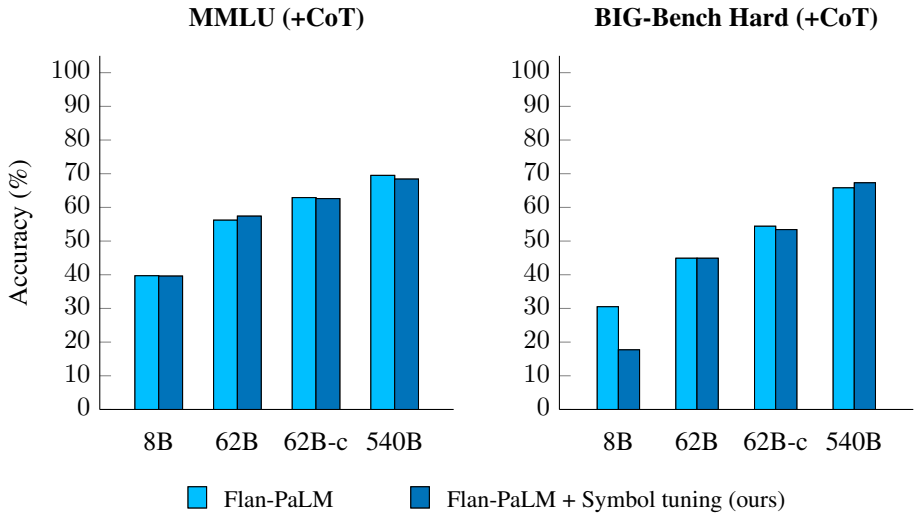

We show these results in \cref{fig:appendix-benchmark-cot-performance}.
We find that performance is mostly consistent between symbol-tuned models and their base variants when using CoT prompting.
One outlier, however, is that Flan-PaLM-8B experienced a significant drop in CoT performance on BIG-Bench Hard after symbol tuning, though it is unclear why this occurred since it did not experience a drop in CoT performance on MMLU.
Other than this outlier, the results are expected, as symbol tuning did not include any CoT prompts and thus should not change a model's performance in CoT settings.

\subsection{Does symbol tuning affect zero-shot performance?}
\label{sec:appendix-does-symbol-tuning-affect-zero-shot-performance}
Our setup for symbol tuning does not include any zero-shot examples, as an arbitrary symbol that maps an input to a label cannot be learned without any exemplars.
This raises the question of whether symbol tuning would harm a model's zero-shot performance, especially since we do not mix in any instruction-tuning data during symbol tuning for the reasons stated in \cref{sec:mixing-instruction-tuning-data}.
Intuitively, symbol tuning should not affect zero-shot performance because it should modify a model's ability to learn in-context and not its prior knowledge (which is what would primarily be used in zero-shot settings).
To test this, we test the models on the MMLU benchmark \citep{Hendrycks2021MMLU} and reformat prompts to a zero-shot setting.

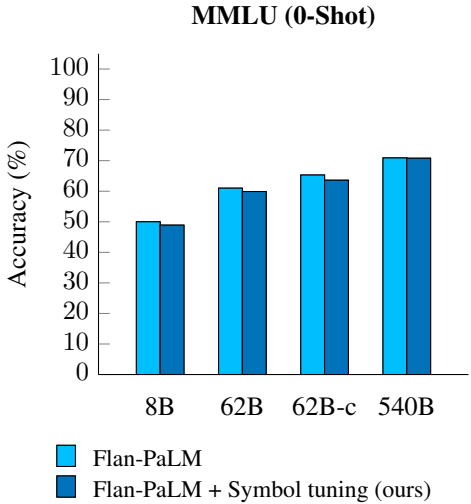
\begin{wrapfigure}{r}{6.5cm}
    \vspace{-8mm}
    \centering
    \pgfplotsset{
        width=\linewidth, 
        height=0.9\linewidth,
        /pgfplots/ybar legend/.style={
            /pgfplots/legend image code/.code={
                \draw[##1,/tikz/.cd,yshift=-0.25em]
                (0cm,0cm) rectangle (7pt,0.8em);
            },
        },
    }
    \centering
    \begin{tikzpicture}
        \begin{groupplot}[
            group style={
            group name=plot,
            horizontal sep=40pt,
            vertical sep=20pt,
            group size=2 by 1},]
            \nextgroupplot[
                ybar=0pt,
                ymin=0, ymax=105,
                ytick={0, 10, 20, 30, 40, 50, 60, 70, 80, 90, 100},
                major x tick style = transparent,
                bar width=9 pt,
                enlarge x limits=0.25,
                typeset ticklabels with strut,
                ylabel={Accuracy (\%)},
                title={\textbf{MMLU (0-Shot)}},
                symbolic x coords={8B, 62B, 62B-c, 540B},  
                xtick=data,  
                axis x line*=bottom,
                axis y line*=none,
                legend cell align=left,
                    legend style={
                            at={(0.4,-0.2)},
                            anchor=north,
                            column sep=1ex,
                            font=\small,
                            draw=none,
                            legend columns=1,
                    },
                ]  
                \addplot[ybar, fill=flancolorone,  postaction={}] coordinates {
                    (8B, 50.0)
                    (62B, 61.0)
                    (62B-c, 65.3)
                    (540B, 70.9)
                };  
                \addplot[ybar, fill=symflancolorone,  postaction={}] coordinates {
                    (8B, 48.9)
                    (62B, 59.9)
                    (62B-c, 63.6)
                    (540B, 70.8)
                };
                \legend{
                    Flan-PaLM,
                    Flan-PaLM + Symbol tuning (ours)
                }
        \end{groupplot}
    \end{tikzpicture}
    \caption{
    Performance on MMLU in a zero-shot setting does not significantly change after symbol tuning.
    Accuracy shown is an unweighted average over all tasks (per-task results are shown in \cref{sec:appendix-zero-shot-mmlu}).
    }
    \label{fig:appendix-benchmark-zero-shot-performance}
    \vspace{-3mm}
\end{wrapfigure}

In \cref{fig:appendix-benchmark-zero-shot-performance}, we compare each of our symbol-tuned model's performance on zero-shot MMLU against their respective Flan-PaLM model.
We find that performance is somewhat consistent after symbol-tuning.
Symbol-tuned models saw a maximum decrease in performance of 1.7\%, though we note that this difference is not sufficiently large to conclude that symbol tuning reduces zero-shot performance due to the variance within the evaluation.
For example, continuing instruction-tuning on Flan-PaLM-8B for 1k steps reduces MMLU 5-shot performance from 49.5\% to 47.2\%, and continuing for another 1k steps improve performance back to 49.0\%, which may indicate that for these benchmarks, small differences in performance are not enough to suggest an actual reduction or improvement in a model's true performance.
For this reason, we posit that the zero-shot performance before and after symbol-tuning is relatively-consistent for all base models, though we note that there is some ambiguity in this conclusion due to the variance in the performance metric.

\subsection{Do symbol-tuned models require fewer in-context exemplars?}
\label{sec:appendix-do-symbol-tuned-models-require-fewer-in-context-exemplars}
In \cref{sec:better-in-context-learning}, we showed that symbol-tuned models perform much better than Flan-PaLM models in difficult ICL settings without relevant labels.
Our evaluations, however, were all in a setting using four in-context exemplars per class, making it unclear how symbol-tuned models perform relative to baselines when there are fewer or more in-context exemplars that the model can use.
Intuitively, symbol tuning should be more effective when there are fewer in-context exemplars available, as having fewer exemplars makes it more difficult to identify the task (and we already showed in \cref{sec:better-in-context-learning} that symbol-tuned models are better in ICL settings where the task is unclear).

\begin{figure}[ht]
    \centering
    \begin{tikzpicture}
        \pgfplotsset{footnotesize,samples=10}
        \begin{groupplot}[
            group style = {group size = 4 by 1, horizontal sep = 20pt},
            width = 0.3\linewidth, 
            height = 0.3\linewidth]
            \nextgroupplot[
                align = center,
                title = {\textbf{Flan-PaLM-8B}}, 
                legend style={at={(-0.12,1.4)},anchor=south},
                xmode=log,
                xmin=0.7, xmax=23,
                ymin=-5, ymax=105,
                xtick={1, 2, 4, 8, 16},
                xticklabels={1, 2, 4, 8, 16},
                axis x line*=bottom,
                axis y line*=left,
                xlabel={\# exemplars per class},
                ylabel={Accuracy (\%)},
                ytick={0, 25, 50, 75, 100},
                yticklabels={0, 25, 50, 75, 100},
                grid style=dashed,
                x label style={at={(axis description cs:0.5,-0.15)},anchor=north},
                y label style={at={(axis description cs:-0.21,0.5)},anchor=south},
                xtick pos=bottom,
                ytick pos=left,
                legend cell align=left,
                    legend style={
                            at={(1.1,0.5)},
                            anchor=west,
                            column sep=1ex,
                            font=\small,
                            draw=none,
                    }
                ]
                \addplot[
                    color=flancolorone,
                    mark=\palmshape,
                    mark size=1.5pt,
                    line width=1pt,
                    ]
                    coordinates {
                    (1, 14.4)
                    (2, 42.5)
                    (4, 44.2)
                    (8, 47.8)
                    (16, 59.3)
                    };
                \addplot[
                    color=symflancolorone,
                    mark=\codexshape,
                    mark size=1.5pt,
                    line width=1pt,
                    ]
                    coordinates {
                    (1, 45.8)
                    (2, 48.8)
                    (4, 52.8)
                    (8, 58.7)
                    (16, 77.5)
                    };
            \nextgroupplot[
                align = center,
                title = {\textbf{Flan-PaLM-62B}}, 
                legend style={at={(-0.12,1.4)},anchor=south},
                xmode=log,
                xmin=0.7, xmax=23,
                ymin=-5, ymax=105,
                xtick={1, 2, 4, 8, 16},
                xticklabels={1, 2, 4, 8, 16},
                axis x line*=bottom,
                axis y line*=left,
                xlabel={\# exemplars per class},
                ytick={0, 25, 50, 75, 100},
                yticklabels={0, 25, 50, 75, 100},
                grid style=dashed,
                x label style={at={(axis description cs:0.5,-0.15)},anchor=north},
                y label style={at={(axis description cs:-0.16,0.5)},anchor=south},
                xtick pos=bottom,
                ytick pos=left,
                legend cell align=left,
                    legend style={
                            at={(-0.6,-0.6)},
                            anchor=west,
                            column sep=1ex,
                            font=\small,
                            draw=none,
                            legend columns=2,
                    }
                ]
                \addplot[
                    color=flancolorone,
                    mark=\palmshape,
                    mark size=1.5pt,
                    line width=1pt,
                    ]
                    coordinates {
                    (1, 11.5)
                    (2, 46.7)
                    (4, 50.5)
                    (8, 53.7)
                    (16, 75.0)
                    };
                    \addlegendentry{Flan-PaLM\ \ \ \ \ \ \ \ \ }
                \addplot[
                    color=symflancolorone,
                    mark=\codexshape,
                    mark size=1.5pt,
                    line width=1pt,
                    ]
                    coordinates {
                    (1, 49.4)
                    (2, 53.4)
                    (4, 60.3)
                    (8, 63.8)
                    (16, 82.8)
                    };
                    \addlegendentry{Flan-PaLM + Symbol tuning (ours)}
            \nextgroupplot[
                align = center,
                title = {\textbf{Flan-cont-PaLM-62B}}, 
                legend style={at={(-0.12,1.4)},anchor=south},
                xmode=log,
                xmin=0.7, xmax=23,
                ymin=-5, ymax=105,
                xtick={1, 2, 4, 8, 16},
                xticklabels={1, 2, 4, 8, 16},
                axis x line*=bottom,
                axis y line*=left,
                xlabel={\# exemplars per class},
                ytick={0, 25, 50, 75, 100},
                yticklabels={0, 25, 50, 75, 100},
                grid style=dashed,
                x label style={at={(axis description cs:0.5,-0.15)},anchor=north},
                y label style={at={(axis description cs:-0.16,0.5)},anchor=south},
                xtick pos=bottom,
                ytick pos=left,
                legend cell align=left,
                    legend style={
                            at={(-1.5,-0.5)},
                            anchor=west,
                            column sep=1ex,
                            font=\small,
                            draw=none,
                            legend columns=2,
                    }
                ]
                \addplot[
                    color=flancolorone,
                    mark=\palmshape,
                    mark size=1.5pt,
                    line width=1pt,
                    ]
                    coordinates {
                    (1, 8.6)
                    (2, 44.5)
                    (4, 51.0)
                    (8, 52.7)
                    (16, 74.0)
                    };
                \addplot[
                    color=symflancolorone,
                    mark=\codexshape,
                    mark size=1.5pt,
                    line width=1pt,
                    ]
                    coordinates {
                    (1, 49.0)
                    (2, 52.7)
                    (4, 62.1)
                    (8, 65.2)
                    (16, 85.5)
                    };
            \nextgroupplot[
                align = center,
                title = {\textbf{Flan-PaLM-540B}}, 
                legend style={at={(-0.12,1.4)},anchor=south},
                xmode=log,
                xmin=0.7, xmax=23,
                ymin=-5, ymax=105,
                xtick={1, 2, 4, 8, 16},
                xticklabels={1, 2, 4, 8, 16},
                axis x line*=bottom,
                axis y line*=left,
                xlabel={\# exemplars per class},
                ytick={0, 25, 50, 75, 100},
                yticklabels={0, 25, 50, 75, 100},
                grid style=dashed,
                x label style={at={(axis description cs:0.5,-0.15)},anchor=north},
                y label style={at={(axis description cs:-0.16,0.5)},anchor=south},
                xtick pos=bottom,
                ytick pos=left,
                legend cell align=left,
                    legend style={
                            at={(-1.5,-0.5)},
                            anchor=west,
                            column sep=1ex,
                            font=\small,
                            draw=none,
                            legend columns=2,
                    }
                ]
                \addplot[
                    color=flancolorone,
                    mark=\palmshape,
                    mark size=1.5pt,
                    line width=1pt,
                    ]
                    coordinates {
                    (1, 9.3)
                    (2, 49.3)
                    (4, 58.1)
                    (8, 60.3)
                    (16, 75.3)
                    };
                \addplot[
                    color=symflancolorone,
                    mark=\codexshape,
                    mark size=1.5pt,
                    line width=1pt,
                    ]
                    coordinates {
                    (1, 50.4)
                    (2, 56.6)
                    (4, 63.6)
                    (8, 67.7)
                    (16, 85.0)
                    };
        \end{groupplot}
    \end{tikzpicture}
    \caption{
    Symbol-tuned models consistently perform better than their respective Flan-PaLM models relative to the number of available in-context exemplars.
    The performance difference is especially significant when there is only one in-context exemplar per class available.
    Accuracy is shown as an unweighted average of the tasks with enough examples to use as in-context exemplars.
    }
    \label{fig:appendix-num-shots-ablation}
\end{figure}
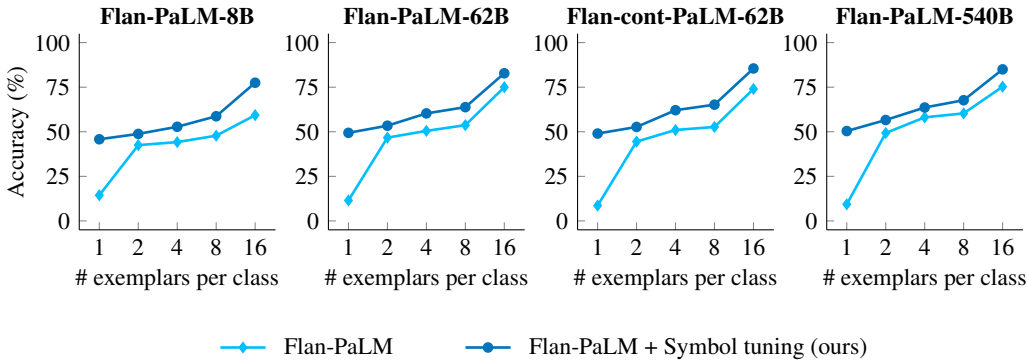

To investigate this, we regenerate evaluations using the same process as described in \cref{sec:evaluation-tasks}, except we vary the number of in-context exemplars per class.\footnote{If a dataset does not have enough examples to create a prompt with a particular number of in-context exemplars, we exclude that dataset from the evaluation for that number of in-context exemplars.}
We then test models on the hardest ICL setting from \cref{sec:better-in-context-learning} in order to study how instruction-tuned and symbol-tuned models behave relative to the number of available exemplars.
These results are shown in \cref{fig:appendix-num-shots-ablation}.
We find that the performance difference between symbol-tuned models and their base variants is relatively consistent in all settings except when there is only one in-context exemplar per class.
In this setting, symbol-tuned models perform much better than base models, and this trend is consistent across all of our tested models.
We posit that this could be a result of the Flan-PaLM not recognizing that arbitrary symbols are meant to be used as labels (which is implied because they perform significantly worse than random guessing), while symbol-tuned models already learned that arbitrary symbols can be used as labels.
These results suggest that in ICL settings where the task is unclear, symbol tuning improves model performance regardless of the number of in-context exemplars that are provided.

\subsection{Does symbol tuning require using all 30k labels?}
\label{sec:appendix-does-symbol-tuning-require-using-all-labels}
As described in \cref{sec:tuning-tasks}, our symbol-tuning procedure remapped original labels using a set of approximately 30k possible arbitrary symbols.
This raises the question, however, of whether symbol tuning requires this large of a label space, and exactly how large of a label space is necessary for successful symbol tuning.
Intuitively, we expect that models that are symbol tuned using larger label spaces should match or outperform those that are symbol tuned using smaller label spaces because a larger label space increases the diversity of the symbol-tuning data, which may make it easier to learn that \textit{any} arbitrary symbol can be used as a label.
We study how the size of the label space used for symbol tuning affects model performance by shrinking the label space for each category in \cref{sec:tuning-tasks}.
As our experiments from \cref{sec:tuning-tasks} use 10k possible labels per category, we decrease the label space size by only using 1k, 100, and 10 labels per category for possible labels.

\begin{figure}[th]
    \centering
    \begin{tikzpicture}
        \pgfplotsset{footnotesize,samples=10}
        \begin{groupplot}[
            group style = {group size = 4 by 1, horizontal sep = 20pt},
            width = 0.33\linewidth, 
            height = 0.33\linewidth]
            \nextgroupplot[
                align = center,
                title = {\textbf{Flan-PaLM-8B}}, 
                legend style={at={(-0.12,1.4)},anchor=south},
                xmode=log,
                xmin=0.7, xmax=11,
                ymin=45, ymax=85,
                xtick={1, 2, 4, 8},
                xticklabels={30, 300, 3k, 30k},
                xticklabel style = {font=\scriptsize},
                axis x line*=bottom,
                axis y line*=left,
                xlabel={\# Datasets},
                ylabel={Accuracy (\%)},
                ytick={50, 60, 70, 80},
                yticklabels={50, 60, 70, 80},
                grid style=dashed,
                x label style={at={(axis description cs:0.5,-0.15)},anchor=north},
                y label style={at={(axis description cs:-0.21,0.5)},anchor=south},
                xtick pos=bottom,
                ytick pos=left,
                legend cell align=left,
                    legend style={
                            at={(1.1,0.5)},
                            anchor=west,
                            column sep=1ex,
                            font=\small,
                            draw=none,
                    }
                ]
                \addplot[
                    color=instructcolorone,
                    mark=\codexshape,
                    mark size=1.5pt,
                    line width=1pt,
                    ]
                    coordinates {
                    (1, 51.6)
                    (2, 55.5)
                    (4, 59.6)
                    (8, 57.6)
                    };
                \addplot[
                    color=instructcolortwo,
                    mark=\basegptshape,
                    mark size=1.5pt,
                    line width=1pt,
                    ]
                    coordinates {
                    (1, 50.9)
                    (2, 54.8)
                    (4, 55.8)
                    (8, 54.3)
                    };
                \addplot[
                    color=instructcolorthree,
                    mark=\instructgptshape,
                    mark size=1.5pt,
                    line width=1pt,
                    ]
                    coordinates {
                    (1, 56.4)
                    (2, 57.5)
                    (4, 56.1)
                    (8, 58.2)
                    };
                \addplot[
                    color=instructcolorfour,
                    mark=\palmshape,
                    mark size=1.5pt,
                    line width=1pt,
                    ]
                    coordinates {
                    (1, 52.6)
                    (2, 54.3)
                    (4, 54.6)
                    (8, 52.8)
                    };
            \nextgroupplot[
                align = center,
                title = {\textbf{Flan-PaLM-62B}}, 
                legend style={at={(-0.12,1.4)},anchor=south},
                xmode=log,
                xmin=0.7, xmax=11,
                ymin=45, ymax=85,
                xtick={1, 2, 4, 8},
                xticklabels={30, 300, 3k, 30k},
                xticklabel style = {font=\scriptsize},
                axis x line*=bottom,
                axis y line*=left,
                xlabel={\# Datasets},
                ytick={50, 60, 70, 80},
                yticklabels={50, 60, 70, 80},
                grid style=dashed,
                x label style={at={(axis description cs:0.5,-0.15)},anchor=north},
                y label style={at={(axis description cs:-0.16,0.5)},anchor=south},
                xtick pos=bottom,
                ytick pos=left,
                legend cell align=left,
                    legend style={
                            at={(-1.4,-0.5)},
                            anchor=west,
                            column sep=1ex,
                            font=\small,
                            draw=none,
                            legend columns=2,
                    }
                ]
                \addplot[
                    color=instructcolorone,
                    mark=\codexshape,
                    mark size=1.5pt,
                    line width=1pt,
                    ]
                    coordinates {
                    (1, 72.4)
                    (2, 78.7)
                    (4, 75.4)
                    (8, 75.5)
                    };
                    \addlegendentry{Relevant Label + Instruction}
                \addplot[
                    color=instructcolortwo,
                    mark=\basegptshape,
                    mark size=1.5pt,
                    line width=1pt,
                    ]
                    coordinates {
                    (1, 65.9)
                    (2, 71.7)
                    (4, 66.8)
                    (8, 70.8)
                    };
                    \addlegendentry{Relevant Label + No Instruction}
                \addplot[
                    color=instructcolorthree,
                    mark=\instructgptshape,
                    mark size=1.5pt,
                    line width=1pt,
                    ]
                    coordinates {
                    (1, 70.6)
                    (2, 72.1)
                    (4, 71.5)
                    (8, 71.4)
                    };
                    \addlegendentry{No Relevant Label + Instruction \ \ \ \ \ \ \ }
                \addplot[
                    color=instructcolorfour,
                    mark=\palmshape,
                    mark size=1.5pt,
                    line width=1pt,
                    ]
                    coordinates {
                    (1, 62.0)
                    (2, 62.0)
                    (4, 59.7)
                    (8, 60.3)
                    };
                    \addlegendentry{No Relevant Label + No Instruction}
            \nextgroupplot[
                align = center,
                title = {\textbf{Flan-cont-PaLM-62B}}, 
                legend style={at={(-0.12,1.4)},anchor=south},
                xmode=log,
                xmin=0.7, xmax=11,
                ymin=45, ymax=85,
                xtick={1, 2, 4, 8},
                xticklabels={30, 300, 3k, 30k},
                xticklabel style = {font=\scriptsize},
                axis x line*=bottom,
                axis y line*=left,
                xlabel={\# Datasets},
                ytick={50, 60, 70, 80},
                yticklabels={50, 60, 70, 80},
                grid style=dashed,
                x label style={at={(axis description cs:0.5,-0.15)},anchor=north},
                y label style={at={(axis description cs:-0.16,0.5)},anchor=south},
                xtick pos=bottom,
                ytick pos=left,
                legend cell align=left,
                    legend style={
                            at={(-1.5,-0.5)},
                            anchor=west,
                            column sep=1ex,
                            font=\small,
                            draw=none,
                            legend columns=2,
                    }
                ]
                \addplot[
                    color=instructcolorone,
                    mark=\codexshape,
                    mark size=1.5pt,
                    line width=1pt,
                    ]
                    coordinates {
                    (1, 76.7)
                    (2, 77.4)
                    (4, 78.5)
                    (8, 78.9)
                    };
                \addplot[
                    color=instructcolortwo,
                    mark=\basegptshape,
                    mark size=1.5pt,
                    line width=1pt,
                    ]
                    coordinates {
                    (1, 70.3)
                    (2, 71.5)
                    (4, 73.6)
                    (8, 74.5)
                    };
                \addplot[
                    color=instructcolorthree,
                    mark=\instructgptshape,
                    mark size=1.5pt,
                    line width=1pt,
                    ]
                    coordinates {
                    (1, 71.5)
                    (2, 72.9)
                    (4, 72.8)
                    (8, 71.8)
                    };
                \addplot[
                    color=instructcolorfour,
                    mark=\palmshape,
                    mark size=1.5pt,
                    line width=1pt,
                    ]
                    coordinates {
                    (1, 61.6)
                    (2, 62.5)
                    (4, 63.5)
                    (8, 62.1)
                    };
        \end{groupplot}
    \end{tikzpicture}
    \caption{
    Symbol tuning using a larger label space slightly improves model performance, though the improvement is greater for the smallest model (Flan-PaLM-8B).
    All models are tuned for 4k steps.
    Performance is shown as the average accuracy across eleven datasets.
    }
    \label{fig:appendix-num-labels-ablation}
\end{figure}
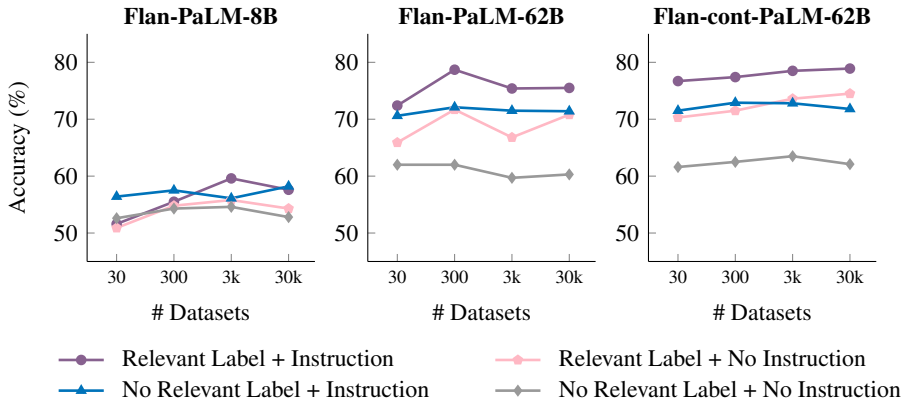

We retune models\footnote{We exclude Flan-PaLM-540B from this ablation study to reduce computational costs.} and evaluate their performance on the ICL settings from \cref{sec:better-in-context-learning}, showing these results in \cref{fig:appendix-num-labels-ablation}.
We find that, in general, models perform slightly better after symbol tuning using larger label spaces, but that the performance improvement from using larger label spaces is greater for the smallest model, Flan-PaLM-8B.
The improvement seen in Flan-PaLM-8B may suggest that the larger label space's ability to increase the diversity of the symbol-tuning data is important for smaller models that may have a harder time learning a general trend from a small sample size.
Combined with the overall trend of improved performance with larger label spaces across model sizes and across ICL settings, we posit that using a larger label space can indeed improve the symbol-tuned model performance to some degree, possibly because the larger label space creates a more-diverse set of prompts for the model to learn from.

\subsection{Which category of symbols is most important during symbol tuning?}
\label{sec:appendix-which-category-of-symbols-is-most-important}
For our symbol-tuning procedure, we used symbols drawn from three categories (integers, combinations of characters, and words).
Here, we investigate whether any particular category is more important for symbol tuning (one might expect, for example, using labels that are more similar to natural language might better teach models to examine in-context exemplars before using prior knowledge since models are more likely to have priors for those labels).
We retune models (we exclude Flan-PaLM-540B to reduce computational costs) using only integers, only character combinations, and only words as labels.
In \cref{tab:appendix-symbol-category-ablation}, we evaluate these models on the algorithmic reasoning tasks from \cref{sec:better-reasoning}, the flipped-label setting from \cref{sec:follow-flipped-labels}, and the ICL settings from \cref{sec:better-in-context-learning}.

We find that for all model sizes, using only words as labels results in the best performance on flipped labels, indicating that this category best teaches models to examine in-context exemplars before using prior knowledge.
Additionally, symbol tuning using words often yields the best performance when relevant labels are unavailable, but for Flan-PaLM-8B, yields the worst performance when relevant labels are available.
This may suggest that small models learn to treat all natural language labels as arbitrary symbols, even when the label is relevant and could be utilized to better learn the task.
Finally, while one might expect symbol tuning with numbers to be key to improving on algorithmic tasks, Flan-PaLM-8B and Flan-PaLM-62B actually perform better when tuned using only words (there is no consistently-better label type for Flan-cont-PaLM-62B).

\begin{table}[bh]
    \centering
    \scriptsize
    \arrayrulecolor{black}
    \setlength{\tabcolsep}{3pt}
    \resizebox{\columnwidth}{!}{
    \begin{tabular}{l cc c cccc}
        \toprule
            & \multicolumn{2}{c}{\thead{\footnotesize Algorithmic Reasoning}} & \multicolumn{5}{c}{\thead{\footnotesize In-Context Learning}} \\
        \cmidrule[0.5pt](lr){2-3}
        \cmidrule[0.5pt](lr){4-8}
            \multirow{2}{*}{\footnotesize Model} & \thead[c]{\scriptsize Turing \\ \scriptsize Concepts} & \thead[c]{\scriptsize List \\ \scriptsize Functions} & \thead[c]{\scriptsize Flipped \\ \scriptsize Labels} & \thead[c]{\scriptsize Relevant Target \\ \scriptsize + Instruction} & \thead[c]{\scriptsize Relevant Target \\ \scriptsize + No Instruction} & \thead[c]{\scriptsize No Relevant Target \\ \scriptsize + Instruction} & \thead[c]{\scriptsize No Relevant Target \\ \scriptsize + No Instruction} \\
        \midrule
            Random Guessing \hspace{17.5mm} & 0 & 0 & 50 & 42.4 & 42.4 & 42.4 & 42.4 \\
        \midrule
            \textbf{Flan-PaLM-8B} & 17.6 & 19.2 & 26.5 & 63.9 & 61.6 & 42.4 & 44.2 \\
            \ \ \ \ \ \ + Symbol tuning (integers) & 34.1 & 38.1 & 33.3 & \textbf{66.9} & \textbf{65.5} & 54.0 & 53.5 \\
            \ \ \ \ \ \ + Symbol tuning (characters) & 32.9 & 32.7 & 34.3 & 63.5 & 61.8 & 56.7 & 54.7 \\
            \ \ \ \ \ \ + Symbol tuning (words) & \textbf{52.9} & \textbf{42.5} & \textbf{54.8} & 60.6 & 56.6 & \textbf{56.9} & \textbf{54.9} \\
            \\
            \textbf{Flan-PaLM-62B} & 61.2 & 56.1 & 23.8 & 74.3 & 70.0 & 57.0 & 50.5 \\
            \ \ \ \ \ \ + Symbol tuning (integers) & 75.3 & 64.4 & 30.7 & 74.4 & 70.4 & 65.4 & 52.7 \\
            \ \ \ \ \ \ + Symbol tuning (characters) & 72.9 & 64.5 & 33.5 & 76.9 & 70.1 & 70.8 & 59.4 \\
            \ \ \ \ \ \ + Symbol tuning (words) & \textbf{78.8} & \textbf{68.9} & \textbf{54.2} & \textbf{77.3} & \textbf{73.4} & \textbf{71.4} & \textbf{60.7} \\
            \\
            \textbf{Flan-cont-PaLM-62B} & 64.7 & 54.7 & 27.3 & 77.3 & 70.3 & 56.3 & 51.0 \\
            \ \ \ \ \ \ + Symbol tuning (integers) & \textbf{77.6} & 68.1 & 32.5 & 78.2 & 71.0 & 67.7 & 58.9 \\
            \ \ \ \ \ \ + Symbol tuning (characters) & 74.1 & \textbf{69.4} & 33.5 & \textbf{78.3} & \textbf{72.1} & \textbf{73.7} & 60.6 \\
            \ \ \ \ \ \ + Symbol tuning (words) & 76.5 & 69.2 & \textbf{59.8} & \textbf{78.3} & 71.7 & 67.7 & \textbf{62.5} \\
        \bottomrule
    \end{tabular}
    }
    \caption{
    Model performance on algorithmic reasoning and in-context learning tasks when symbol-tuned using only integers, only character combinations, and only words as labels.
    }
    \label{tab:appendix-symbol-category-ablation}
\end{table}

\subsection{Can symbol tuning be successful using random labels?}
\label{sec:appendix-can-symbol-tuning-be-successful-using-random-labels}
As a sanity check, we want to show that symbol tuning cannot improve in-context learning when the tuning data is randomized.
We expect this behavior since if the input--label mappings are randomized, there is no task to learn from the in-context exemplars and thus no reason to learn to use exemplars.
To show this, we use the same symbol-tuning procedure as before but when remapping labels, we randomly select a symbol for each in-context exemplar rather than assigning a symbol for each label and consistently remapping all instances of that label to the new symbol.
This ensures that the labels (despite being arbitrary symbols) are randomized and that there is no meaningful task to learn.
We then retune models using symbol-tuning data generated using this modified process.\footnote{We exclude Flan-PaLM-540B from this ablation study to reduce computational costs.}

\begin{figure}[bh]
    \centering
    \pgfplotsset{
        width=0.45\linewidth, 
        height=0.35\linewidth,
        /pgfplots/ybar legend/.style={
            /pgfplots/legend image code/.code={
                \draw[##1,/tikz/.cd,yshift=-0.25em]
                (0cm,0cm) rectangle (7pt,0.8em);
            },
        },
    }
    \centering
    \begin{tikzpicture}
        \begin{groupplot}[
            group style={
            group name=plot,
            horizontal sep=30pt,
            vertical sep=30pt,
            group size=2 by 2},]
            \nextgroupplot[
                ybar=0pt,
                ymin=0, ymax=105,
                ytick={0, 20, 40, 60, 80, 100},
                major x tick style = transparent,
                bar width=9 pt,
                enlarge x limits=0.35,
                typeset ticklabels with strut,
                ylabel={Accuracy (\%)},
                title={\textbf{Relevant Label + Instruction}},
                title style = {text width=2.5cm,align=center},
                symbolic x coords={8B, 62B, 62B-c, 540B},
                xtick=data,  
                axis x line*=bottom,
                axis y line*=none,
                xticklabel=\empty,
                title style={yshift=-8pt},
                legend cell align=left,
                    legend style={
                            at={(1.05,-0.2)},
                            anchor=north,
                            column sep=1ex,
                            font=\small,
                            draw=none,
                            legend columns=2,
                    },
                ]  
                \addplot[ybar, fill=flancolorone,  postaction={}] coordinates {
                    (8B, 63.9)
                    (62B, 74.3)
                    (62B-c, 77.3)
                };  
                \addplot[ybar, fill=symflancolorone,  postaction={}] coordinates {
                    (8B, 57.6)
                    (62B, 75.5)
                    (62B-c, 78.9)
                };  
                \addplot[ybar, fill=frenchlilac,  postaction={}] coordinates {
                    (8B, 60.1)
                    (62B, 77.8)
                    (62B-c, 71.2)
                };  
            \nextgroupplot[
                ybar=0pt,
                ymin=0, ymax=105,
                ytick={0, 20, 40, 60, 80, 100},
                major x tick style = transparent,
                bar width=9 pt,
                enlarge x limits=0.35,
                typeset ticklabels with strut,
                title={\textbf{Relevant Label + No Instruction}},
                title style = {text width=2.5cm,align=center},
                symbolic x coords={8B, 62B, 62B-c, 540B},
                xtick=data,  
                axis x line*=bottom,
                axis y line*=none,
                xticklabel=\empty,
                title style={yshift=-8pt},
                legend cell align=left,
                    legend style={
                            at={(1.05,-0.2)},
                            anchor=north,
                            column sep=1ex,
                            font=\small,
                            draw=none,
                            legend columns=2,
                    },
                ]  
                \addplot[ybar, fill=flancolorone,  postaction={}] coordinates {
                    (8B, 61.6)
                    (62B, 70.0)
                    (62B-c, 70.3)
                };  
                \addplot[ybar, fill=symflancolorone,  postaction={}] coordinates {
                    (8B, 54.3)
                    (62B, 70.8)
                    (62B-c, 74.5)
                };  
                \addplot[ybar, fill=frenchlilac,  postaction={}] coordinates {
                    (8B, 59.1)
                    (62B, 70.3)
                    (62B-c, 57.0)
                };  
            \nextgroupplot[
                ybar=0pt,
                ymin=0, ymax=105,
                ytick={0, 20, 40, 60, 80, 100},
                major x tick style = transparent,
                bar width=9 pt,
                enlarge x limits=0.35,
                typeset ticklabels with strut,
                ylabel={Accuracy (\%)},
                title={\textbf{No Relevant Label + Instruction}},
                title style = {text width=3cm,align=center},
                symbolic x coords={8B, 62B, 62B-c, 540B},
                xtick=data,  
                axis x line*=bottom,
                axis y line*=none,
                title style={yshift=-8pt},
                legend cell align=left,
                    legend style={
                            at={(1.05,-0.2)},
                            anchor=north,
                            column sep=1ex,
                            font=\small,
                            draw=none,
                            legend columns=2,
                    },
                ]  
                \addplot[ybar, fill=flancolorone,  postaction={}] coordinates {
                    (8B, 42.4)
                    (62B, 57.0)
                    (62B-c, 56.3)
                };  
                \addplot[ybar, fill=symflancolorone,  postaction={}] coordinates {
                    (8B, 58.2)
                    (62B, 71.4)
                    (62B-c, 71.8)
                };  
                \addplot[ybar, fill=frenchlilac,  postaction={}] coordinates {
                    (8B, 45.0)
                    (62B, 53.7)
                    (62B-c, 47.8)
                };  
            \nextgroupplot[
                ybar=0pt,
                ymin=0, ymax=105,
                ytick={0, 20, 40, 60, 80, 100},
                major x tick style = transparent,
                bar width=9 pt,
                enlarge x limits=0.35,
                typeset ticklabels with strut,
                title={\textbf{No Relevant Label + No Instruction}},
                title style = {text width=3cm,align=center},
                symbolic x coords={8B, 62B, 62B-c, 540B},
                xtick=data,  
                axis x line*=bottom,
                axis y line*=none,
                title style={yshift=-8pt},
                legend cell align=left,
                    legend style={
                            at={(-0.1,-0.3)},
                            anchor=north,
                            column sep=1ex,
                            font=\small,
                            draw=none,
                            legend columns=1,
                    },
                ]  
                \addplot[ybar, fill=flancolorone,  postaction={}] coordinates {
                    (8B, 44.2)
                    (62B, 50.5)
                    (62B-c, 51.0)
                };  
                \addplot[ybar, fill=symflancolorone,  postaction={}] coordinates {
                    (8B, 52.8)
                    (62B, 60.3)
                    (62B-c, 62.1)
                };  
                \addplot[ybar, fill=frenchlilac,  postaction={}] coordinates {
                    (8B, 43.4)
                    (62B, 44.1)
                    (62B-c, 43.3)
                };  
                \legend{
                    Flan-PaLM \hspace{3mm},
                    Flan-PaLM + Symbol tuning (ours),
                    Flan-PaLM + Randomized symbol tuning
                }
        \end{groupplot}
    \end{tikzpicture}
    \caption{
    Models that are symbol tuned using randomized labels do not learn to better utilize in-context exemplars and often perform worse than standard symbol-tuned models, particularly when the model size is large or when relevant labels are not available.
    }
    \label{fig:appendix-randomized-labels}
\end{figure}
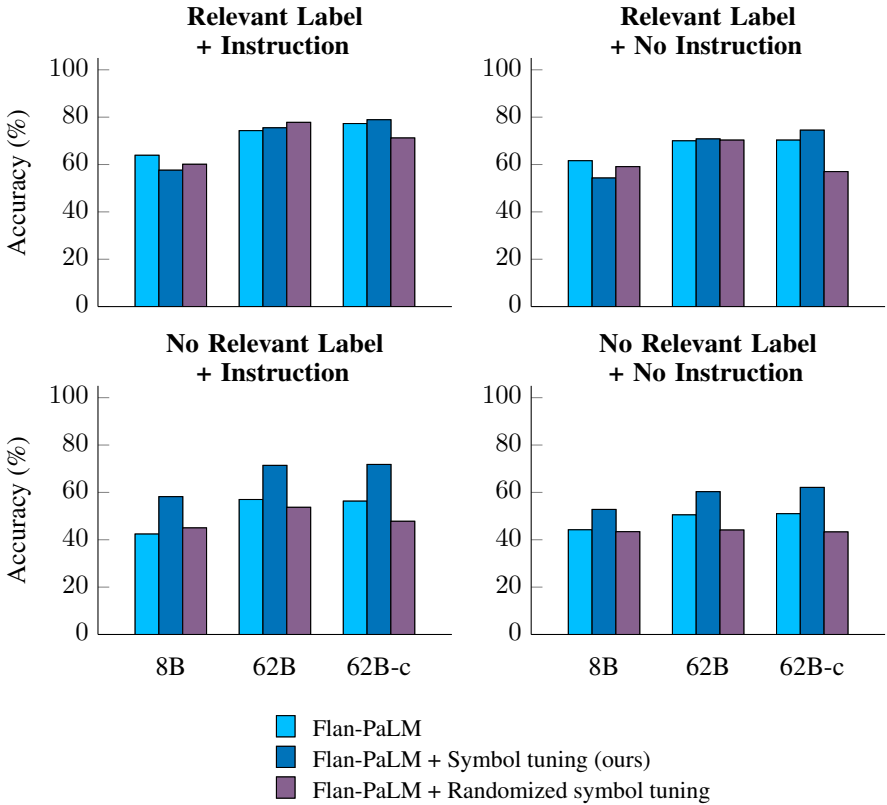

In \cref{fig:appendix-randomized-labels}, we show these models' performance on the ICL settings from \cref{sec:better-in-context-learning}.
We find that the randomized symbol-tuning procedure is almost always worse than the standard symbol-tuning procedure.
In settings without relevant targets, symbol tuning with randomized labels results in equal or worse performance compared with no symbol tuning at all, and model performance is strictly worse than that achieved by standard symbol tuning.
In settings with relevant targets, while randomized symbol tuning results in worse performance than no symbol tuning, it outperforms standard symbol tuning for Flan-PaLM-8B, our smallest model.
This result is not surprising, however, since in \cref{sec:better-in-context-learning}, we observed a large drop in model performance after symbol tuning for Flan-PaLM-8B in settings with relevant labels (which we posited resulted from the model treating all labels as arbitrary symbols, even when the label could have helped the model learn the task).
Overall, these results indicate that, as expected, models do not learn to better utilize in-context exemplars when symbol tuned using exemplars with randomized labels.

\clearpage
\section{Dataset Details}
\label{sec:appendix-dataset-details}
\subsection{Symbol-tuning datasets}
\label{sec:appendix-symbol-tuning-datasets}
Here, we show details of the tasks we used for symbol tuning as described in \cref{sec:tuning-tasks}.
We selected 22 publicly-available tasks from HuggingFace \citep{Lhoest2021Huggingface}, ensuring that each task has discrete labels so that there would be labels to swap with our symbols.
For each dataset, we used examples from the training split, and because some datasets had more examples than other datasets by multiple orders of magnitude, we cap the number of examples taken from any singular dataset at 25,000.
As shown in \cref{tab:appendix-tuning-datasets}, our tuning dataset consists of 291,693 total unique examples.

We selected datasets from several task types as follows:
natural language inference 
\citep[][\textbf{RTE}]{wang2019superglue},
\citep[][\textbf{WNLI}]{wang2018glue},
\citep[][\textbf{QNLI}]{rajpurkar-etal-2016-squad,wang2018glue},
\citep[][\textbf{MNLI}]{wang2018glue},
\citep[][\textbf{SNLI}]{Bowman2015Large}, and
\citep[][\textbf{CB}]{wang2019superglue};
sentiment analysis
\citep[][\textbf{SST2}]{socher-etal-2013-recursive},
\citep[][\textbf{RT}]{Pang2005Seeing}, and
\citep[][\textbf{TES}]{rosenthal2017semeval};
paraphrase detection
\citep[][\textbf{QQP}]{Chen2017QuoraQP,wang2018glue},
\citep[][\textbf{MRPC}]{wang2018glue}, and
\citep[][\textbf{PAWS}]{Zhang2019PAWS};
common sense answering
\citep[][\textbf{COPA}]{wang2019superglue} and 
\citep[][\textbf{PIQA}]{Bisk2020PIQA};
topic classification
\citep[][\textbf{AGN}]{Zhang2015Character} and
\citep[][\textbf{TREC}]{Li2002Learning};
coreference resolution
\citep[][\textbf{WSC}]{levesque2012winograd,wang2019superglue} and
\citep[][\textbf{WINO}]{Keisuke2019Winogrande};
offensive language identification
\citep[][\textbf{TEO}]{zampieri2019semeval};
irony detection
\citep[][\textbf{TEI}]{van2018semeval};
equal-meaning identification
\citep[][\textbf{WIC}]{wang2019superglue};
and sentence acceptability classification
\citep[][\textbf{COLA}]{wang2018glue}.

\begin{table}[bh]
    \centering
    \small
    \resizebox{\columnwidth}{!}{
    \begin{tabular}{l c ccc}
        \toprule
        \thead[l]{\small Task Type} \hspace{35mm} & Datasets & \# Classes & \# Available Examples & \# Examples Used \\
        \midrule
        \multirow{6}{*}{Natural Language Inference} &   RTE     &   2   &   2,488   &   2,488   \\
                                                    &   WNLI    &   2   &   635     &   635     \\
                                                    &   QNLI    &   2   &   104,743 &   25,000  \\
                                                    &   MNLI    &   3   &   392,577 &   25,000  \\
                                                    &   SNLI    &   3   &   549,526 &   25,000  \\
                                                    &   CB      &   3   &   250     &   250     \\
        \midrule
        \multirow{3}{*}{Sentiment Analysis}         &   SST2    &   2   &   66,978  &   25,000  \\
                                                    &   RT      &   2   &   8,530   &   8,530   \\
                                                    &   TES     &   3   &   45,586  &   25,000  \\
        \midrule
        \multirow{3}{*}{Paraphrase Detection}       &   QQP     &   2   &   363,846 &   25,000  \\
                                                    &   MRPC    &   2   &   3,668   &   3,668   \\
                                                    &   PAWS    &   2   &   49,349  &   25,000  \\
        \midrule
        \multirow{2}{*}{Common Sense}               &   COPA    &   2   &   400     &   400     \\
                                                    &   PIQA    &   2   &   16,107  &   16,107  \\
        \midrule
        \multirow{2}{*}{Topic Classification}       &   AGN     &   4   &   120,000 &   25,000  \\
                                                    &   TREC    &   6   &   5,381   &   5,381   \\
        \midrule
        \multirow{2}{*}{Coreference}                &   WSC     &   2   &   529     &   529     \\
                                                    &   WINO    &   2   &   40,394  &   25,000  \\
        \midrule
        \multirow{4}{*}{Miscellaneous}              &   TEO     &   2   &   11,883  &   11,883  \\
                                                    &   TEI     &   2   &   2,862   &   2,862   \\
                                                    &   WIC     &   2   &   5,428   &   5,428   \\
                                                    &   COLA    &   2   &   8,532   &   8,532   \\
        \midrule
        \textbf{Total} & --- & -- & 1,799,692 & 291,693 \\
        \bottomrule
    \end{tabular}
    }
    \caption{
    Tuning tasks used in this paper.
    }
    \label{tab:appendix-tuning-datasets}
\end{table}

\subsection{Evaluation datasets}
\label{sec:appendix-evaluation-datasets}
In this section, we list the eleven tasks from \cref{sec:evaluation-tasks} that we used for our evaluation.
We selected eleven publicly-available tasks from HuggingFace \citep{Lhoest2021Huggingface}.
In order to ensure that evaluation tasks were not seen during tuning, we select datasets that were not used in symbol tuning (\cref{sec:appendix-symbol-tuning-datasets}) and not used in instruction tuning (specifically, the datasets used in \citet{chung2022scaling}, \citet{wei2021finetuned}, and \citet{sanh2022multitask}).
For each dataset, we select examples from the validation split when available (we use the train split if there is no validation split).
Some evaluation tasks had significantly more available examples than other evaluation tasks, so we cap the number of examples per evaluation task at 100 in order to make evaluation set sizes similar and reduce the computational costs of each evaluation.

As shown in \cref{tab:appendix-eval-datasets}, we use the following tasks:
subjectivity detection \citep[][\textbf{SUBJ}]{conneau2018senteval},
hate speech detection \citep[][\textbf{TEH}]{basile2019semeval},
abortion stance classification \citep[][\textbf{TEAB}]{mohammad2016semeval},
atheism stance classification \citep[][\textbf{TEAT}]{mohammad2016semeval},
feminism stance classification \citep[][\textbf{TEFE}]{mohammad2016semeval},
Hillary Clinton stance classification \citep[][\textbf{TEHI}]{mohammad2016semeval},
adverse drug event classification \citep[][\textbf{ADEC}]{Alex2021RAFT},
overruling classification \citep[][\textbf{OR}]{Alex2021RAFT},
organization classification \citep[][\textbf{SOT}]{Alex2021RAFT},
potentially-unfair terms-of-service detection \citep[][\textbf{TOS}]{Alex2021RAFT},
and
Twitter complaint detection \citep[][\textbf{TC}]{Alex2021RAFT}.
In \cref{tab:appendix-evaluation-task-instructions}, we also show the instructions that we provided for each dataset when instructions are included in the prompt setting.

\begin{table}[hb]
    \centering
    \small
    \renewcommand{\arraystretch}{1.25}
    \resizebox{\columnwidth}{!}{
    \begin{tabular}{l ccc}
        \toprule
        Dataset Name (Abbreviation) & \# Classes & \# Available Examples & \# Examples Used \\
        \midrule
        Subjectivity detection (SUBJ) & 2 & 2,000 & 100 \\
        Hate speech detection (TEH) & 2 & 1,000 & 100 \\
        Abortion stance classification (TEAB) & 3 & 66 & 66 \\
        Atheism stance classification (TEAT) & 3 & 52 & 52 \\
        Feminism stance classification (TEFE) & 3 & 67 & 67 \\
        Hillary Clinton stance classification (TEHI) & 3 & 69 & 69 \\
        Adverse drug event classification (ADEC) & 2 & 50 & 50 \\
        Overruling detection (OR) & 2 & 50 & 50 \\
        Organization classification (SOT) & 3 & 50 & 50 \\
        Unfair terms of service detection (TOS) & 2 & 50 & 50 \\
        Twitter complaint detection (TC) & 2 & 50 & 50 \\
        \midrule
        \textbf{Total} & -- & 3,504 & 704 \\
        \bottomrule
    \end{tabular}
    }
    \caption{
    Evaluation tasks used in this paper.
    }
    \label{tab:appendix-eval-datasets}
\end{table}

\begin{table}[hb]
    \centering
    \small
    \renewcommand{\arraystretch}{1.5}
    \resizebox{\columnwidth}{!}{
    \begin{tabular}{l  l}
        \toprule
        \thead[c]{\normalsize Dataset\ \ \ \ \ \ \ } & \multicolumn{1}{c}{\normalsize Instruction} \\
        \midrule
        SUBJ    & ``Is the following sentence subjective or objective?'' \\
        TEH     & ``Label the following tweet based on whether it contains hate speech.'' \\
        TEAB    & ``Read the following tweet and determine its stance on abortion.'' \\
        TEAT    & ``Read the following tweet and determine its stance on atheism.'' \\
        TEFE    & ``Read the following tweet and determine its stance on feminism.'' \\
        TEHI    & ``Read the following tweet and determine its stance on Hillary Clinton.'' \\
        ADEC    & ``Label the following sentence based on whether it is related to an adverse drug event.'' \\
        OR      & ``Label the following sentence based on whether it is overruling or not.'' \\
        SOT     & \thead[l]{\small ``Read the following paper title and institution name and classify the institution as \\ \small a university, company, or research institute.''} \\
        TOS     & \thead[l]{\small ``Label the following sentence from a Terms of Service based on whether it is \\ \small potentially unfair.''} \\
        TC      & ``Label the following tweet text based on whether it contains a complaint.'' \\
        \bottomrule
    \end{tabular}
    }
    \caption{
        Instructions used for each evaluation dataset.
    }
    \label{tab:appendix-evaluation-task-instructions}
\end{table}

\clearpage
\section{Symbol tuning details}
\label{sec:appendix-symbol-tuning-details}
\subsection{Symbol selection}
\label{sec:appendix-symbol-selection}
In this paper, we experimented using a set of $\sim$300k arbitrary symbols as shown in \cref{fig:symbols-used}.
When selecting a symbol to replace natural language labels with, we first randomly select a type of symbol from the three categories (integers, combinations of characters\footnote{Obtained by converting integers to characters (e.g., $0 \to A$, $1 \to B$, $26 \to AA$, etc.).}, and words\footnote{Obtained from MIT's list of 10k words (\url{www.mit.edu/~ecprice/wordlist.10000}) and list of 100k words (\url{www.mit.edu/~ecprice/wordlist.100000}).}) and then select a random symbol from the available symbols for that category.
We did not test other ways of generating arbitrary symbols (e.g., picking random words from the prompt, combining multiple words, combining alphabetical characters and numbers, etc.) and leave this for future work.

\subsection{Prompt formatting}
\label{sec:appendix-prompt-formatting}
We used ten distinct prompt templates to format inputs and outputs into prompts.
During both tuning and evaluation, prompts are randomly generated using one of the following templates ([input] and [label] stand for the input and label of a given example, respectively):

\begin{itemize}[noitemsep,leftmargin=*]
    \item ``Input: [input] \textbackslash n Output: [label]''
    \item ``Input: [input] \textbackslash n Target: [label]''
    \item ``Input: [input] \textbackslash n Symbol: [label]''
    \item ``Input: [input] \textbackslash n Label: [label]''
    \item ``Question: [input] \textbackslash n Answer: [label]''
    \item ``Student: [input] \textbackslash n Teacher: [label]''
    \item ``X = [input] \textbackslash n Y = [label]''
    \item ``Q: [input] \textbackslash n A: [label]''
    \item ``[input] -> [label]''
    \item ``Sentences: [input] \textbackslash n Mapped To: [label]''
\end{itemize}

For evaluation prompts with instructions, however, we format the prompt as ``Question: [instruction] \textbackslash n [input] \textbackslash n Answer: [label]'' where [instruction] stands for the instruction for a given task (see \cref{tab:appendix-evaluation-task-instructions} for instructions that we used).
\cref{sec:appendix-evaluation-prompts} contains examples of prompts that were generated using these prompt templates with instructions.

\subsection{Tuning procedure}
\label{sec:appendix-tuning-procedure}
In \cref{tab:appendix-model-finetuning-details}, we show tuning details for each model that we symbol tuned.
We primarily follow the hyperparameter selection from \citet{chung2022scaling}---in particular, we use the same batch size, dropout, and learning rate for each model.
On the other hand, we showed in \cref{sec:number-of-tuning-steps} that symbol tuning does not require tuning for as long as instruction tuning does.
Because we use packing \citep{Raffel2020Exploring}, the effective batch size is larger than the reported number.

\begin{table}[hb]
    \centering
    \renewcommand{\arraystretch}{1.4}
    \resizebox{0.8\columnwidth}{!}{
    \begin{tabular}{l l c c c c}
         \toprule
         Params & Model & Batch size & Dropout & LR & Steps \\
         \midrule
         8B     &   Flan-PaLM   &   32  &   0.05    &   $3 \times 10^{-3}$  &   4k   \\
         62B    &   Flan-PaLM   &   32  &   0.05    &   $3 \times 10^{-3}$  &   4k   \\
         540B   &   Flan-PaLM   &   32  &   0.1     &   $1 \times 10^{-3}$  &   1k   \\
         \\
         62B    &   Flan-cont-PaLM  &   32  &   0.05    &    $3 \times 10^{-3}$ & 4k \\
         \bottomrule
    \end{tabular}
    }
    \caption{
    Hyperparameters for all symbol-tuned models.
    }
    \label{tab:appendix-model-finetuning-details}
\end{table}

\clearpage
\section{Full experimental results}
\subsection{BIG-Bench list functions}
\label{sec:appendix-big-bench-list-functions}
We experimented on twenty list function tasks from the List Functions benchmark from BIG-Bench \citep{bigbench}.
These list function tasks were selected as the tasks with the highest human accuracy baseline reported in \citet{rule2020thesis}.
We describe each of the tasks that we tested in \cref{fig:algorithmic-reasoning-main-comparison} and categorize them into five distinct categories based on the list function used by that task.

The pairings in all tasks are composed of input and output lists that contain numbers from 0 to 9 or numbers from 0 to 99 (these two ranges are separated such that a single list function can have two associated tasks, one for each range).
Each task contains 32 input--output pairs---each pairing is used as an evaluation example and for each evaluation example, in-context exemplars examples are randomly selected from the remaining 31 pairs.
In \cref{sec:better-in-context-learning}, we evaluated models on evaluation examples generated with four in-context exemplars.
We show per-task results from this experiment for base models, continued instruction-tuned variants, and symbol-tuned variants in \cref{tab:list-functions-per-task-1}.

\begin{figure}[hb]
    \centering
    \includegraphics[width=\linewidth]{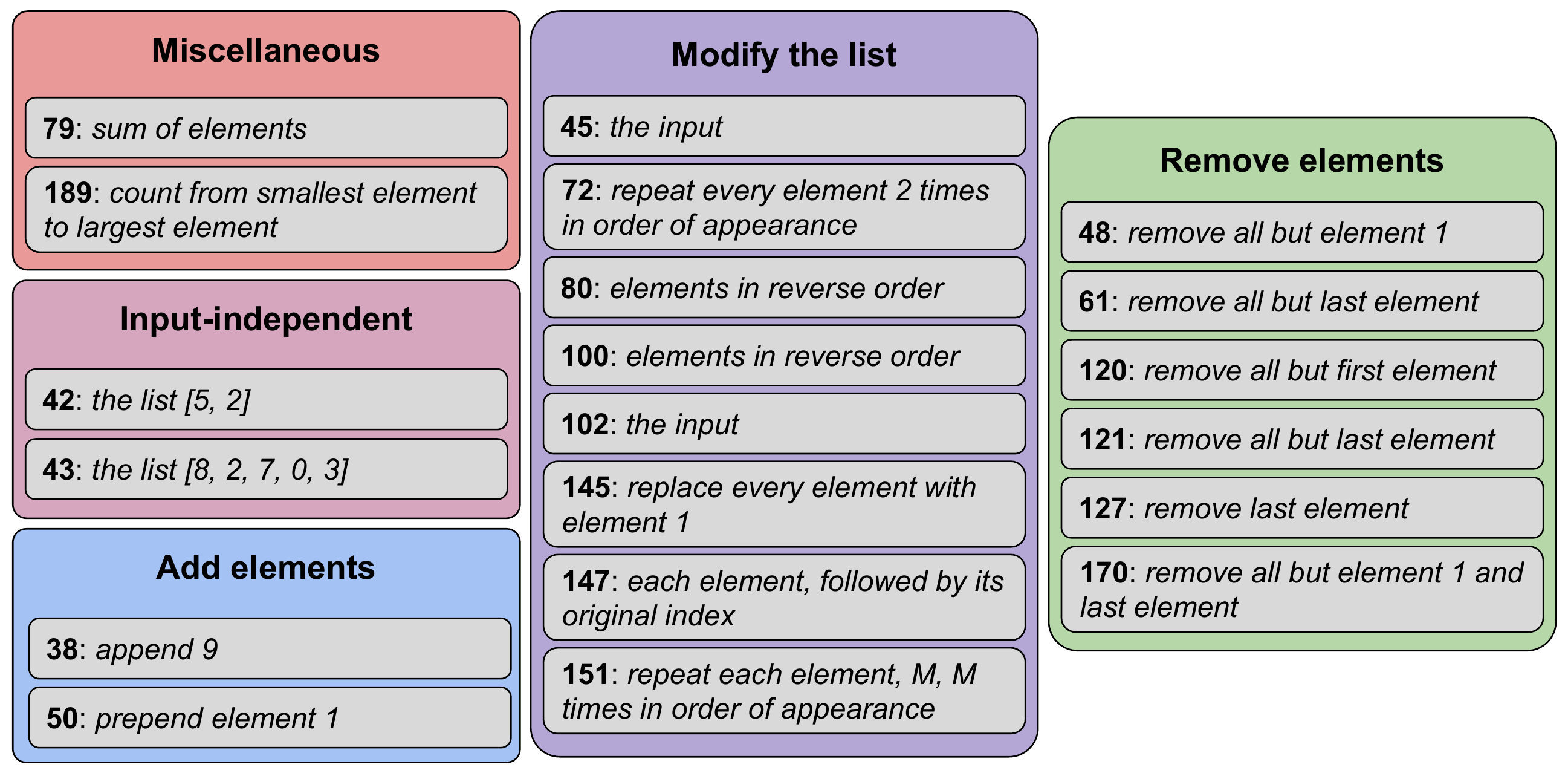}
    \caption{
    The twenty list function tasks used in \cref{sec:better-reasoning} grouped by each task categories used in \cref{fig:algorithmic-reasoning-main-comparison}.
    Task numbers for reference are bolded.
    Descriptions of each task are italicized---some task descriptions are identical because one variant uses only numbers from 0 to 9 while the other variant uses numbers from 0 to 99 (following the setup from \citet{bigbench}).
    }
    \label{fig:appendix-list-functions-tasks}
\end{figure}

\begin{table}[hb]
    \vspace{4mm}
    \centering
    \scriptsize
    \arrayrulecolor{black}
    \setlength{\tabcolsep}{2.5pt}
    \resizebox{\columnwidth}{!}{
    \begin{tabular}{ll ccccccccccccccccccccc}
        \toprule
        \multicolumn{2}{l}{\multirow{2}{*}{\thead{\scriptsize \\ Model}}} & & \multicolumn{19}{c}{\thead{Task Number}} \\
        \cmidrule{3-23}
         & & 38 & 42 & 43 & 45 & 48 & 50 & 61 & 72 & 79 & 80 & 100 & 102 & 120 & 121 & 127 & 145 & 147 & 151 & 170 & 189 & \textbf{Avg.} \\
        \midrule
        8B & Flan-PaLM & 0.0 & 18.8 & 9.4 & 96.9 & 9.4 & 0.0 & 9.4 & 12.5 & 18.8 & 6.2 & 15.6 & 93.8 & 15.6 & 31.2 & 0.0 & 15.6 & 12.5 & 15.6 & 3.1 & 0.0 & 19.2 \\
        & + Instruction tuning & 0.0 & 25.0 & 9.4 & 100.0 & 28.1 & 0.0 & 21.9 & 3.1 & 18.8 & 9.4 & 15.6 & 96.9 & 28.1 & 53.1 & 0.0 & 21.9 & 12.5 & 12.5 & 6.2 & 0.0 & 23.1 \\
        & + Symbol tuning & 9.4 & 87.5 & 75.0 & 96.9 & 62.5 & 3.1 & 37.5 & 9.4 & 34.4 & 12.5 & 12.5 & 100.0 & 59.4 & 71.9 & 12.5 & 18.8 & 18.8 & 18.8 & 6.2 & 0.0 & 37.4 \\
        \\
        62B & Flan-PaLM & 81.2 & 90.6 & 84.4 & 100.0 & 75.0 & 12.5 & 59.4 & 43.8 & 65.6 & 43.8 & 34.4 & 100.0 & 62.5 & 81.2 & 21.9 & 75.0 & 34.4 & 15.6 & 25.0 & 15.6 & 56.1 \\
        & + Instruction tuning & 62.5 & 96.9 & 90.6 & 100.0 & 68.8 & 21.9 & 53.1 & 40.6 & 71.9 & 46.9 & 37.5 & 100.0 & 65.6 & 68.8 & 40.6 & 71.9 & 34.4 & 15.6 & 15.6 & 21.9 & 56.3 \\
        & + Symbol tuning & 96.9 & 96.9 & 100.0 & 100.0 & 96.9 & 46.9 & 75.0 & 68.8 & 78.1 & 56.2 & 46.9 & 100.0 & 93.8 & 84.4 & 21.9 & 90.6 & 46.9 & 12.5 & 15.6 & 15.6 & 67.2 \\
        \\
        62B & Flan-cont-PaLM & 56.2 & 87.5 & 71.9 & 96.9 & 62.5 & 12.5 & 68.8 & 50.0 & 53.1 & 59.4 & 46.9 & 100.0 & 75.0 & 75.0 & 31.2 & 62.5 & 40.6 & 9.4 & 18.8 & 15.6 & 54.7 \\
        & + Instruction tuning & 75.0 & 93.8 & 90.6 & 100.0 & 90.6 & 9.4 & 81.2 & 71.9 & 65.6 & 62.5 & 46.9 & 100.0 & 90.6 & 78.1 & 50.0 & 65.6 & 53.1 & 15.6 & 28.1 & 31.2 & 65.0 \\
        & + Symbol tuning & 93.8 & 100.0 & 96.9 & 100.0 & 100.0 & 31.2 & 81.2 & 90.6 & 59.4 & 71.9 & 50.0 & 100.0 & 93.8 & 87.5 & 28.1 & 84.4 & 53.1 & 12.5 & 40.6 & 28.1 & 70.2 \\
        \\
        540B & Flan-PaLM & 90.6 & 81.2 & 100.0 & 100.0 & 46.9 & 81.2 & 50.0 & 96.9 & 59.4 & 65.6 & 50.0 & 100.0 & 78.1 & 46.9 & 78.1 & 96.9 & 84.4 & 18.8 & 18.8 & 46.9 & 69.5 \\
        & + Instruction tuning & 93.8 & 75.0 & 90.6 & 100.0 & 46.9 & 84.4 & 37.5 & 96.9 & 62.5 & 59.4 & 46.9 & 100.0 & 78.1 & 46.9 & 78.1 & 100.0 & 93.8 & 18.8 & 21.9 & 46.9 & 68.9 \\
        & + Symbol tuning & 93.8 & 100.0 & 100.0 & 100.0 & 68.8 & 81.2 & 56.2 & 100.0 & 71.9 & 75.0 & 56.2 & 100.0 & 65.6 & 50.0 & 81.2 & 93.8 & 87.5 & 18.8 & 18.8 & 43.8 & 73.1 \\
        \bottomrule
    \end{tabular}
    }
    \caption{
    List functions individual task performance.
    }
    \label{tab:list-functions-per-task-1}
\end{table}

\clearpage
\subsection{In-context learning}
\label{sec:appendix-icl-settings}
We evaluated each model's in-context learning abilities on a set of eleven datasets as described in \cref{sec:evaluation-tasks}.
We reported results on these tasks using an unweighted average of the per-task accuracies.
In \cref{tab:icl-settings-per-task-1}, \cref{tab:icl-settings-per-task-2}, \cref{tab:icl-settings-per-task-3}, and \cref{tab:icl-settings-per-task-4}, we show base model, continued instruction-tuned model, and symbol-tuned model performance for each task.
Models have been tuned with the same specifications described in \cref{sec:appendix-tuning-procedure}.

\begin{table}[hb]
    \vspace{4mm}
    \caption{
    ADEC, OR, and SOT 4-shot task performance.
    }
    \label{tab:icl-settings-per-task-1}
    \centering
    \scriptsize
    \arrayrulecolor{black}
    \resizebox{\columnwidth}{!}{
    \begin{tabular}{l cccc cccc cccc}
        \toprule
        & \multicolumn{4}{c}{\thead{\scriptsize \textbf{ADEC}}} & \multicolumn{4}{c}{\thead{\scriptsize \textbf{OR}}} & \multicolumn{4}{c}{\thead{\scriptsize \textbf{SOT}}}\\
        \cmidrule[0.5pt](lr){2-5}
        \cmidrule[0.5pt](lr){6-9}
        \cmidrule[0.5pt](lr){10-13}
        \hspace{9mm} \textbf{Relevant labels:} & \colorbox{babyblue}{\textbf{\cmark}} & \colorbox{babyblue}{\textbf{\cmark}} & \colorbox{fluorescentorange}{\textbf{\xmark}} & \colorbox{fluorescentorange}{\textbf{\xmark}} & \colorbox{babyblue}{\textbf{\cmark}} & \colorbox{babyblue}{\textbf{\cmark}} & \colorbox{fluorescentorange}{\textbf{\xmark}} & \colorbox{fluorescentorange}{\textbf{\xmark}} & \colorbox{babyblue}{\textbf{\cmark}} & \colorbox{babyblue}{\textbf{\cmark}} & \colorbox{fluorescentorange}{\textbf{\xmark}} & \colorbox{fluorescentorange}{\textbf{\xmark}} \\
        \hspace{7mm} \textbf{Task instructions:} & \colorbox{babyblue}{\textbf{\cmark}} & \colorbox{fluorescentorange}{\textbf{\xmark}} & \colorbox{babyblue}{\textbf{\cmark}} & \colorbox{fluorescentorange}{\textbf{\xmark}} & \colorbox{babyblue}{\textbf{\cmark}} & \colorbox{fluorescentorange}{\textbf{\xmark}} & \colorbox{babyblue}{\textbf{\cmark}} & \colorbox{fluorescentorange}{\textbf{\xmark}} & \colorbox{babyblue}{\textbf{\cmark}} & \colorbox{fluorescentorange}{\textbf{\xmark}} & \colorbox{babyblue}{\textbf{\cmark}} & \colorbox{fluorescentorange}{\textbf{\xmark}} \\
        \midrule
        Random Guessing \hspace{10mm} & 50 & 50 & 50 & 50 & 50 & 50 & 50 & 50 & 33.3 & 33.3 & 33.3 & 33.3 \\
        \midrule
        Flan-PaLM-8B & 86 & 80 & 60 & 48 & 96 & 86 & 62 & 62 & 80 & 84 & 12 & 34 \\
        \ \ \ \ \ \ + Instruction tuning & 82 & 82 & 50 & 54 & 96 & 96 & 58 & 70 & 82 & 86 & 20 & 32 \\
        \ \ \ \ \ \ + Symbol tuning (ours) & 76 & 62 & 74 & 48 & 82 & 90 & 80 & 82 & 78 & 74 & 46 & 40 \\
        \\
        Flan-PaLM-62B & 56 & 78 & 70 & 56 & 96 & 92 & 76 & 74 & 88 & 86 & 30 & 48 \\
        \ \ \ \ \ \ + Instruction tuning & 70 & 78 & 76 & 56 & 92 & 92 & 72 & 72 & 90 & 88 & 50 & 50 \\
        \ \ \ \ \ \ + Symbol tuning (ours) & 82 & 88 & 90 & 66 & 98 & 98 & 98 & 90 & 78 & 88 & 76 & 36 \\
        \\
        Flan-cont-PaLM-62B & 70 & 74 & 70 & 50 & 96 & 86 & 80 & 70 & 96 & 96 & 52 & 42 \\
        \ \ \ \ \ \ + Instruction tuning & 80 & 84 & 80 & 52 & 96 & 94 & 88 & 72 & 94 & 94 & 56 & 46 \\
        \ \ \ \ \ \ + Symbol tuning (ours) & 90 & 84 & 86 & 58 & 98 & 98 & 98 & 96 & 96 & 90 & 84 & 40 \\
        \\
        Flan-PaLM-540B & 90 & 84 & 66 & 54 & 98 & 94 & 98 & 66 & 96 & 86 & 62 & 42 \\
        \ \ \ \ \ \ + Instruction tuning & 86 & 82 & 68 & 62 & 98 & 94 & 98 & 66 & 96 & 86 & 70 & 50 \\
        \ \ \ \ \ \ + Symbol tuning (ours) & 90 & 88 & 88 & 64 & 98 & 98 & 98 & 94 & 94 & 86 & 90 & 50 \\
        \bottomrule
    \end{tabular}
    }

    \vspace{8mm}
    \caption{
    SUBJ, TC, and TEAB 4-shot task performance.
    }
    \label{tab:icl-settings-per-task-2}
    \centering
    \scriptsize
    \arrayrulecolor{black}
    \resizebox{\columnwidth}{!}{
    \begin{tabular}{l cccc cccc cccc}
        \toprule
        & \multicolumn{4}{c}{\thead{\scriptsize \textbf{SUBJ}}} & \multicolumn{4}{c}{\thead{\scriptsize \textbf{TC}}} & \multicolumn{4}{c}{\thead{\scriptsize \textbf{TEAB}}}\\
        \cmidrule[0.5pt](lr){2-5}
        \cmidrule[0.5pt](lr){6-9}
        \cmidrule[0.5pt](lr){10-13}
        \hspace{9mm} \textbf{Relevant labels:} & \colorbox{babyblue}{\textbf{\cmark}} & \colorbox{babyblue}{\textbf{\cmark}} & \colorbox{fluorescentorange}{\textbf{\xmark}} & \colorbox{fluorescentorange}{\textbf{\xmark}} & \colorbox{babyblue}{\textbf{\cmark}} & \colorbox{babyblue}{\textbf{\cmark}} & \colorbox{fluorescentorange}{\textbf{\xmark}} & \colorbox{fluorescentorange}{\textbf{\xmark}} & \colorbox{babyblue}{\textbf{\cmark}} & \colorbox{babyblue}{\textbf{\cmark}} & \colorbox{fluorescentorange}{\textbf{\xmark}} & \colorbox{fluorescentorange}{\textbf{\xmark}} \\
        \hspace{7mm} \textbf{Task instructions:} & \colorbox{babyblue}{\textbf{\cmark}} & \colorbox{fluorescentorange}{\textbf{\xmark}} & \colorbox{babyblue}{\textbf{\cmark}} & \colorbox{fluorescentorange}{\textbf{\xmark}} & \colorbox{babyblue}{\textbf{\cmark}} & \colorbox{fluorescentorange}{\textbf{\xmark}} & \colorbox{babyblue}{\textbf{\cmark}} & \colorbox{fluorescentorange}{\textbf{\xmark}} & \colorbox{babyblue}{\textbf{\cmark}} & \colorbox{fluorescentorange}{\textbf{\xmark}} & \colorbox{babyblue}{\textbf{\cmark}} & \colorbox{fluorescentorange}{\textbf{\xmark}} \\
        \midrule
        Random Guessing \hspace{10mm} & 50 & 50 & 50 & 50 & 50 & 50 & 50 & 50 & 33.3 & 33.3 & 33.3 & 33.3 \\
        \midrule
        Flan-PaLM-8B & 68 & 68 & 48 & 55 & 82 & 74 & 54 & 50 & 28.8 & 36.4 & 33.3 & 31.8 \\
        \ \ \ \ \ \ + Instruction tuning & 62 & 65 & 55 & 55 & 80 & 82 & 58 & 52 & 19.7 & 30.3 & 34.8 & 33.3 \\
        \ \ \ \ \ \ + Symbol tuning (ours) & 81 & 71 & 77 & 64 & 84 & 84 & 72 & 82 & 21.2 & 21.2 & 31.8 & 30.3 \\
        \\
        Flan-PaLM-62B & 79 & 82 & 51 & 63 & 90 & 72 & 70 & 62 & 66.7 & 54.5 & 56.1 & 40.9 \\
        \ \ \ \ \ \ + Instruction tuning & 82 & 85 & 56 & 69 & 88 & 68 & 72 & 62 & 68.2 & 60.6 & 59.1 & 47.0 \\
        \ \ \ \ \ \ + Symbol tuning (ours) & 82 & 79 & 89 & 72 & 88 & 84 & 84 & 84 & 57.6 & 47.0 & 47.0 & 50.0 \\
        \\
        Flan-cont-PaLM-62B & 93 & 84 & 32 & 59 & 88 & 86 & 54 & 62 & 66.7 & 56.1 & 56.1 & 39.4 \\
        \ \ \ \ \ \ + Instruction tuning & 91 & 87 & 42 & 67 & 88 & 92 & 70 & 58 & 59.1 & 45.5 & 38.5 & 39.4 \\
        \ \ \ \ \ \ + Symbol tuning (ours) & 92 & 90 & 82 & 77 & 86 & 84 & 82 & 88 & 65.2 & 47.0 & 54.5 & 48.5 \\
        \\
        Flan-PaLM-540B & 93 & 89 & 84 & 77 & 90 & 90 & 78 & 62 & 71.2 & 69.7 & 66.7 & 60.6 \\
        \ \ \ \ \ \ + Instruction tuning & 94 & 92 & 86 & 75 & 90 & 92 & 84 & 60 & 72.7 & 71.2 & 71.2 & 65.2 \\
        \ \ \ \ \ \ + Symbol tuning (ours) & 97 & 88 & 93 & 60 & 92 & 90 & 90 & 92 & 81.8 & 78.8 & 72.7 & 65.2 \\
        \bottomrule
    \end{tabular}
    }
\end{table}

\begin{table}[th]
    \vspace{8mm}
    \caption{
    TEAT, TEFE, and TEH 4-shot task performance.
    }
    \label{tab:icl-settings-per-task-3}
    \centering
    \scriptsize
    \arrayrulecolor{black}
    \resizebox{\columnwidth}{!}{
    \begin{tabular}{l cccc cccc cccc}
        \toprule
        & \multicolumn{4}{c}{\thead{\scriptsize \textbf{TEAT}}} & \multicolumn{4}{c}{\thead{\scriptsize \textbf{TEFE}}} & \multicolumn{4}{c}{\thead{\scriptsize \textbf{TEH}}}\\
        \cmidrule[0.5pt](lr){2-5}
        \cmidrule[0.5pt](lr){6-9}
        \cmidrule[0.5pt](lr){10-13}
        \hspace{9mm} \textbf{Relevant labels:} & \colorbox{babyblue}{\textbf{\cmark}} & \colorbox{babyblue}{\textbf{\cmark}} & \colorbox{fluorescentorange}{\textbf{\xmark}} & \colorbox{fluorescentorange}{\textbf{\xmark}} & \colorbox{babyblue}{\textbf{\cmark}} & \colorbox{babyblue}{\textbf{\cmark}} & \colorbox{fluorescentorange}{\textbf{\xmark}} & \colorbox{fluorescentorange}{\textbf{\xmark}} & \colorbox{babyblue}{\textbf{\cmark}} & \colorbox{babyblue}{\textbf{\cmark}} & \colorbox{fluorescentorange}{\textbf{\xmark}} & \colorbox{fluorescentorange}{\textbf{\xmark}} \\
        \hspace{7mm} \textbf{Task instructions:} & \colorbox{babyblue}{\textbf{\cmark}} & \colorbox{fluorescentorange}{\textbf{\xmark}} & \colorbox{babyblue}{\textbf{\cmark}} & \colorbox{fluorescentorange}{\textbf{\xmark}} & \colorbox{babyblue}{\textbf{\cmark}} & \colorbox{fluorescentorange}{\textbf{\xmark}} & \colorbox{babyblue}{\textbf{\cmark}} & \colorbox{fluorescentorange}{\textbf{\xmark}} & \colorbox{babyblue}{\textbf{\cmark}} & \colorbox{fluorescentorange}{\textbf{\xmark}} & \colorbox{babyblue}{\textbf{\cmark}} & \colorbox{fluorescentorange}{\textbf{\xmark}} \\
        \midrule
        Random Guessing \hspace{10mm} & 33.3 & 33.3 & 33.3 & 33.3 & 33.3 & 33.3 & 33.3 & 33.3 & 50 & 50 & 50 & 50 \\
        \midrule
        Flan-PaLM-8B & 23.1 & 23.1 & 28.8 & 30.8 & 49.3 & 37.3 & 32.8 & 28.4 & 69 & 70 & 47 & 52 \\
        \ \ \ \ \ \ + Instruction tuning & 19.2 & 21.2 & 36.5 & 30.8 & 43.3 & 32.8 & 29.9 & 31.3 & 68 & 71 & 50 & 47 \\
        \ \ \ \ \ \ + Symbol tuning (ours) & 19.2 & 17.3 & 44.2 & 55.8 & 37.3 & 32.8 & 46.3 & 23.9 & 60 & 61 & 59 & 62 \\
        \\
        Flan-PaLM-62B & 44.2 & 36.5 & 42.3 & 38.5 & 73.1 & 58.2 & 56.7 & 40.3 & 78 & 72 & 59 & 52 \\
        \ \ \ \ \ \ + Instruction tuning & 46.2 & 30.8 & 44.2 & 40.4 & 74.6 & 61.2 & 58.2 & 47.8 & 76 & 72 & 59 & 57 \\
        \ \ \ \ \ \ + Symbol tuning (ours) & 59.6 & 46.2 & 48.1 & 44.2 & 65.7 & 44.8 & 59.7 & 58.2 & 73 & 70 & 58 & 60 \\
        \\
        Flan-cont-PaLM-62B & 46.2 & 23.1 & 44.2 & 42.3 & 76.1 & 59.7 & 59.7 & 44.8 & 71 & 79 & 51 & 56 \\
        \ \ \ \ \ \ + Instruction tuning & 53.8 & 34.6 & 38.5 & 38.5 & 64.2 & 59.7 & 49.3 & 41.8 & 73 & 77 & 60 & 57 \\
        \ \ \ \ \ \ + Symbol tuning (ours) & 63.5 & 53.8 & 57.7 & 57.7 & 62.7 & 64.2 & 62.7 & 50.7 & 75 & 69 & 64 & 60 \\
        \\
        Flan-PaLM-540B & 73.1 & 59.6 & 69.2 & 57.7 & 80.6 & 68.7 & 70.1 & 56.7 & 73 & 74 & 65 & 60 \\
        \ \ \ \ \ \ + Instruction tuning & 73.1 & 69.2 & 69.2 & 65.4 & 79.1 & 74.6 & 70.1 & 53.7 & 76 & 75 & 66 & 58 \\
        \ \ \ \ \ \ + Symbol tuning (ours) & 78.8 & 61.5 & 67.3 & 59.6 & 71.6 & 61.2 & 71.6 & 47.8 & 71 & 70 & 65 & 64 \\
        \bottomrule
    \end{tabular}
    }

    \vspace{8mm}
    \caption{
    TEHI, TOS, and average across eleven tasks 4-shot task performance.
    }
    \label{tab:icl-settings-per-task-4}
    \centering
    \scriptsize
    \arrayrulecolor{black}
    \resizebox{\columnwidth}{!}{
    \begin{tabular}{l cccc cccc cccc}
        \toprule
        & \multicolumn{4}{c}{\thead{\scriptsize \textbf{TEHI}}} & \multicolumn{4}{c}{\thead{\scriptsize \textbf{TOS}}} & \multicolumn{4}{c}{\thead{\scriptsize \textbf{Average}}}\\
        \cmidrule[0.5pt](lr){2-5}
        \cmidrule[0.5pt](lr){6-9}
        \cmidrule[0.5pt](lr){10-13}
        \hspace{9mm} \textbf{Relevant labels:} & \colorbox{babyblue}{\textbf{\cmark}} & \colorbox{babyblue}{\textbf{\cmark}} & \colorbox{fluorescentorange}{\textbf{\xmark}} & \colorbox{fluorescentorange}{\textbf{\xmark}} & \colorbox{babyblue}{\textbf{\cmark}} & \colorbox{babyblue}{\textbf{\cmark}} & \colorbox{fluorescentorange}{\textbf{\xmark}} & \colorbox{fluorescentorange}{\textbf{\xmark}} & \colorbox{babyblue}{\textbf{\cmark}} & \colorbox{babyblue}{\textbf{\cmark}} & \colorbox{fluorescentorange}{\textbf{\xmark}} & \colorbox{fluorescentorange}{\textbf{\xmark}} \\
        \hspace{7mm} \textbf{Task instructions:} & \colorbox{babyblue}{\textbf{\cmark}} & \colorbox{fluorescentorange}{\textbf{\xmark}} & \colorbox{babyblue}{\textbf{\cmark}} & \colorbox{fluorescentorange}{\textbf{\xmark}} & \colorbox{babyblue}{\textbf{\cmark}} & \colorbox{fluorescentorange}{\textbf{\xmark}} & \colorbox{babyblue}{\textbf{\cmark}} & \colorbox{fluorescentorange}{\textbf{\xmark}} & \colorbox{babyblue}{\textbf{\cmark}} & \colorbox{fluorescentorange}{\textbf{\xmark}} & \colorbox{babyblue}{\textbf{\cmark}} & \colorbox{fluorescentorange}{\textbf{\xmark}} \\
        \midrule
        Random Guessing \hspace{10mm} & 33.3 & 33.3 & 33.3 & 33.3 & 50 & 50 & 50 & 50 & 42.4 & 42.4 & 42.4 & 42.4 \\
        \midrule
        Flan-PaLM-8B & 40.6 & 39.1 & 30.4 & 40.6 & 80 & 80 & 58 & 54 & 63.9 & 61.6 & 42.4 & 44.2 \\
        \ \ \ \ \ \ + Instruction tuning & 30.4 & 26.1 & 37.7 & 42.0 & 76 & 82 & 58 & 54 & 59.9 & 61.3 & 44.4 & 45.6 \\
        \ \ \ \ \ \ + Symbol tuning (ours) & 30.4 & 26.1 & 43.5 & 33.3 & 64 & 58 & 66 & 60 & 57.6 & 54.3 & 58.2 & 52.8 \\
        \\
        Flan-PaLM-62B & 58.0 & 55.1 & 46.4 & 29.0 & 88 & 84 & 70 & 52 & 74.3 & 70.0 & 57.0 & 50.5 \\
        \ \ \ \ \ \ + Instruction tuning & 59.4 & 53.6 & 46.4 & 40.6 & 84 & 90 & 66 & 56 & 75.5 & 70.8 & 59.9 & 54.3 \\
        \ \ \ \ \ \ + Symbol tuning (ours) & 60.9 & 52.2 & 55.1 & 44.9 & 86 & 82 & 80 & 58 & 75.5 & 70.8 & 71.4 & 60.3 \\
        \\
        Flan-cont-PaLM-62B & 59.4 & 49.3 & 47.8 & 42.0 & 88 & 80 & 72 & 54 & 77.3 & 70.3 & 56.3 & 51.0 \\
        \ \ \ \ \ \ + Instruction tuning & 60.9 & 44.9 & 50.7 & 40.6 & 88 & 82 & 64 & 64 & 77.1 & 72.2 & 59.0 & 52.4 \\
        \ \ \ \ \ \ + Symbol tuning (ours) & 58.0 & 56.5 & 44.9 & 34.8 & 82 & 88 & 74 & 72 & 78.9 & 74.5 & 71.8 & 62.1 \\
        \\
        Flan-PaLM-540B & 59.4 & 60.9 & 56.5 & 44.9 & 80 & 76 & 62 & 58 & 82.2 & 77.4 & 70.7 & 58.1 \\
        \ \ \ \ \ \ + Instruction tuning & 59.4 & 62.3 & 60.9 & 43.5 & 82 & 80 & 66 & 56 & 82.4 & 79.8 & 73.6 & 59.5 \\
        \ \ \ \ \ \ + Symbol tuning (ours) & 63.8 & 60.9 & 56.5 & 33.3 & 90 & 84 & 88 & 70 & 84.4 & 78.8 & 80.0 & 63.6 \\
        \bottomrule
    \end{tabular}
    }
\end{table}

\clearpage
\subsection{MMLU}
\label{sec:appendix-mmlu}
MMLU consists of 57 tasks that test a model's knowledge and problem-solving abilities \citep{Hendrycks2021MMLU}.
We evaluate on MMLU in a five-shot setting where few-shot exemplars are from the ``dev'' set, following \citet{chung2022scaling}.
In this section, we report the ``validation'' set performance on MMLU for each task.
We use the same prompts as \citet{chung2022scaling}, which can be found at \url{https://github.com/jasonwei20/flan-2}.
Prompts for STEM datasets are also the same as in \citet{chung2022scaling}, which originated from \citet{lewkowycz2022solving}.
We show full experimental results for Flan-PaLM models and symbol-tuned variants (after tuning for 4k steps for 8B and 62B models and 1k steps for 540B models) on MMLU in \cref{tab:mmlu-per-task-1}, \cref{tab:mmlu-per-task-2}, \cref{tab:mmlu-per-task-3}, \cref{tab:mmlu-per-task-4}, \cref{tab:mmlu-per-task-5}, and \cref{tab:mmlu-per-task-6}.

\begin{table}[hb]
\centering
\caption{MMLU [:10] 5-shot individual task performance.}
\label{tab:mmlu-per-task-1}
\setlength{\tabcolsep}{3pt}
\resizebox{\columnwidth}{!}{
\begin{tabular}{llcccccccccccccccccccc}
    \toprule
        & \multicolumn{11}{c}\thead{MMLU} \\
    \cmidrule(lr){3-22}
        &
        & \multicolumn{2}{c}{\thead{Abstract \\Algebra}} &
        \multicolumn{2}{c}{\thead{Anatomy}} &
        \multicolumn{2}{c}{\thead{Astronomy}} &
        \multicolumn{2}{c}{\thead{Business\\Ethics}} &
        \multicolumn{2}{c}{\thead{Clinical\\Knowledge}} &
        \multicolumn{2}{c}{\thead{College\\Biology}} &
        \multicolumn{2}{c}{\thead{College\\Chemistry}} &
        \multicolumn{2}{c}{\thead{College\\Comp. Sci.}} &
        \multicolumn{2}{c}{\thead{College\\Math}} &
        \multicolumn{2}{c}{\thead{College\\Medicine}} \\
    \cmidrule(lr){3-4}
    \cmidrule(lr){5-6}
    \cmidrule(lr){7-8}
    \cmidrule(lr){9-10}
    \cmidrule(lr){11-12}
    \cmidrule(lr){13-14}
    \cmidrule(lr){15-16}
    \cmidrule(lr){17-18}
    \cmidrule(lr){19-20}
    \cmidrule(lr){21-22} 
        \multicolumn{2}{l}{Model} 
        & \thead{Direct} &
        \thead{CoT} &
        \thead{Direct} &
        \thead{CoT} &
        \thead{Direct} &
        \thead{CoT} &
        \thead{Direct} &
        \thead{CoT} &
        \thead{Direct} &
        \thead{CoT} &
        \thead{Direct} &
        \thead{CoT} &
        \thead{Direct} &
        \thead{CoT} &
        \thead{Direct} &
        \thead{CoT} &
        \thead{Direct} &
        \thead{CoT} &
        \thead{Direct} &
        \thead{CoT} \\
    \midrule
        8B & Flan-PaLM & 36.4 & 9.1 & 42.9 & 35.7 & 43.8 & 43.8 & 36.4 & 45.5 & 44.8 & 41.4 & 56.2 & 50.0 & 25.0 & 25.0 & 45.5 & 27.3 & 18.2 & 0.0 & 45.5 & 40.9 \\
        & + Symbol tuning & 18.2 & 9.1 & 50.0 & 50.0 & 56.2 & 25.0 & 45.5 & 45.5 & 34.5 & 44.8 & 56.2 & 50.0 & 25.0 & 12.5 & 45.5 & 54.5 & 27.3 & 0.0 & 59.1 & 27.3 \\
        \\
        62B & Flan-PaLM & 18.2 & 27.3 & 57.1 & 35.7 & 68.8 & 62.5 & 63.6 & 54.5 & 55.2 & 58.6 & 75.0 & 75.0 & 12.5 & 37.5 & 54.5 & 36.4 & 36.4 & 18.2 & 81.8 & 68.2 \\
        & + Symbol tuning & 18.2 & 36.4 & 42.9 & 28.6 & 68.8 & 62.5 & 54.5 & 45.5 & 62.1 & 62.1 & 62.5 & 68.8 & 37.5 & 37.5 & 36.4 & 27.3 & 27.3 & 18.2 & 77.3 & 77.3 \\
        \\
        62B & Flan-cont-PaLM & 27.3 & 18.2 & 71.4 & 64.3 & 81.2 & 68.8 & 63.6 & 54.5 & 69.0 & 62.1 & 75.0 & 81.2 & 37.5 & 37.5 & 54.5 & 27.3 & 45.5 & 36.4 & 72.7 & 81.8 \\
        & + Symbol tuning & 64.9 & 9.1 & 27.3 & 50.0 & 57.1 & 62.5 & 62.5 & 63.6 & 63.6 & 58.6 & 75.9 & 56.2 & 75.0 & 37.5 & 37.5 & 27.3 & 45.5 & 54.5 & 54.5 & 68.2 \\
        \\
        540B & Flan-PaLM & 0.0 & 9.1 & 57.1 & 71.4 & 81.2 & 68.8 & 63.6 & 63.6 & 79.3 & 69.0 & 87.5 & 62.5 & 50.0 & 50.0 & 81.8 & 63.6 & 36.4 & 36.4 & 86.4 & 81.8 \\
        & + Symbol tuning & 0.0 & 9.1 & 64.3 & 64.3 & 81.2 & 68.8 & 63.6 & 63.6 & 86.2 & 75.9 & 87.5 & 62.5 & 50.0 & 50.0 & 72.7 & 63.6 & 36.4 & 9.1 & 86.4 & 86.4 \\
    \bottomrule
\end{tabular}
}

\centering
\vspace{4mm}
\caption{MMLU [10:20] 5-shot individual task performance.}
\label{tab:mmlu-per-task-2}
\setlength{\tabcolsep}{3pt}
\resizebox{\columnwidth}{!}{%
\begin{tabular}{llcccccccccccccccccccc}
    \toprule
        & \multicolumn{11}{c}\thead{MMLU} \\
    \cmidrule(lr){3-22}
        &
        & \multicolumn{2}{c}{\thead{College\\Physics}} & \multicolumn{2}{c}{\thead{Computer\\Security}} & \multicolumn{2}{c}{\thead{Conceptual\\physics}} & \multicolumn{2}{c}{\thead{Econometrics}} & \multicolumn{2}{c}{\thead{Electrical\\Engineering}} & \multicolumn{2}{c}{\thead{Elementary\\Mathematics}} & \multicolumn{2}{c}{\thead{Formal\\Logic}} & \multicolumn{2}{c}{\thead{Global\\Facts}} & \multicolumn{2}{c}{\thead{High School\\Biology}} & \multicolumn{2}{c}{\thead{High School\\Chemistry}} \\
    \cmidrule(lr){3-4}
    \cmidrule(lr){5-6}
    \cmidrule(lr){7-8}
    \cmidrule(lr){9-10}
    \cmidrule(lr){11-12}
    \cmidrule(lr){13-14}
    \cmidrule(lr){15-16}
    \cmidrule(lr){17-18}
    \cmidrule(lr){19-20}
    \cmidrule(lr){21-22} 
        \multicolumn{2}{l}{Model} 
        & \thead{Direct} &
        \thead{CoT} &
        \thead{Direct} &
        \thead{CoT} &
        \thead{Direct} &
        \thead{CoT} &
        \thead{Direct} &
        \thead{CoT} &
        \thead{Direct} &
        \thead{CoT} &
        \thead{Direct} &
        \thead{CoT} &
        \thead{Direct} &
        \thead{CoT} &
        \thead{Direct} &
        \thead{CoT} &
        \thead{Direct} &
        \thead{CoT} &
        \thead{Direct} &
        \thead{CoT} \\
    \midrule
        8B & Flan-PaLM & 45.5 & 18.2 & 81.8 & 45.5 & 30.8 & 26.9 & 41.7 & 16.7 & 31.2 & 50.0 & 29.3 & 29.3 & 28.6 & 14.3 & 30.0 & 30.0 & 50.0 & 40.6 & 22.7 & 22.7 \\
        & + Symbol tuning & 27.3 & 27.3 & 36.4 & 9.1 & 34.6 & 34.6 & 33.3 & 8.3 & 37.5 & 50.0 & 31.7 & 31.7 & 21.4 & 28.6 & 0.0 & 50.0 & 40.6 & 25.0 & 27.3 & 31.8 \\
        \\
        62B & Flan-PaLM & 72.7 & 54.5 & 54.5 & 54.5 & 61.5 & 57.7 & 50.0 & 50.0 & 56.2 & 43.8 & 43.9 & 51.2 & 28.6 & 21.4 & 20.0 & 50.0 & 75.0 & 62.5 & 31.8 & 36.4 \\
        & + Symbol tuning & 54.5 & 36.4 & 54.5 & 45.5 & 61.5 & 53.8 & 41.7 & 33.3 & 50.0 & 50.0 & 46.3 & 63.4 & 21.4 & 28.6 & 30.0 & 30.0 & 75.0 & 59.4 & 40.9 & 50.0 \\
        \\
        62B & Flan-cont-PaLM & 63.6 & 54.5 & 72.7 & 54.5 & 61.5 & 65.4 & 50.0 & 33.3 & 56.2 & 68.8 & 53.7 & 80.5 & 21.4 & 14.3 & 40.0 & 50.0 & 68.8 & 62.5 & 27.3 & 45.5 \\
        & + Symbol tuning & 81.8 & 45.5 & 63.6 & 54.5 & 54.5 & 61.5 & 65.4 & 33.3 & 33.3 & 75.0 & 50.0 & 78.0 & 46.3 & 50.0 & 42.9 & 50.0 & 50.0 & 59.4 & 71.9 & 31.8 \\
        \\
        540B & Flan-PaLM & 63.6 & 72.7 & 72.7 & 63.6 & 65.4 & 65.4 & 66.7 & 66.7 & 87.5 & 75.0 & 63.4 & 70.7 & 57.1 & 57.1 & 50.0 & 70.0 & 75.0 & 71.9 & 63.6 & 54.5 \\
        & + Symbol tuning & 63.6 & 54.5 & 81.8 & 72.7 & 65.4 & 61.5 & 66.7 & 58.3 & 87.5 & 81.2 & 61.0 & 68.3 & 57.1 & 64.3 & 50.0 & 60.0 & 75.0 & 78.1 & 59.1 & 54.5 \\
    \bottomrule
\end{tabular}
}

\centering
\vspace{4mm}
\caption{MMLU [20:30] 5-shot individual task performance.}
\label{tab:mmlu-per-task-3}
\setlength{\tabcolsep}{3pt}
\resizebox{\columnwidth}{!}{%
\begin{tabular}{llcccccccccccccccccccc}
    \toprule
        & \multicolumn{11}{c}\thead{MMLU} \\
    \cmidrule(lr){3-22}
        &
        & \multicolumn{2}{c}{\thead{High School\\Comp. Sci.}} & \multicolumn{2}{c}{\thead{High School\\European History}} & \multicolumn{2}{c}{\thead{High School\\Geography}} & \multicolumn{2}{c}{\thead{High School\\Gvmt \& Politics}} & \multicolumn{2}{c}{\thead{High School\\Macroeconomics}} & \multicolumn{2}{c}{\thead{High School\\Math}} & \multicolumn{2}{c}{\thead{High School\\Microeconomics}} & \multicolumn{2}{c}{\thead{High School\\Physics}} & \multicolumn{2}{c}{\thead{High School\\Psychology}} & \multicolumn{2}{c}{\thead{High School\\Statistics}} \\
    \cmidrule(lr){3-4}
    \cmidrule(lr){5-6}
    \cmidrule(lr){7-8}
    \cmidrule(lr){9-10}
    \cmidrule(lr){11-12}
    \cmidrule(lr){13-14}
    \cmidrule(lr){15-16}
    \cmidrule(lr){17-18}
    \cmidrule(lr){19-20}
    \cmidrule(lr){21-22} 
        \multicolumn{2}{l}{Model} 
        & \thead{Direct} &
        \thead{CoT} &
        \thead{Direct} &
        \thead{CoT} &
        \thead{Direct} &
        \thead{CoT} &
        \thead{Direct} &
        \thead{CoT} &
        \thead{Direct} &
        \thead{CoT} &
        \thead{Direct} &
        \thead{CoT} &
        \thead{Direct} &
        \thead{CoT} &
        \thead{Direct} &
        \thead{CoT} &
        \thead{Direct} &
        \thead{CoT} &
        \thead{Direct} &
        \thead{CoT} \\
    \midrule
        8B & Flan-PaLM & 44.4 & 33.3 & 72.2 & 61.1 & 68.2 & 54.5 & 57.1 & 57.1 & 44.2 & 39.5 & 24.1 & 17.2 & 57.7 & 38.5 & 35.3 & 17.6 & 66.7 & 45.0 & 39.1 & 39.1 \\
        & + Symbol tuning & 66.7 & 55.6 & 77.8 & 50.0 & 63.6 & 59.1 & 66.7 & 66.7 & 39.5 & 46.5 & 34.5 & 20.7 & 57.7 & 30.8 & 35.3 & 23.5 & 61.7 & 48.3 & 43.5 & 34.8 \\
        \\
        62B & Flan-PaLM & 55.6 & 55.6 & 88.9 & 66.7 & 77.3 & 81.8 & 76.2 & 71.4 & 58.1 & 55.8 & 13.8 & 27.6 & 69.2 & 57.7 & 23.5 & 17.6 & 88.3 & 83.3 & 52.2 & 43.5 \\
        & + Symbol tuning & 44.4 & 55.6 & 88.9 & 77.8 & 86.4 & 72.7 & 76.2 & 71.4 & 58.1 & 67.4 & 24.1 & 27.6 & 73.1 & 69.2 & 17.6 & 17.6 & 88.3 & 86.7 & 47.8 & 39.1 \\
        \\
        62B & Flan-cont-PaLM & 55.6 & 55.6 & 88.9 & 83.3 & 95.5 & 86.4 & 85.7 & 85.7 & 62.8 & 72.1 & 24.1 & 41.4 & 88.5 & 80.8 & 23.5 & 47.1 & 91.7 & 86.7 & 56.5 & 47.8 \\
        & + Symbol tuning & 36.4 & 55.6 & 44.4 & 66.7 & 83.3 & 86.4 & 95.5 & 85.7 & 81.0 & 62.8 & 65.1 & 37.9 & 34.5 & 80.8 & 80.8 & 41.2 & 17.6 & 86.7 & 91.7 & 43.5 \\
        \\
        540B & Flan-PaLM & 100.0 & 100.0 & 77.8 & 77.8 & 100.0 & 95.5 & 95.2 & 85.7 & 76.7 & 72.1 & 34.5 & 37.9 & 100.0 & 88.5 & 23.5 & 23.5 & 93.3 & 90.0 & 65.2 & 47.8 \\
        & + Symbol tuning & 88.9 & 88.9 & 77.8 & 77.8 & 100.0 & 95.5 & 95.2 & 85.7 & 76.7 & 72.1 & 41.4 & 24.1 & 100.0 & 80.8 & 17.6 & 23.5 & 93.3 & 90.0 & 65.2 & 60.9 \\
    \bottomrule
\end{tabular}
}
\end{table}

\begin{table}[tb]
\centering
\vspace{4mm}
\caption{MMLU [30:40] 5-shot individual task performance.}
\label{tab:mmlu-per-task-4}
\setlength{\tabcolsep}{3pt}
\resizebox{\columnwidth}{!}{%
\begin{tabular}{llcccccccccccccccccccc}
    \toprule
        & \multicolumn{11}{c}\thead{MMLU} \\
    \cmidrule(lr){3-22}
        &
        &\multicolumn{2}{c}{\thead{High School\\US History}} &
        \multicolumn{2}{c}{\thead{High School\\World History}} &
        \multicolumn{2}{c}{\thead{Human\\Aging}} &
        \multicolumn{2}{c}{\thead{Human\\Sexuality}} &
        \multicolumn{2}{c}{\thead{International\\Law}} &
        \multicolumn{2}{c}{\thead{Jurisprudence}} &
        \multicolumn{2}{c}{\thead{Logical\\Fallacies}} &
        \multicolumn{2}{c}{\thead{Machine\\Learning}} &
        \multicolumn{2}{c}{\thead{Management}} &
        \multicolumn{2}{c}{\thead{Marketing}} \\
    \cmidrule(lr){3-4}
    \cmidrule(lr){5-6}
    \cmidrule(lr){7-8}
    \cmidrule(lr){9-10}
    \cmidrule(lr){11-12}
    \cmidrule(lr){13-14}
    \cmidrule(lr){15-16}
    \cmidrule(lr){17-18}
    \cmidrule(lr){19-20}
    \cmidrule(lr){21-22} 
        \multicolumn{2}{l}{Model} 
        & \thead{Direct} &
        \thead{CoT} &
        \thead{Direct} &
        \thead{CoT} &
        \thead{Direct} &
        \thead{CoT} &
        \thead{Direct} &
        \thead{CoT} &
        \thead{Direct} &
        \thead{CoT} &
        \thead{Direct} &
        \thead{CoT} &
        \thead{Direct} &
        \thead{CoT} &
        \thead{Direct} &
        \thead{CoT} &
        \thead{Direct} &
        \thead{CoT} &
        \thead{Direct} &
        \thead{CoT} \\
    \midrule
        8B & Flan-PaLM & 72.7 & 54.5 & 57.7 & 50.0 & 56.5 & 47.8 & 66.7 & 58.3 & 76.9 & 53.8 & 72.7 & 36.4 & 61.1 & 61.1 & 45.5 & 45.5 & 81.8 & 36.4 & 68.0 & 68.0 \\
        & + Symbol tuning & 72.7 & 45.5 & 46.2 & 46.2 & 60.9 & 52.2 & 66.7 & 50.0 & 69.2 & 30.8 & 54.5 & 45.5 & 61.1 & 55.6 & 36.4 & 27.3 & 81.8 & 45.5 & 68.0 & 56.0 \\
        \\
        62B & Flan-PaLM & 81.8 & 72.7 & 80.8 & 69.2 & 60.9 & 65.2 & 75.0 & 50.0 & 84.6 & 69.2 & 63.6 & 54.5 & 61.1 & 66.7 & 27.3 & 27.3 & 81.8 & 90.9 & 72.0 & 68.0 \\
        & + Symbol tuning & 63.6 & 77.3 & 73.1 & 73.1 & 60.9 & 65.2 & 66.7 & 50.0 & 84.6 & 69.2 & 54.5 & 72.7 & 61.1 & 61.1 & 45.5 & 36.4 & 81.8 & 90.9 & 76.0 & 76.0 \\
        \\
        62B & Flan-cont-PaLM & 81.8 & 63.6 & 80.8 & 84.6 & 69.6 & 73.9 & 66.7 & 41.7 & 84.6 & 84.6 & 54.5 & 72.7 & 72.2 & 72.2 & 36.4 & 36.4 & 100.0 & 90.9 & 84.0 & 72.0 \\
        & + Symbol tuning & 69.6 & 77.3 & 81.8 & 76.9 & 69.2 & 69.6 & 69.6 & 75.0 & 66.7 & 76.9 & 84.6 & 81.8 & 54.5 & 72.2 & 72.2 & 36.4 & 45.5 & 100.0 & 100.0 & 76.0 \\
        \\
        540B & Flan-PaLM & 90.9 & 90.9 & 84.6 & 76.9 & 82.6 & 82.6 & 83.3 & 75.0 & 92.3 & 76.9 & 72.7 & 63.6 & 77.8 & 72.2 & 45.5 & 36.4 & 81.8 & 90.9 & 88.0 & 80.0 \\
        & + Symbol tuning & 90.9 & 95.5 & 84.6 & 76.9 & 87.0 & 78.3 & 83.3 & 75.0 & 92.3 & 69.2 & 72.7 & 81.8 & 77.8 & 72.2 & 45.5 & 54.5 & 81.8 & 90.9 & 84.0 & 76.0 \\
    \bottomrule
\end{tabular}
}

\vspace{4mm}
\centering
\caption{MMLU [40:50] 5-shot individual task performance.}
\label{tab:mmlu-per-task-5}
\setlength{\tabcolsep}{3pt}
\resizebox{\columnwidth}{!}{%
\begin{tabular}{llcccccccccccccccccccc}
    \toprule
        & \multicolumn{11}{c}\thead{MMLU} \\
    \cmidrule(lr){3-22}
        &
        &\multicolumn{2}{c}{\thead{Medical\\Genetics}} &
        \multicolumn{2}{c}{\thead{Misc.}} &
        \multicolumn{2}{c}{\thead{Moral\\Disputes}} &
        \multicolumn{2}{c}{\thead{Moral\\Scenarios}} &
        \multicolumn{2}{c}{\thead{Nutrition}} &
        \multicolumn{2}{c}{\thead{Philosophy}} &
        \multicolumn{2}{c}{\thead{Prehistory}} &
        \multicolumn{2}{c}{\thead{Professional\\Accounting}} &
        \multicolumn{2}{c}{\thead{Professional\\Law}} &
        \multicolumn{2}{c}{\thead{Professional\\Medicine}} \\
    \cmidrule(lr){3-4}
    \cmidrule(lr){5-6}
    \cmidrule(lr){7-8}
    \cmidrule(lr){9-10}
    \cmidrule(lr){11-12}
    \cmidrule(lr){13-14}
    \cmidrule(lr){15-16}
    \cmidrule(lr){17-18}
    \cmidrule(lr){19-20}
    \cmidrule(lr){21-22} 
        \multicolumn{2}{l}{Model} 
        & \thead{Direct} &
        \thead{CoT} &
        \thead{Direct} &
        \thead{CoT} &
        \thead{Direct} &
        \thead{CoT} &
        \thead{Direct} &
        \thead{CoT} &
        \thead{Direct} &
        \thead{CoT} &
        \thead{Direct} &
        \thead{CoT} &
        \thead{Direct} &
        \thead{CoT} &
        \thead{Direct} &
        \thead{CoT} &
        \thead{Direct} &
        \thead{CoT} &
        \thead{Direct} &
        \thead{CoT} \\
    \midrule
        8B & Flan-PaLM & 63.6 & 54.5 & 68.6 & 58.1 & 42.1 & 36.8 & 29.0 & 33.0 & 54.5 & 36.4 & 55.9 & 52.9 & 42.9 & 42.9 & 35.5 & 25.8 & 33.5 & 31.8 & 51.6 & 35.5 \\
        & + Symbol tuning & 72.7 & 63.6 & 67.4 & 62.8 & 39.5 & 39.5 & 32.0 & 37.0 & 51.5 & 39.4 & 47.1 & 58.8 & 48.6 & 34.3 & 45.2 & 22.6 & 35.9 & 26.5 & 45.2 & 48.4 \\
        \\
        62B & Flan-PaLM & 90.9 & 90.9 & 80.2 & 76.7 & 65.8 & 63.2 & 22.0 & 46.0 & 72.7 & 51.5 & 64.7 & 67.6 & 51.4 & 60.0 & 32.3 & 35.5 & 47.1 & 35.3 & 61.3 & 71.0 \\
        & + Symbol tuning & 90.9 & 90.9 & 81.4 & 80.2 & 68.4 & 65.8 & 31.0 & 50.0 & 63.6 & 60.6 & 67.6 & 67.6 & 54.3 & 62.9 & 35.5 & 29.0 & 42.9 & 34.7 & 61.3 & 67.7 \\
        \\
        62B & Flan-cont-PaLM & 90.9 & 100.0 & 79.1 & 79.1 & 71.1 & 55.3 & 24.0 & 41.0 & 75.8 & 60.6 & 73.5 & 73.5 & 74.3 & 68.6 & 64.5 & 45.2 & 42.4 & 37.1 & 64.5 & 71.0 \\
        & + Symbol tuning & 88.0 & 100.0 & 90.9 & 76.7 & 77.9 & 73.7 & 71.1 & 42.0 & 43.0 & 66.7 & 75.8 & 67.6 & 73.5 & 71.4 & 71.4 & 48.4 & 64.5 & 40.6 & 45.3 & 61.3 \\
        \\
        540B & Flan-PaLM & 90.9 & 100.0 & 82.6 & 82.6 & 78.9 & 60.5 & 56.0 & 74.0 & 84.8 & 78.8 & 88.2 & 70.6 & 82.9 & 80.0 & 51.6 & 58.1 & 60.6 & 52.4 & 93.5 & 77.4 \\
        & + Symbol tuning & 90.9 & 90.9 & 83.7 & 84.9 & 78.9 & 60.5 & 50.0 & 48.0 & 84.8 & 72.7 & 85.3 & 76.5 & 82.9 & 82.9 & 58.1 & 58.1 & 59.4 & 53.5 & 90.3 & 74.2 \\
    \bottomrule
\end{tabular}
}

\vspace{4mm}
\centering
\caption{MMLU [50:57] 5-shot individual task performance.}
\label{tab:mmlu-per-task-6}
\setlength{\tabcolsep}{3pt}
\resizebox{\columnwidth}{!}{%
\begin{tabular}{llcccccccccccccccc}
    \toprule
        & \multicolumn{10}{c}\thead{MMLU} \\
    \cmidrule(lr){3-18}
        &
        &\multicolumn{2}{c}{\thead{Professional\\Psychology}} &
        \multicolumn{2}{c}{\thead{Public\\Relations}} &
        \multicolumn{2}{c}{\thead{Security\\Studies}} &
        \multicolumn{2}{c}{\thead{Sociology}} &
        \multicolumn{2}{c}{\thead{US Foreign\\Policy}} &
        \multicolumn{2}{c}{\thead{Virology}} &
        \multicolumn{2}{c}{\thead{World Religions}} &
        \multicolumn{2}{c}{\thead{\textbf{Average}}} \\
    \cmidrule(lr){3-4}
    \cmidrule(lr){5-6}
    \cmidrule(lr){7-8}
    \cmidrule(lr){9-10}
    \cmidrule(lr){11-12}
    \cmidrule(lr){13-14}
    \cmidrule(lr){15-16}
    \cmidrule(lr){17-18}
        \multicolumn{2}{l}{Model} 
        & \thead{Direct} &
        \thead{CoT} &
        \thead{Direct} &
        \thead{CoT} &
        \thead{Direct} &
        \thead{CoT} &
        \thead{Direct} &
        \thead{CoT} &
        \thead{Direct} &
        \thead{CoT} &
        \thead{Direct} &
        \thead{CoT} &
        \thead{Direct} &
        \thead{CoT} &
        \thead{Direct} &
        \thead{CoT} \\
    \midrule
        8B & Flan-PaLM & 46.4 & 43.5 & 50.0 & 41.7 & 44.4 & 37.0 & 68.2 & 54.5 & 63.6 & 45.5 & 38.9 & 27.8 & 78.9 & 78.9 & 49.5 & 39.7 \\
        & + Symbol tuning & 46.4 & 42.0 & 41.7 & 41.7 & 37.0 & 18.5 & 54.5 & 63.6 & 63.6 & 54.5 & 27.8 & 16.7 & 78.9 & 78.9 & 47.5 & 39.6 \\
        \\
        62B & Flan-PaLM & 71.0 & 66.7 & 50.0 & 50.0 & 70.4 & 48.1 & 81.8 & 68.2 & 90.9 & 100.0 & 55.6 & 38.9 & 89.5 & 84.2 & 59.8 & 56.2 \\
        & + Symbol tuning & 63.8 & 66.7 & 50.0 & 58.3 & 70.4 & 63.0 & 86.4 & 63.6 & 81.8 & 90.9 & 66.7 & 61.1 & 89.5 & 89.5 & 58.6 & 57.4\\
        \\
        62B & Flan-cont-PaLM & 66.7 & 69.6 & 58.3 & 75.0 & 74.1 & 59.3 & 90.9 & 81.8 & 100.0 & 90.9 & 61.1 & 44.4 & 94.7 & 89.5 & 65.3 & 62.9 \\
        & + Symbol tuning & 64.5 & 72.5 & 66.7 & 66.7 & 66.7 & 59.3 & 74.1 & 68.2 & 81.8 & 90.9 & 90.9 & 55.6 & 55.6 & 89.5 & 64.9 & 62.6\\
        \\
        540B & Flan-PaLM & 76.8 & 71.0 & 58.3 & 50.0 & 70.4 & 59.3 & 100.0 & 90.9 & 100.0 & 100.0 & 50.0 & 61.1 & 84.2 & 89.5 & 73.0 & 69.5 \\
        & + Symbol tuning & 78.3 & 72.5 & 58.3 & 58.3 & 66.7 & 55.6 & 100.0 & 86.4 & 100.0 & 100.0 & 50.0 & 55.6 & 78.9 & 89.5 & 72.8 & 68.4 \\
    \bottomrule
\end{tabular}
}
\end{table}

\clearpage
\subsection{BIG-Bench Hard}
\label{sec:appendix-big-bench-hard}
BIG-Bench Hard is a collection of challenging tasks from BIG-Bench.
Tasks were selected in \citet{suzgun2022challenging} by choosing tasks where model performance as recorded by \citet{bigbench} was better than the average human rater.
There are a total of 23 tasks in BIG-Bench Hard---two of these tasks have three subtasks \citep{suzgun2022challenging}.
Following \citet{chung2022scaling}, we treat these subtasks as distinct tasks and take an unweighted average.
Our prompts are the same as those used in \citet{chung2022scaling} which are also the same as the ones given in \citet{suzgun2022challenging}.
These prompts contain three in-context exemplars.
We show full experimental results for Flan-PaLM models and symbol-tuned variants (after tuning for 4k steps for 8B and 62B models and 1k steps for 540B models) on BIG-Bench Hard in \cref{tab:bbh-per-task-1}, \cref{tab:bbh-per-task-2}, and \cref{tab:bbh-per-task-3}.

\begin{table}[hb]
\centering
\caption{BIG-Bench Hard [:9] individual task performance.}
\label{tab:bbh-per-task-1}
\setlength{\tabcolsep}{3pt}
\resizebox{\columnwidth}{!}{%
\begin{tabular}{llcccccccccccccccccc}
    \toprule
        & & \multicolumn{18}{c}{BIG-Bench Hard}  \\
    \cmidrule(lr){3-20}
        &
        & \multicolumn{2}{c}{\thead{Boolean\\Expressions}} &
        \multicolumn{2}{c}{\thead{Causal\\Judgement}} &
        \multicolumn{2}{c}{\thead{Date\\Understanding}} &
        \multicolumn{2}{c}{\thead{Disambiguation\\QA}} &
        \multicolumn{2}{c}{\thead{Dyck\\Languages}} &
        \multicolumn{2}{c}{\thead{Formal\\Fallacies}} &
        \multicolumn{2}{c}{\thead{Geometric\\Shapes}} &
        \multicolumn{2}{c}{\thead{Hyperbaton}} &
        \multicolumn{2}{c}{\thead{Logical Deduction\\Five Objects}} \\
    \cmidrule(lr){3-4}
    \cmidrule(lr){5-6}
    \cmidrule(lr){7-8}
    \cmidrule(lr){9-10}
    \cmidrule(lr){11-12}
    \cmidrule(lr){13-14}
    \cmidrule(lr){15-16}
    \cmidrule(lr){17-18}
    \cmidrule(lr){19-20}
        \multicolumn{2}{l}{Model} 
        & \thead{Direct} &
        \thead{CoT} &
        \thead{Direct} &
        \thead{CoT} &
        \thead{Direct} &
        \thead{CoT} &
        \thead{Direct} &
        \thead{CoT} &
        \thead{Direct} &
        \thead{CoT} &
        \thead{Direct} &
        \thead{CoT} &
        \thead{Direct} &
        \thead{CoT} &
        \thead{Direct} &
        \thead{CoT} &
        \thead{Direct} &
        \thead{CoT} \\
    \midrule
        8B & Flan-PaLM & 46.8 & 44.4 & 60.4 & 54.5 & 10.4 & 34.0 & 58.0 & 39.2 & 15.6 & 0.0 & 49.2 & 51.6 & 13.6 & 4.4 & 62.4 & 32.8 & 23.6 & 22.0 \\
        & + Symbol tuning & 46.0 & 18.0 & 59.4 & 48.1 & 8.8 & 34.0 & 39.2 & 15.6 & 17.2 & 0.0 & 50.8 & 44.8 & 17.2 & 0.8 & 42.8 & 4.4 & 26.4 & 11.2 \\
        \\
        62B & Flan-PaLM & 66.8 & 74.4 & 64.7 & 65.8 & 43.6 & 63.6 & 69.2 & 26.4 & 1.6 & 0.4 & 55.6 & 48.8 & 17.2 & 16.8 & 74.8 & 56.8 & 53.6 & 35.6 \\
        & + Symbol tuning & 64.4 & 69.6 & 66.8 & 59.4 & 47.6 & 64.8 & 68.8 & 32.0 & 4.0 & 0.0 & 53.2 & 54.0 & 18.8 & 18.4 & 66.0 & 50.0 & 49.6 & 34.4 \\
        \\
        62B & Flan-cont-PaLM & 77.2 & 82.4 & 66.3 & 64.7 & 52.4 & 61.2 & 68.4 & 68.8 & 27.2 & 3.2 & 55.2 & 55.2 & 34.8 & 22.8 & 73.2 & 88.4 & 52.0 & 42.0 \\
        & + Symbol tuning & 78.8 & 83.2 & 69.0 & 62.0 & 54.4 & 68.8 & 71.2 & 68.0 & 22.0 & 8.0 & 57.2 & 52.4 & 38.0 & 31.6 & 77.2 & 74.8 & 47.6 & 34.8 \\
        \\
        540B & Flan-PaLM & 86.4 & 79.6 & 63.6 & 64.7 & 60.0 & 75.2 & 76.0 & 64.8 & 31.2 & 21.6 & 60.4 & 55.6 & 40.0 & 42.4 & 66.0 & 94.8 & 55.6 & 56.8 \\
        & + Symbol tuning & 85.2 & 83.2 & 67.4 & 64.7 & 61.6 & 77.2 & 74.4 & 79.2 & 34.0 & 20.8 & 64.4 & 56.4 & 42.8 & 44.4 & 66.0 & 94.8 & 56.0 & 58.0 \\
    \bottomrule
\end{tabular}
}

\vspace{5mm}
\caption{BIG-Bench Hard [9:18] individual task performance.}
\label{tab:bbh-per-task-2}
\setlength{\tabcolsep}{3pt}
\resizebox{\columnwidth}{!}{%
\begin{tabular}{llcccccccccccccccccc}
    \toprule
        & & \multicolumn{18}{c}{BIG-Bench Hard} \\
    \cmidrule(lr){3-20}
        &
        & \multicolumn{2}{c}{\thead{Logical Deduction\\Seven Objects}} &
        \multicolumn{2}{c}{\thead{Logical Deduction\\Three Objects}} &
        \multicolumn{2}{c}{\thead{Movie\\Recommendation}} &
        \multicolumn{2}{c}{\thead{Multistep\\Arithmetic}} &
        \multicolumn{2}{c}{\thead{Navigate}} &
        \multicolumn{2}{c}{\thead{Object\\Counting}} &
        \multicolumn{2}{c}{\thead{Penguins\\in a Table}} &
        \multicolumn{2}{c}{\thead{Reasoning about\\Colored Objects}} &
        \multicolumn{2}{c}{\thead{Ruin\\Names}} \\
    \cmidrule(lr){3-4}
    \cmidrule(lr){5-6}
    \cmidrule(lr){7-8}
    \cmidrule(lr){9-10}
    \cmidrule(lr){11-12}
    \cmidrule(lr){13-14}
    \cmidrule(lr){15-16}
    \cmidrule(lr){17-18}
    \cmidrule(lr){19-20}
        \multicolumn{2}{l}{Model} 
        & \thead{Direct} &
        \thead{CoT} &
        \thead{Direct} &
        \thead{CoT} &
        \thead{Direct} &
        \thead{CoT} &
        \thead{Direct} &
        \thead{CoT} &
        \thead{Direct} &
        \thead{CoT} &
        \thead{Direct} &
        \thead{CoT} &
        \thead{Direct} &
        \thead{CoT} &
        \thead{Direct} &
        \thead{CoT} &
        \thead{Direct} &
        \thead{CoT} \\
    \midrule
        8B & Flan-PaLM & 25.2 & 14.8 & 46.0 & 40.0 & 74.4 & 46.8 & 0.8 & 0.8 & 57.6 & 44.4 & 32.0 & 29.2 & 30.8 & 31.5 & 30.4 & 32.8 & 42.4 & 28.0 \\
        & + Symbol tuning & 13.6 & 4.4 & 41.2 & 24.8 & 64.4 & 29.6 & 0.4 & 0.8 & 58.4 & 2.8 & 26.0 & 29.6 & 34.2 & 26.7 & 26.0 & 29.2 & 46.4 & 6.4 \\
        \\
        62B & Flan-PaLM & 48.4 & 34.8 & 73.6 & 57.6 & 82.0 & 73.2 & 2.0 & 1.2 & 61.6 & 44.4 & 51.2 & 48.8 & 37.0 & 50.0 & 50.0 & 46.4 & 64.0 & 48.4 \\
        & + Symbol tuning & 48.0 & 29.6 & 66.4 & 55.6 & 82.0 & 79.2 & 1.2 & 2.0 & 62.4 & 52.4 & 48.4 & 56.0 & 39.0 & 52.1 & 48.4 & 52.0 & 60.4 & 52.4 \\
        \\
        62B & Flan-cont-PaLM & 52.0 & 33.2 & 70.8 & 52.0 & 83.2 & 84.0 & 0.8 & 17.2 & 62.4 & 69.6 & 54.0 & 68.4 & 43.2 & 56.8 & 50.0 & 60.4 & 64.4 & 74.0 \\
        & + Symbol tuning & 50.8 & 29.2 & 64.8 & 48.8 & 82.8 & 88.0 & 1.6 & 21.2 & 60.4 & 64.8 & 56.0 & 64.4 & 45.9 & 54.1 & 54.8 & 62.0 & 60.8 & 68.8 \\
        \\
        540B & Flan-PaLM & 54.0 & 49.6 & 86.0 & 90.4 & 84.0 & 86.0 & 0.4 & 32.8 & 67.2 & 78.0 & 55.6 & 88.4 & 56.8 & 69.2 & 67.2 & 81.6 & 80.8 & 63.2 \\
        & + Symbol tuning & 52.0 & 48.4 & 87.2 & 91.6 & 83.6 & 84.8 & 1.6 & 30.0 & 66.4 & 78.0 & 57.6 & 89.6 & 52.1 & 69.2 & 66.0 & 82.8 & 82.0 & 59.2 \\
    \bottomrule
\end{tabular}
}

\centering
\vspace{4mm}
\caption{BIG-Bench Hard [18:27] individual task performance.}
\label{tab:bbh-per-task-3}
\setlength{\tabcolsep}{3pt}
\resizebox{\columnwidth}{!}{%
\begin{tabular}{llcccccccccccccccccccc}
    \toprule
        & & \multicolumn{19}{c}{BIG-Bench Hard}  \\
    \cmidrule(lr){3-22}
        &
        & \multicolumn{2}{c}{\thead{Salient Translation\\Error Detection}} &
        \multicolumn{2}{c}{\thead{Snarks}} &
        \multicolumn{2}{c}{\thead{Sports\\Understanding}} &
        \multicolumn{2}{c}{\thead{Temporal\\Sequences}} &
        \multicolumn{2}{c}{\thead{Tracking Shuffled\\Objects (5)}} &
        \multicolumn{2}{c}{\thead{Tracking Shuffled\\Objects (7)}} &
        \multicolumn{2}{c}{\thead{Tracking Shuffled\\Objects (3)}} &
        \multicolumn{2}{c}{\thead{Web of\\Lies}} &
        \multicolumn{2}{c}{\thead{Word\\Sorting}} &
        \multicolumn{2}{c}{\thead{\textbf{Average}}} \\
    \cmidrule(lr){3-4}
    \cmidrule(lr){5-6}
    \cmidrule(lr){7-8}
    \cmidrule(lr){9-10}
    \cmidrule(lr){11-12}
    \cmidrule(lr){13-14}
    \cmidrule(lr){15-16}
    \cmidrule(lr){17-18}
    \cmidrule(lr){19-20}
    \cmidrule(lr){21-22}
        \multicolumn{2}{l}{Model} 
        & \thead{Direct} &
        \thead{CoT} &
        \thead{Direct} &
        \thead{CoT} &
        \thead{Direct} &
        \thead{CoT} &
        \thead{Direct} &
        \thead{CoT} &
        \thead{Direct} &
        \thead{CoT} &
        \thead{Direct} &
        \thead{CoT} &
        \thead{Direct} &
        \thead{CoT} &
        \thead{Direct} &
        \thead{CoT} &
        \thead{Direct} &
        \thead{CoT} &
        \thead{Direct} &
        \thead{CoT} \\
    \midrule
        8B & Flan-PaLM & 27.2 & 0.0 & 69.1 & 60.7 & 63.6 & 69.6 & 14.4 & 25.6 & 18.0 & 18.0 & 16.4 & 14.8 & 33.2 & 32.0 & 51.6 & 49.6 & 5.2 & 2.0 & 36.2 & 30.5\\
        & + Symbol tuning & 17.2 & 0.0 & 56.7 & 44.9 & 58.0 & 73.6 & 18.0 & 12.8 & 15.6 & 0.8 & 15.6 & 2.4 & 32.8 & 8.0 & 51.6 & 3.6 & 6.8 & 1.2 & 33.0 & 17.7 \\
        \\
        62B & Flan-PaLM & 44.4 & 38.4 & 82.6 & 83.1 & 79.2 & 82.4 & 31.6 & 39.6 & 22.0 & 23.2 & 14.8 & 20.8 & 22.4 & 32.8 & 48.4 & 89.6 & 10.4 & 9.2 & 47.1 & 44.9 \\
        & + Symbol tuning & 41.2 & 35.6 & 76.4 & 82.6 & 73.6 & 82.8 & 34.4 & 38.4 & 22.8 & 20.0 & 16.4 & 15.2 & 24.8 & 33.2 & 48.8 & 83.2 & 9.6 & 7.6 & 46.1 & 44.9 \\
        \\
        62B & Flan-cont-PaLM & 48.8 & 42.0 & 83.1 & 80.3 & 82.4 & 84.0 & 33.6 & 67.6 & 20.0 & 25.2 & 19.6 & 16.4 & 23.2 & 37.6 & 48.8 & 95.2 & 16.0 & 16.0 & 50.5 & 54.4 \\
        & + Symbol tuning & 50.8 & 42.0 & 79.8 & 82.6 & 74.4 & 85.2 & 26.8 & 59.6 & 21.2 & 26.4 & 21.6 & 18.8 & 22.0 & 34.4 & 47.2 & 94.8 & 15.6 & 14.0 & 50.1 & 53.4 \\
        \\
        540B & Flan-PaLM & 54.4 & 48.4 & 83.1 & 75.8 & 82.0 & 88.0 & 76.4 & 89.2 & 25.2 & 46.8 & 23.6 & 36.0 & 32.4 & 64.0 & 59.6 & 100.0 & 32.8 & 34.8 & 57.8 & 65.8 \\
        & + Symbol tuning & 54.4 & 53.6 & 71.9 & 77.5 & 83.2 & 88.0 & 71.6 & 90.0 & 24.0 & 54.0 & 26.8 & 44.0 & 30.0 & 65.2 & 67.6 & 100.0 & 32.4 & 31.6 & 57.9 & 67.3 \\
    \bottomrule
\end{tabular}
}
\end{table}

\clearpage
\subsection{MMLU (zero-shot)}
\label{sec:appendix-zero-shot-mmlu}
In this section, we show full experimental results for Flan-PaLM models and symbol-tuned variants (after tuning for 4k steps for 8B and 62B models and 1k steps for 540B models) on MMLU \citep{Hendrycks2021MMLU}.
These results are from evaluating models in a zero-shot setting rather than in a five-shot setting as was tested in \cref{sec:appendix-mmlu}.

\begin{table}[hb]
\centering
\caption{MMLU [:10] 0-shot individual task performance.}
\label{tab:mmlu-zero-shot-per-task-1}
\setlength{\tabcolsep}{3pt}
\resizebox{\columnwidth}{!}{
\begin{tabular}{llcccccccccc}
    \toprule
        & & \multicolumn{10}{c}{\thead{MMLU}} \\
    \cmidrule(lr){3-12}
        Model &  &
        \multicolumn{1}{c}{\thead{Abstract \\Algebra}} &
        \multicolumn{1}{c}{\thead{Anatomy}} &
        \multicolumn{1}{c}{\thead{Astronomy}} &
        \multicolumn{1}{c}{\thead{Business\\Ethics}} &
        \multicolumn{1}{c}{\thead{Clinical\\Knowledge}} &
        \multicolumn{1}{c}{\thead{College\\Biology}} &
        \multicolumn{1}{c}{\thead{College\\Chemistry}} &
        \multicolumn{1}{c}{\thead{College\\Comp. Sci.}} &
        \multicolumn{1}{c}{\thead{College\\Math}} &
        \multicolumn{1}{c}{\thead{College\\Medicine}} \\
    \midrule
        8B & Flan-PaLM & 27.3 & 57.1 & 68.8 & 36.4 & 41.4 & 56.2 & 37.5 & 36.4 & 9.1 & 45.5 \\
        & + Symbol tuning & 36.4 & 57.1 & 56.2 & 45.5 & 48.3 & 50.0 & 25.0 & 45.5 & 9.1 & 54.5 \\
        \\
        62B & Flan-PaLM & 27.3 & 64.3 & 75.0 & 63.6 & 55.2 & 75.0 & 37.5 & 63.6 & 36.4 & 72.7 \\
        & + Symbol tuning & 9.1 & 50.0 & 75.0 & 63.6 & 58.6 & 68.8 & 50.0 & 45.5 & 36.4 & 72.7 \\
        \\
        62B & Flan-cont-PaLM & 27.3 & 64.3 & 75.0 & 63.6 & 75.9 & 68.8 & 37.5 & 54.5 & 54.5 & 72.7 \\
        & + Symbol tuning & 9.1 & 50.0 & 75.0 & 63.6 & 72.4 & 68.8 & 37.5 & 36.4 & 36.4 & 81.8 \\
        \\
        540B & Flan-PaLM & 0.0 & 50.0 & 75.0 & 63.6 & 82.8 & 81.2 & 50.0 & 72.7 & 36.4 & 81.8 \\
        & + Symbol tuning & 9.1 & 57.1 & 75.0 & 63.6 & 82.8 & 87.5 & 50.0 & 63.6 & 36.4 & 81.8 \\
    \bottomrule
\end{tabular}
}

\centering
\vspace{4mm}
\caption{MMLU [10:20] 0-shot individual task performance.}
\label{tab:mmlu-zero-shot-per-task-2}
\setlength{\tabcolsep}{3pt}
\resizebox{\columnwidth}{!}{
\begin{tabular}{llcccccccccc}
    \toprule
        & & \multicolumn{10}{c}{\thead{MMLU}} \\
    \cmidrule(lr){3-12}
        Model &  &
        \multicolumn{1}{c}{\thead{College\\Physics}} &
        \multicolumn{1}{c}{\thead{Computer\\Security}} &
        \multicolumn{1}{c}{\thead{Conceptual\\physics}} &
        \multicolumn{1}{c}{\thead{Econometrics}} &
        \multicolumn{1}{c}{\thead{Electrical\\Engineering}} & \multicolumn{1}{c}{\thead{Elementary\\Mathematics}} &
        \multicolumn{1}{c}{\thead{Formal\\Logic}} &
        \multicolumn{1}{c}{\thead{Global\\Facts}} &
        \multicolumn{1}{c}{\thead{High School\\Biology}} &
        \multicolumn{1}{c}{\thead{High School\\Chemistry}} \\
    \midrule
        8B & Flan-PaLM & 54.5 & 54.5 & 38.5 & 25.0 & 56.2 & 29.3 & 28.6 & 50.0 & 43.8 & 22.7 \\
        & + Symbol tuning & 36.4 & 36.4 & 38.5 & 25.0 & 56.2 & 26.8 & 21.4 & 40.0 & 34.4 & 27.3 \\
        \\
        62B & Flan-PaLM & 72.7 & 54.5 & 53.8 & 50.0 & 43.8 & 39.0 & 35.7 & 30.0 & 68.8 & 31.8 \\
        & + Symbol tuning & 54.5 & 54.5 & 50.0 & 50.0 & 50.0 & 41.5 & 35.7 & 30.0 & 68.8 & 36.4 \\
        \\
        62B & Flan-cont-PaLM & 63.6 & 63.6 & 61.5 & 50.0 & 50.0 & 53.7 & 28.6 & 40.0 & 68.8 & 31.8 \\
        & + Symbol tuning & 54.5 & 63.6 & 61.5 & 50.0 & 50.0 & 46.3 & 57.1 & 40.0 & 65.6 & 36.4 \\
        \\
        540B & Flan-PaLM & 72.7 & 63.6 & 69.2 & 58.3 & 81.2 & 51.2 & 50.0 & 50.0 & 75.0 & 63.6 \\
        & + Symbol tuning & 72.7 & 72.7 & 65.4 & 66.7 & 81.2 & 51.2 & 50.0 & 50.0 & 78.1 & 54.5 \\
    \bottomrule
\end{tabular}
}

\centering
\vspace{4mm}
\caption{MMLU [20:30] 0-shot individual task performance.}
\label{tab:mmlu-zero-shot-per-task-3}
\setlength{\tabcolsep}{3pt}
\resizebox{\columnwidth}{!}{
\begin{tabular}{llcccccccccc}
    \toprule
        & & \multicolumn{10}{c}{\thead{MMLU}} \\
    \cmidrule(lr){3-12}
        Model &  &
        \multicolumn{1}{c}{\thead{High School\\Comp. Sci.}} &
        \multicolumn{1}{c}{\thead{High School\\European History}} &
        \multicolumn{1}{c}{\thead{High School\\Geography}} &
        \multicolumn{1}{c}{\thead{High School\\Gvmt \& Politics}} &
        \multicolumn{1}{c}{\thead{High School\\Macroeconomics}} &
        \multicolumn{1}{c}{\thead{High School\\Math}} &
        \multicolumn{1}{c}{\thead{High School\\Microeconomics}} &
        \multicolumn{1}{c}{\thead{High School\\Physics}} &
        \multicolumn{1}{c}{\thead{High School\\Psychology}} &
        \multicolumn{1}{c}{\thead{High School\\Statistics}} \\
    \midrule
        8B & Flan-PaLM & 33.3 & 66.7 & 68.2 & 61.9 & 44.2 & 27.6 & 61.5 & 47.1 & 65.0 & 39.1 \\
        & + Symbol tuning & 44.4 & 72.2 & 77.3 & 61.9 & 39.5 & 37.9 & 61.5 & 41.2 & 61.7 & 34.8 \\
        \\
        62B & Flan-PaLM & 55.6 & 88.9 & 81.8 & 76.2 & 62.8 & 20.7 & 69.2 & 29.4 & 88.3 & 47.8 \\
        & + Symbol tuning & 55.6 & 88.9 & 86.4 & 71.4 & 67.4 & 24.1 & 69.2 & 23.5 & 86.7 & 47.8 \\
        \\
        62B & Flan-cont-PaLM & 55.6 & 88.9 & 90.9 & 81.0 & 62.8 & 24.1 & 88.5 & 29.4 & 93.3 & 60.9 \\
        & + Symbol tuning & 44.4 & 83.3 & 90.9 & 76.2 & 65.1 & 37.9 & 80.8 & 17.6 & 95.0 & 56.5 \\
        \\
        540B & Flan-PaLM & 100.0 & 77.8 & 95.5 & 95.2 & 79.1 & 31.0 & 96.2 & 11.8 & 95.0 & 73.9 \\
        & + Symbol tuning & 88.9 & 77.8 & 95.5 & 95.2 & 76.7 & 34.5 & 96.2 & 11.8 & 95.0 & 65.2 \\
    \bottomrule
\end{tabular}
}
\end{table}

\begin{table}[tb]
\centering
\vspace{4mm}
\caption{MMLU [30:40] 0-shot individual task performance.}
\label{tab:mmlu-zero-shot-per-task-4}
\setlength{\tabcolsep}{3pt}
\resizebox{\columnwidth}{!}{
\begin{tabular}{llcccccccccc}
    \toprule
        & & \multicolumn{10}{c}{\thead{MMLU}} \\
    \cmidrule(lr){3-12}
        Model &  &
        \multicolumn{1}{c}{\thead{High School\\US History}} &
        \multicolumn{1}{c}{\thead{High School\\World History}} &
        \multicolumn{1}{c}{\thead{Human\\Aging}} &
        \multicolumn{1}{c}{\thead{Human\\Sexuality}} &
        \multicolumn{1}{c}{\thead{International\\Law}} &
        \multicolumn{1}{c}{\thead{Jurisprudence}} &
        \multicolumn{1}{c}{\thead{Logical\\Fallacies}} &
        \multicolumn{1}{c}{\thead{Machine\\Learning}} &
        \multicolumn{1}{c}{\thead{Management}} &
        \multicolumn{1}{c}{\thead{Marketing}} \\
    \midrule
        8B & Flan-PaLM & 72.7 & 73.1 & 43.5 & 66.7 & 84.6 & 72.7 & 61.1 & 36.4 & 81.8 & 80.0 \\
        & + Symbol tuning & 68.2 & 50.0 & 47.8 & 66.7 & 84.6 & 63.6 & 55.6 & 36.4 & 81.8 & 72.0 \\
        \\
        62B & Flan-PaLM & 81.8 & 80.8 & 65.2 & 75.0 & 84.6 & 72.7 & 66.7 & 36.4 & 81.8 & 88.0 \\
        & + Symbol tuning & 68.2 & 73.1 & 56.5 & 75.0 & 84.6 & 63.6 & 66.7 & 45.5 & 81.8 & 80.0 \\
        \\
        62B & Flan-cont-PaLM & 86.4 & 84.6 & 69.6 & 66.7 & 84.6 & 54.5 & 72.2 & 36.4 & 100.0 & 80.0 \\
        & + Symbol tuning & 86.4 & 76.9 & 65.2 & 66.7 & 84.6 & 54.5 & 72.2 & 45.5 & 100.0 & 80.0 \\
        \\
        540B & Flan-PaLM & 86.4 & 88.5 & 69.6 & 83.3 & 92.3 & 72.7 & 77.8 & 45.5 & 90.9 & 76.0 \\
        & + Symbol tuning & 90.9 & 88.5 & 73.9 & 83.3 & 84.6 & 72.7 & 77.8 & 45.5 & 90.9 & 76.0 \\
    \bottomrule
\end{tabular}
}

\vspace{4mm}
\centering
\caption{MMLU [40:50] 0-shot individual task performance.}
\label{tab:mmlu-zero-shot-per-task-5}
\setlength{\tabcolsep}{3pt}
\resizebox{\columnwidth}{!}{
\begin{tabular}{llcccccccccc}
    \toprule
        & & \multicolumn{10}{c}{\thead{MMLU}} \\
    \cmidrule(lr){3-12}
        Model &  &
        \multicolumn{1}{c}{\thead{Medical\\Genetics}} &
       \multicolumn{1}{c}{\thead{Misc.}} &
       \multicolumn{1}{c}{\thead{Moral\\Disputes}} &
       \multicolumn{1}{c}{\thead{Moral\\Scenarios}} &
       \multicolumn{1}{c}{\thead{Nutrition}} &
       \multicolumn{1}{c}{\thead{Philosophy}} &
       \multicolumn{1}{c}{\thead{Prehistory}} &
       \multicolumn{1}{c}{\thead{Professional\\Accounting}} &
       \multicolumn{1}{c}{\thead{Professional\\Law}} &
       \multicolumn{1}{c}{\thead{Professional\\Medicine}} \\
    \midrule
        8B & Flan-PaLM & 63.6 & 68.6 & 42.1 & 27.0 & 51.5 & 58.8 & 45.7 & 29.0 & 31.2 & 51.6 \\
        & + Symbol tuning & 72.7 & 65.1 & 36.8 & 29.0 & 42.4 & 52.9 & 48.6 & 45.2 & 32.9 & 51.6 \\
        \\
        62B & Flan-PaLM & 90.9 & 79.1 & 60.5 & 27.0 & 69.7 & 61.8 & 54.3 & 29.0 & 44.7 & 61.3 \\
        & + Symbol tuning & 90.9 & 80.2 & 60.5 & 29.0 & 63.6 & 64.7 & 60.0 & 35.5 & 41.8 & 61.3 \\
        \\
        62B & Flan-cont-PaLM & 90.9 & 82.6 & 71.1 & 34.0 & 72.7 & 79.4 & 74.3 & 58.1 & 41.2 & 64.5 \\
        & + Symbol tuning & 90.9 & 79.1 & 68.4 & 49.0 & 69.7 & 79.4 & 65.7 & 58.1 & 40.6 & 71.0 \\
        \\
        540B & Flan-PaLM & 90.9 & 83.7 & 78.9 & 53.0 & 81.8 & 76.5 & 71.4 & 61.3 & 58.8 & 87.1 \\
        & + Symbol tuning & 90.9 & 83.7 & 76.3 & 57.0 & 78.8 & 76.5 & 71.4 & 61.3 & 58.8 & 87.1 \\
    \bottomrule
\end{tabular}
}

\vspace{4mm}
\centering
\caption{MMLU [50:57] 0-shot individual task performance.}
\label{tab:mmlu-zero-shot-per-task-6}
\setlength{\tabcolsep}{3pt}
\resizebox{\columnwidth}{!}{
\begin{tabular}{llcccccccc}
    \toprule
        & & \multicolumn{8}{c}{\thead{MMLU}} \\
    \cmidrule(lr){3-10}
        Model &  &
        \multicolumn{1}{c}{\thead{Professional\\Psychology}} &
        \multicolumn{1}{c}{\thead{Public\\Relations}} &
        \multicolumn{1}{c}{\thead{Security\\Studies}} &
        \multicolumn{1}{c}{\thead{Sociology}} &
        \multicolumn{1}{c}{\thead{US Foreign\\Policy}} &
        \multicolumn{1}{c}{\thead{Virology}} &
        \multicolumn{1}{c}{\thead{World Religions}} &
        \multicolumn{1}{c}{\thead{\textbf{Average}}} \\
    \midrule
        8B & Flan-PaLM & 46.4 & 33.3 & 44.4 & 77.3 & 72.7 & 33.3 & 68.4 & 50.0 \\
        & + Symbol tuning & 42.0 & 50.0 & 48.1 & 68.2 & 63.6 & 38.9 & 68.4 & 48.9 \\
        \\
        62B & Flan-PaLM & 65.2 & 50.0 & 70.4 & 86.4 & 72.7 & 66.7 & 84.2 & 61.0 \\
        & + Symbol tuning & 63.8 & 50.0 & 70.4 & 90.9 & 72.7 & 72.2 & 89.5 & 59.9 \\
        \\
        62B & Flan-cont-PaLM & 65.2 & 58.3 & 74.1 & 90.9 & 90.9 & 61.1 & 94.7 & 65.3 \\
        & + Symbol tuning & 63.8 & 58.3 & 74.1 & 81.8 & 81.8 & 61.1 & 94.7 & 63.6 \\
        \\
        540B & Flan-PaLM & 73.9 & 50.0 & 77.8 & 95.5 & 100.0 & 50.0 & 84.2 & 70.9 \\
        & + Symbol tuning & 73.9 & 58.3 & 70.4 & 90.9 & 100.0 & 50.0 & 78.9 & 70.8 \\
    \bottomrule
\end{tabular}
}
\end{table}

\clearpage
\section{Example Prompts}
\label{sec:appendix-example-prompts}
\subsection{Symbol tuning prompts}
\label{sec:appendix-symbol-tuning-prompts}
In this section, we provide an example of a full few-shot prompt for each of the 22 datasets used in the main paper.
When generating these prompts, we follow the procedure describe in \cref{sec:tuning-tasks}.
Namely, prompts use one of ten possible formats shown in \cref{sec:appendix-prompt-formatting} and contain 2-10 in-context exemplars per class.
Original labels are remapped to arbitrary symbols as described in \cref{sec:tuning-tasks}.

\subsubsection{RTE}
\textbf{Overview}.
This prompt contains $k = 2$ in-context exemplars per class.
The original natural language labels [``entailment'', ``not entailment''] have been remapped to  [``4348'', ``forests''], respectively.

\textbf{Prompt:}

Input: A zoo worker is dead and two visitors are seriously injured after a Siberian tiger escaped from the San Francisco Zoo in San Francisco, California in the United States and attacked three people who were inside a cafe. The tiger was shot dead by police who were called to the scene. They found the tiger attacking one of the zoo visitors when they killed it.

A tiger attacked three people in San Francisco.

Output: 4348

Input: After the 1979 Soviet invasion and occupation, 3 million Afghans fled to Pakistan, which was encouraged by hefty Western aid to take them in.

Afghanistan was invaded by the Soviet Union in 1979.

Output: 4348

Input: In the May 2005 general election Michael Howard failed to unseat the Labour Government, although the Conservatives did gain 33 seats, playing the most significant role in reducing Labour's majority from 167 to 66.

In the May 2005 general election Conservatives got 33 seats.

Output: forests

Input: David Millar retained his Tour de France leader's yellow jersey despite crashing in the final two kilometres of the third stage of the race to Nantes.

Tour de France winner is likely to lose the crown.

Output: forests

Input: New Zealand's Qualifications Authority said Friday that it still strongly discourages students from using anything other than full English, but that credit will be given if the answer "clearly shows the required understanding," even if it contains text-speak.

Full English is recommended by New Zealand's Qualifications Authority.

Output:

\textbf{Answer:}

4348

\subsubsection{WNLI}
\textbf{Overview}.
This prompt contains $k = 6$ in-context exemplars per class.
The original natural language labels [``entailment'', ``not entailment''] have been remapped to  [``MIC'', ``certification''], respectively.

\textbf{Prompt:}

Input: Even before they reached town, they could hear a sound like corn popping. Dora asked what it was, and Dad said it was firecrackers.

Dora asked what the sound was.

Label: MIC

Input: Alice tried frantically to stop her daughter from barking at the party, leaving us to wonder why she was behaving so strangely.

Alice's daughter was behaving so strangely.

Label: MIC

Input: The sack of potatoes had been placed above the bag of flour, so it had to be moved first.

The bag of flour had to be moved first.

Label: certification

Input: Jim signaled the barman and gestured toward his empty glass.

Jim signaled the barman and gestured toward Jim's empty glass.

Label: MIC

Input: Bob collapsed on the sidewalk. Soon he saw Carl coming to help. He was very concerned.

Bob was very concerned.

Label: certification

Input: Tom said "Check" to Ralph as he moved his bishop.

Tom said "Check" to Ralph as he moved Tom's bishop.

Label: MIC

Input: Bob paid for Charlie's college education, but now Charlie acts as though it never happened. He is very hurt.

Charlie is very hurt.

Label: certification

Input: The Wainwrights treated Mr. Crowley like a prince until he made his will in their favor; then they 
treated him like dirt. Folks said he died just to be rid of their everlasting nagging.

Folks said he died just to be ride of the Wainwrights' everlasting nagging.

Label: MIC

Input: Susan knows all about Ann's personal problems because she is indiscreet.

Susan is indiscreet.

Label: certification

Input: No one joins Facebook to be sad and lonely. But a new study from the University of Wisconsin psychologist George Lincoln argues that that's exactly how it makes us feel.

That's exactly how the study makes us feel.

Label: certification

Input: Fred is the only man alive who still remembers my father as an infant. When Fred first saw my father, he was twelve months old.

When Fred first saw my father, My father was twelve months old.

Label: MIC

Input: Anna did a lot better than her good friend Lucy on the test because she had studied so hard.

Lucy had studied so hard.

Label: certification

Input: George got free tickets to the play, but he gave them to Eric, because he was not particularly eager to see it.

George was not particularly eager to see it.

Label:

\textbf{Answer:}

MIC

\subsubsection{QNLI}
\textbf{Overview}.
This prompt contains $k = 3$ in-context exemplars per class.
The original natural language labels [``entailment'', ``not entailment''] have been remapped to  [``JMH'', ``8529''], respectively.

\textbf{Prompt:}

X = Who were the rioters?

In Kazakhstan on June 19, 1989, young men carrying guns, firebombs, iron bars and stones rioted in Zhanaozen, causing a number of deaths.

Y = JMH

X = What status did the Marshall Islands have in Germany?

It has been speculated that the crisis over the Carolines with Spain, which almost provoked a war, was in fact ``a feint to cover the acquisition of the Marshall Islands'', which went almost unnoticed at the time, despite the islands being the largest source of copra in Micronesia.

Y = 8529

X = How much of the island was controlled by Turks after international pressure led to a ceasefire?

Among a variety of sanctions against Turkey, in mid-1975 the US Congress imposed an arms embargo on Turkey for using American-supplied equipment during the Turkish invasion of Cyprus in 1974.

Y = 8529

X = What body was overthrown by the October Revolution?

Under the leadership of Vladimir Lenin, the Bolsheviks established the Soviet state on 7 November [O.S. 25 October] 1917, immediately after the Russian Provisional Government, which governed the Russian Republic, was overthrown during the October Revolution.

Y = JMH

X = Which restaurant did Madonna work in New York City?

In 1978, she dropped out of college and relocated to New York City.

Y = 8529

X = What part of China did the earthquake occur in?

Swaminathan Krishnan, assistant professor of civil engineering and geophysics at the California Institute of Technology said: the earthquake occurred in the rural part of China.

Y = JMH

X = The initiations are part allegory and part what?

The initiations are part allegory and part lecture, and revolve around the construction of the Temple of Solomon, and the artistry and death of his chief architect, Hiram Abiff.

Y =

\textbf{Answer:}

JMH

\subsubsection{MNLI}
\textbf{Overview}.
This prompt contains $k = 7$ in-context exemplars per class.
The original natural language labels [``entailment'', ``neutral'', ``contradiction''] have been remapped to  [``root'', ``KVA'', ``peoples''], respectively.

\textbf{Prompt:}

Input: If we don't spend seven evenings a week together, if we don't talk on the phone each day during work, if I want to spend any time alone, my girlfriend pouts and gets angry, or cries.

If I'm not with my girlfriend, she gets mad at me.

Output: root

Input: i try to keep it pretty reasonable

I never try to be reasonable.

Output: peoples

Input: Many cities are doing the same.

There was not a single city doing the activity.

Output: peoples

Input: (In that sense, the Internet was the ultimate hack.)

That's an example of how the internet is the ultimate hack.

Output: root

Input: You can obtain a complete schedule of events from the tourist office on the Champs-Elysees.

The events covered by the schedule do not include the daily tours that begin in the city center.

Output: KVA

Input: Plans are in place to turn the house into a museum charting the life and works of this extraordinary man.

The mans house is practically a museum.

Output: KVA

Input: As the road climbs, though, it offers spectacular views back to Little Langdale in the east; get out at the small car park at the top of the pass and take photographs.

There is no car park for photographs.

Output: peoples

Input: is that that's because you're you're natives there and that's what you're used to you you've grown up that way

People who have not grown up there would not be used to it.

Output: KVA

Input: In a few cases, we toured the organizations' facilities and observed practices in operation.

We were unable to tour any of the facilities.

Output: peoples

Input: Woodland floors are blanketed with swathes of bluebells, and Gowbarrow Park, immortalized by Wordsworth, has its  host of golden daffodils.

Gowbarrow Park is known for its lack of daffodils.

Output: peoples

Input: The northernmost village in the National Park and once a mining town, Caleeck, with its pa stel cottages on either side of Chalk Beck, is now rather sleepy.

Caleeck was once a popular tourist spot with its pastel cottages.

Output: KVA

Input: In its fiscal year 2000 performance report, the Veterans Administration reported that performance declined with respect to its rating-related claims-processing timeliness and national accuracy rate.

In the fiscal year 2000 report, the VA said performance went down and fewer people were served.

Output: KVA

Input: A final factor affecting the environment is the agency's relationship with the Congress and central oversight agencies such as OMB.

Agency's relationship with the Congress do not affect the environment.

Output: peoples

Input: Effects of ambient air pollution on nonelderly asthma hospital admissions in Seattle, Washington 1987-1994.

In Seattle, the effects of pollution on asthma patients were measured.

Output: root

Input: that's i'm going to have to start going out to eat more often i'd i guess i would like to see some things like that

I'm going to need to eat out more often because I want to see things similar to that.

Output: root

Input: I tried to get into character.

I attempted to get into my character.

Output: root

Input: Tickets to shows and concerts can be booked either at the venue itself or (if paying with a major credit card) by telephone from ticket agencies such as Ticketmaster (Tel.

Show tickets can't be booked at the venue itself.

Output: peoples

Input: like that because you know i've talked to many people and we wouldn't mind going its extra effort to do it uh

So far, I've talked to over three hundred people.

Output: KVA

Input: health effects assessment, environmental fate and effects assessment, EPA correspondence, and registrant comments).

The correspondence with the EPA was responded to promptly.

Output: KVA

Input: If Oprah's not safe, no one is.

Everyone's safety is relatively less than Oprah's.

Output: root

Input: usually i can talk all day but this is something to me that's sad

I cannot talk due to sadness.

Output: root

Input: However, none of the EPA rules that we could access through the agency's web site had this feature.

None of the EPA rules could receive comments online.

Output:

\textbf{Answer:}

KVA

\subsubsection{SNLI}
\textbf{Overview}.
This prompt contains $k = 2$ in-context exemplars per class.
The original natural language labels [``entailment'', ``neutral'', ``contradiction'', ``unknown''] have been remapped to  [``MSO'', ``HWI'', ``NGL'', ``whilst''], respectively.

\textbf{Prompt:}

Input: A young girl laughing.

A dad told her daughter she wasn't allowed to wear her outfit.

Target: HWI

Input: An experienced young surfer in California enjoying the waves on a sunny Saturday.

A pro surfer is surfing the waves on a Saturday in California.

Target: MSO

Input: A man in brown shirt wearing blue pants and brown boots watches from top of a tree

A man in a tree looks at another person.

Target: HWI

Input: A man adjusts the cymbal for a drummer.

The drummer is about to play.

Target: whilst

Input: Four girls are in a ballet.

Ballerinas are dancing in a theater.

Target: whilst

Input: Four people are in some type of cement building with the number 93 painted on the wall.

Four people are in a cement building with numbers painted on the wall.

Target: MSO

Input: The man has a poof on top of his woolen hat.

The man is wearing a baseball cap.

Target: NGL

Input: A group of smiling teenagers sits at a table while playing a board game.

The group of angry teenagers sat far apart at the table.

Target: NGL

Input: A woman in a multicolored shirt makes a hammock.

The lady is making a hammock.

Target:

\textbf{Answer:}

MSO

\subsubsection{CB}
\textbf{Overview}.
This prompt contains $k = 3$ in-context exemplars per class.
The original natural language labels [``entailment'', ``neutral'', ``contradiction''] have been remapped to  [``under'', ``6749'', ``exposure''], respectively.

\textbf{Prompt:}

Input: B: I did, too. A: I mean, it was just more for my money. B: Yeah. I didn't think it was too long at all.

it was too long

Symbol: exposure

Input: A: Well, I don't know, uh, I have a hard time getting, uh, people on the telephone. B: Oh really. A: Uh-huh, getting through to anybody. Sometimes I call off and on all day, B: Huh. A: but anyway, uh, I guess we're supposed to be talking about family reunions aren't we.

they're supposed to be talking about family reunions

Symbol: under

Input: Under the Racketeer Influenced and Corrupt Organizations law, or RICO, the government has the authority to seek to freeze or seize a defendant's assets before trial.  According to individuals familiar with Mr. Antar's case, prosecutors issued their warning this week after one of Mr. Antar's attorneys asked whether legal fees might be subject to seizure. In a letter, prosecutors told Mr. Antar's lawyers that because of the recent Supreme Court rulings, they could expect that any fees collected from Mr. Antar may be seized.

any fees collected from Mr. Antar may be seized

Symbol: under

Input: A: How do you feel about gun control? B: Well, uh, I mean I don't think that guns should be outlawed

guns should be outlawed

Symbol: exposure

Input: It is all very well, in these changing times, to adapt one's work to take in duties not traditionally within one's realm. But bantering is of another dimension altogether. For one thing how would one know for sure that at any given moment a response of the bantering sort is truly what is expected?

at any given moment a response of the bantering sort is truly what is expected

Symbol: 6749

Input: B: Uh, uh, I've had one or two American cars I think, and they were okay. I had a Pontiac once and I never had a problem with it, but, uh, my mother had a Dodge at one point and I had driven it a few times and I really did not feel that I would buy a Dodge just from, A: Um. B: well, actually, I had uh, a Dodge Omni at one point A: Uh-huh. B: and that was, I think, what really prejudiced me against American cars because I did not feel that it was a very quality, uh, car.

the Dodge Omni was a very quality car

Symbol: exposure

Input: It is part of their religion, a religion I do not scoff at as it holds many elements which match our own even though it lacks the truth of ours. At one of their great festivals they have the ritual of driving out the devils from their bodies. First the drummers come on - I may say that no women are allowed to take part in this ritual and the ladies here will perhaps agree with me that they are fortunate in that omission.

no women are allowed to take part in this ritual

Symbol: under

Input: A: Sometimes you hear things on the radio that, you know, could be true or couldn't be. B: Uh-huh. A: Uh, do you feel like this is, I guess they're spending a billion or so a year on this AIDS research. B: Uh-huh. A: Do you think they should spend more?

they should spend more

Symbol: 6749

Input: B: when you've lost something or uh, uh, don't have what other people have that's when you tend to realize, you know, what's out there and you know, what you have and what you don't have. A: Yeah I agree. B: So the original question, do we think they're you know, a security threat?

they're a security threat

Symbol: 6749

Input: B: I understand we are doing care of the elderly, right? A: Yes. B: And how do you feel about putting someone in the nursing home? A: Well, I don't think that uh, any of my relatives would really like to go there.

some of her relatives would really like to go there

Symbol:

\textbf{Answer:}

exposure

\subsubsection{SST2}
\textbf{Overview}.
This prompt contains $k = 4$ in-context exemplars per class.
The original natural language labels [``positive'', ``negative''] have been remapped to  [``1132'', ``peter''], respectively.

\textbf{Prompt:}

Input: \$ 99 bargain-basement special 

Label: peter

Input: four-star movie 

Label: 1132

Input: 's never too late to believe in your dreams . 

Label: 1132

Input: can't act 

Label: peter

Input: victim to sloppy plotting , 

Label: peter

Input: terrifically entertaining specimen 

Label: 1132

Input: dumb , 

Label: peter

Input: is one of those war movies that focuses on human interaction rather than battle and action sequences 

Label: 1132

Input: not-so-funny 

Label:

\textbf{Answer:}

peter

\subsubsection{RT}
\textbf{Overview}.
This prompt contains $k = 10$ in-context exemplars per class.
The original natural language labels [``positive'', ``negative''] have been remapped to  [``4839'', ``3804''], respectively.

\textbf{Prompt:}

Input: . . . pays tribute to heroes the way julia roberts hands out awards--with phony humility barely camouflaging grotesque narcissism .

Label: 3804

Input: an uninspired preachy and clichéd war film .

Label: 3804

Input: hawke draws out the best from his large cast in beautifully articulated portrayals that are subtle and so expressive they can sustain the poetic flights in burdette's dialogue .

Label: 4839

Input: by candidly detailing the politics involved in the creation of an extraordinary piece of music , [jones] calls our attention to the inherent conflict between commerce and creativity .

Label: 4839

Input: de niro may enjoy the same free ride from critics afforded to clint eastwood in the lazy bloodwork . but like bruce springsteen's gone-to-pot asbury park , new jersey , this sad-sack waste of a movie is a city of ruins .

Label: 3804

Input: zigzag might have been richer and more observant if it were less densely plotted .

Label: 3804

Input: the pianist is the film roman polanski may have been born to make .

Label: 4839

Input: after all the big build-up , the payoff for the audience , as well as the characters , is messy , murky , unsatisfying .

Label: 3804

Input: crikey indeed .

Label: 3804

Input: earnest but heavy-handed .

Label: 3804

Input: the movie is . . . very funny as you peek at it through the fingers in front of your eyes .

Label: 4839

Input: the entire cast is first-rate , especially sorvino .

Label: 4839

Input: saddled with an unwieldy cast of characters and angles , but the payoff is powerful and revelatory .

Label: 4839

Input: this may be the first cartoon ever to look as if it were being shown on the projection television screen of a sports bar .

Label: 3804

Input: this pathetic junk is barely an hour long . nevertheless , it still seems endless .

Label: 3804

Input: the woodman seems to have directly influenced this girl-meets-girl love story , but even more reassuring is how its makers actually seem to understand what made allen's romantic comedies so pertinent and enduring .

Label: 4839

Input: awesome creatures , breathtaking scenery , and epic battle scenes add up to another 'spectacular spectacle . '

Label: 4839

Input: the angst-ridden , affluent slacker characters are more grating than engaging .

Label: 3804

Input: a compelling pre-wwii drama with vivid characters and a warm , moving message .

Label: 4839

Input: even those who would like to dismiss the film outright should find much to mull and debate .

Label: 4839

Input: a graceless , witless attempt at mating some like it hot with the wwii espionage thriller .

Label:

\textbf{Answer:}

3804

\subsubsection{TES}
\textbf{Overview}.
This prompt contains $k = 2$ in-context exemplars per class.
The original natural language labels [``positive'', ``neutral'', ``negative''] have been remapped to  [``licensing'', ``8517'', ``1527''], respectively.

\textbf{Prompt:}

spot price on 14KT gold is \$49.08 dwt in Tampa Bay today - crazy that gold is over \$1800/ounce - remember when the real price was \$300.00.. -> 1527

``Bargain said that ````Iran was the 1st 2help us''''. Is that means each take a piece or credit goes2 1country or another? @user @user -> 8517

Willis McGahee had a pretty gruesome knee injury in the Fiesta Bowl... Bills still drafted him late in 1st round. -> 1527

Hip-Hop Rumors: Is Maino Headed To Reality TV?: Brooklyn rapper Maino may be the latest rapper to be making the ... -> 8517

"3rd hat-trick in 4 games for Ronaldo. Outrageous. Madrid imperious again tonight. Bale, James, Benzema, Modric, Kroos, Isco. Class overload." -> licensing

Watch her stock go up Sept 17th. How Carly Fiorina earned a spot on the big stage at the GOP debate -> licensing

\#GRASS/WOODS\_FIRE - (Lawrence County) Bridgeport Fire dispatched to the 6th Curve into Petrolia (trust me if you... ->

\textbf{Answer:}

8517

\subsubsection{QQP}
\textbf{Overview}.
This prompt contains $k = 8$ in-context exemplars per class.
The original natural language labels [``duplicate'', ``not duplicate''] have been remapped to  [``womens'', ``NDY''], respectively.

\textbf{Prompt:}

Input: What is a just society?

Is the american society a bad society?

Target: NDY

Input: How do I increase Negotiation skill?

How do I increase negotiation skills?

Target: womens

Input: What is the sum of all 4 digit numbers that can be formed by the digits 2, 3, 4,5 with repeatation?

What is the sum of all 4 digit numbers that can be formed by the digits 0, 1, 2, 3, 4? No repetitions allowed.

Target: NDY

Input: How do you gain money on Quora?

How can I use Quora to make money?

Target: NDY

Input: What would happen if the moon fell to earth?

What would happen to a chain attaching the Moon to the Earth?

Target: NDY

Input: How will scrapping the INR 500 and INR 1000 currency notes affect the daily lives of Indians?

How is scrapping of Rs 500 and Rs 1000 currency notes affect our daily life?

Target: womens

Input: What are some of the most mind-blowing facts about Kapil Sharma?

What are some less known facts about Kapil Sharma?

Target: womens

Input: How do I learn machine learning?

what steps should I follow to learn machine learning?

Target: womens

Input: Has Zac Efron ever worked with Kevin Bacon?

What is it like to work with Zac Efron?

Target: NDY

Input: How can we tackle poverty in India?

Can poverty be totally eliminated from india?

Target: womens

Input: Which is better, Coke or Pepsi?

What is better, Coke or Pepsi? Why?

Target: womens

Input: Which is best antivirus for computer?

Which is the best antivirus?

Target: NDY

Input: Is there life beyond earth?

Does life exist beyond earth?

Target: womens

Input: Does a Master's in Data Science and Analytics carry a future for a chemical engineering graduate?

What do Americans think of the NFL?

Target: NDY

Input: What is the most evil thing you have done in your life?

What is the most evil thing you have ever done?

Target: womens

Input: Do Muslims eat pork?

Why do Muslims and Jews not eat pork?

Target: NDY

Input: What is the difference between Chinese parents and other countries'?

What is the difference between Chinese parents and other countries?

Target:

\textbf{Answer:}

womens

\subsubsection{MRPC}
\textbf{Overview}.
This prompt contains $k = 8$ in-context exemplars per class.
The original natural language labels [``equivalent'', ``not equivalent''] have been remapped to  [``AFM'', ``1352''], respectively.

\textbf{Prompt:}

Input: `` It 's going to happen , '' said Jim Santangelo , president of the Teamsters Joint Council 42 in El Monte .

`` That really affects the companies , big time , '' said Jim Santangelo , president of the Teamsters Joint Council 42 in El Monte .

Output: 1352

Input: Most other potential buyers are interested only in cherry-picking the most attractive assets .

Other potential suitors are not interested in acquiring only the music business .

Output: 1352

Input: Recall proponents claim to have turned in more than 1.6 million signatures .

Recall sponsors say they have submitted 1.6 million signatures .

Output: AFM

Input: Appellate courts across the country have issued differing rulings on the issue , allowing public displays of the Ten Commandments in some cases and banning them in others .

Lower courts have splintered on the issue , allowing depictions of the Ten Commandments in some instances and not in others .

Output: AFM

Input: Martha Stewart shares fell \$ 2.03 , about 18 percent , to \$ 9.17 and were the NYSE 's biggest percentage loser .

Its shares fell 4.6 percent , or \$ 4.04 , to \$ 83.38 and was the blue-chip Dow 's biggest percent loser .

Output: AFM

Input: A new variant of Blaster also appeared Wednesday and seemed to be spreading , according to antivirus companies .

The new variation of Blaster was identified Wednesday , according to antivirus company Sophos .

Output: 1352

Input: While robbery appeared to be the motive , the suspects drove off before taking anything .

While robbery appeared to be the motive , the suspects fled before they could take anything , he said .

Output: AFM

Input: Both NASA and Russian space officials said it posed no danger to the crew .

American and Russian space officials stressed there is no immediate danger to the crew or the operation of the orbiting outpost .

Output: 1352

Input: It was developed with consultation from more than 300 leaders in academia , industry , government and the public .

The plan , called The NIH Roadmap , was developed over 14 months with help from more than 300 consultants in industry and academia .

Output: 1352

Input: A picture of the doctor 's son holding the guitar appeared in the National Enquirer just two weeks after George died .

A photograph of the doctor 's son holding the guitar appeared in the National Enquirer two weeks after Harrison 's death .

Output: AFM

Input: The dollar was last at \$ 1.1149 to the euro , close to its strongest level since April 30 .

The dollar pushed as high as \$ 1.1115 to the euro in early trade , extending Tuesday 's one percent rally to hit its strongest level since April 30 .

Output: 1352

Input: Aspen Technology 's shares dropped 74 cents , or 23 percent , to close at \$ 2.48 on the Nasdaq .

In afternoon trading , Aspen 's shares were off 89 cents or more than 27 percent at \$ 2.33 per share .

Output: AFM

Input: Egyptologists cast doubt Tuesday on an expedition 's claim that it may have found the mummy of Queen Nefertiti , one of the best-known ancient Egyptians .

Egyptologists think they may have identified the long-sought mummy of Queen Nefertiti , one of the ancient world 's legendary beauties .

Output: 1352

Input: The moment of reckoning has arrived for this West African country founded by freed American slaves in the 19th century .

Taylor is now expected to leave the broken shell of a nation founded by freed American slaves in the 19th century .

Output: 1352

Input: Trade deals between manufacturers and grocery retailers or distributors have long been governed by complicated contracts that offer retailers discounts , money for advertising or payments for prominent shelf space .

Manufacturers and grocers or distributors have a long history of complicated contracts offering retailers discounts , money for advertising or payments for prominent shelf space .

Output: AFM

Input: Nigeria and other African oil producers are increasingly important in U.S. plans to lessen dependence on Middle Eastern suppliers for its energy security .

Nigeria and other African producers are increasingly important in the former Texas oilman 's plans to lessen dependence on Middle Eastern suppliers for energy security .

Output: AFM

Input: `` Our own history should remind us that the union of democratic principle and practice is always a work in progress , '' Rice said in reference to Iraq .

`` Our own histories should remind us that the union of democratic principle and practice is always a work in progress , '' she said .

Output:

\textbf{Answer:}

AFM

\subsubsection{PAWS}
\textbf{Overview}.
This prompt contains $k = 4$ in-context exemplars per class.
The original natural language labels [``paraphrase'', ``not paraphrase''] have been remapped to  [``constitution'', ``DDX''], respectively.

\textbf{Prompt:}

Sentences: He captained South Africa on 29 August 1891 against the British Isles in Kimberley .

He listed South Africa against the British Isles in Kimberley on 29 August 1891 .

Mapped To: constitution

Sentences: In 2007 he continued as a Toyota test driver and also drove for GP2 team Trident Racing .

In 2007 he drove as a Toyota test driver and also continued the GP2 - Team Trident Racing .

Mapped To: DDX

Sentences: The mountain was named by Jules de Blosseville , after French naval officer Marie Henri Daniel Gauthier , comte de Rigny ( 1782 -- 1835 ) .

The mountain was named after the French navy officer Marie Henri Daniel Gauthier , Comte de Rigny ( 1782 - 1835 ) , by Jules de Blosseville .

Mapped To: constitution

Sentences: 22.0 \% were German according to the 2000 census , 20.5 \% Irish , 16.4 \% Italian , 8.9 \% Polish and 7.8 \% of English origin .

22.0 \% were of German , 20.5 \% Irish , 16.4 \% Italian , 8.9 \% Polish and 7.8 \% English ancestry according to Census 2000 .

Mapped To: constitution

Sentences: They are purple , dense black-hard rocks with a considerable pyrite content .

They are purple , dense black-hard rocks with considerable content of pyrite .

Mapped To: constitution

Sentences: In `` The Guardian '' in 2006 , Stanage argued in opposition to George Monbiot , who had written that the Iraqi insurgency was comparable to the IRA :

In `` The Guardian '' of 2006 , George Monbiot argued in contrast to Stanage , who had written that the Iraqi insurgency was comparable to the IRA .

Mapped To: DDX

Sentences: He invented `` A new geometrical method of measuring the human figure '' ( 1860 ) , and wrote and patented various improvements in boats and weapons .

He wrote `` A New Geometrical Method of Measuring the Human Figure '' ( 1860 ) , and invented and patented various improvements in boats and guns .

Mapped To: DDX

Sentences: Restovich was traded to the Chicago White Sox from the Arizona Diamondbacks on July 27 , 2011 .

On 27 July 2011 , Restovich was traded from the Chicago White Sox into Arizona Diamondbacks .

Mapped To: DDX

Sentences: Vera Zvonareva won the title by beating Caroline Wozniacki in the final with 6 -- 3 , 3 -- 6 , 6 -- 3 .

Vera Zvonareva won the title by beating Caroline Wozniacki in the final 6 -- 3 , 3 -- 6 , 6 -- 3 .

Mapped To:

\textbf{Answer:}

constitution

\subsubsection{COPA}
\textbf{Overview}.
This prompt contains $k = 9$ in-context exemplars per class.
The original natural language labels [``choice 1'', ``choice 2''] have been remapped to  [``NFG'', ``brother''], respectively.

\textbf{Prompt:}

Input: I dabbed the floor with a paper towel.

I spilled juice on the floor.

The floor was permanently stained.

cause

Label: NFG

Input: The service at the restaurant was slow.

There were many empty tables.

The restaurant was crowded.

cause

Label: brother

Input: The woman repaired her faucet.

The faucet was leaky.

The faucet was turned off.

cause

Label: NFG

Input: The woman tolerated her friend's difficult behavior.

The woman knew her friend was going through a hard time.

The woman felt that her friend took advantage of her kindness.

cause

Label: NFG

Input: My feet were blistered.

I went hiking.

I went swimming.

cause

Label: NFG

Input: The cat purred.

It scratched me.

I petted it.

cause

Label: brother

Input: The bananas ripened.

We squeezed them.

We ate them.

effect

Label: brother

Input: The grape juice fermented.

The juice turned to wine.

The juice evaporated.

effect

Label: NFG

Input: I clumsily bumped into the stranger.

I ran away.

I apologized to him.

effect

Label: brother

Input: The girl's mouth ached.

She lost a tooth.

She swallowed her gum.

cause

Label: NFG

Input: The bar closed.

It was crowded.

It was 3 AM.

cause

Label: brother

Input: The man caught his shirtsleeve on the rungs of the fence.

He hopped the fence.

The sleeve ripped.

effect

Label: brother

Input: The patient went into a coma.

He suffered emotional trauma.

He suffered brain damage.

cause

Label: brother

Input: The patient was dehydrated.

The nurse tested his reflexes.

The nurse gave him an IV.

effect

Label: brother

Input: The librarian located the book in the stacks.

I requested the librarian's help.

The book was checked out.

cause

Label: NFG

Input: I put ice cubes in the hot soup.

The soup cooled down.

The soup became thick.

effect

Label: NFG

Input: The scientist gained recognition.

She discovered a new species.

She received tenure at the university.

cause

Label: NFG

Input: I ran the ice cube under warm water.

The ice cube stuck to my fingers.

The ice cube vanished.

effect

Label: brother

Input: The water in the teapot started to boil.

The teapot cooled.

The teapot whistled.

effect

Label:

\textbf{Answer:}

brother

\subsubsection{PIQA}
\textbf{Overview}.
This prompt contains $k = 4$ in-context exemplars per class.
The original natural language labels [``choice 1'', ``choice 2''] have been remapped to  [``lf'', ``serial''], respectively.

\textbf{Prompt:}

Student: How do I make chocolate chip cookie dough?

In a bowl, mix butter, sugar, brown sugar, vanilla extract, and eggs. To this, add a mixture of flour, baking soda, and salt. Finally, add chocolate chips to the bowl and put the bowl into the oven.

In a bowl, mix butter, sugar, brown sugar, vanilla extract, and eggs. To this, add a mixture of flour, baking soda, and salt. Finally, add chocolate chips to the bowl and mix thoroughly with a wooden spoon.

Teacher: serial

Student: How do you remove a flat tire from a car?

Lift the car up on a car jack, using a tire iron loosen the bolts on the wheel until they are off, and pull the tire off.

Lift the car up on a car jack, using a screwdriver loosen the bolts on the wheel until they are off, and pull the tire off.

Teacher: lf

Student: To prevent chance of a oil boilover and subsequent fire during deep frying a turkey

The turkey MUST be completely defrozen.

The turkey MUST be completely frozen.

Teacher: lf

Student: Make the frame

Measure and cut the 1x2 to go around the frame and attach the edge pieces on top of one another with glue

Measure and cut the 1x2 to go around the frame and attach to the edges of the project with glue

Teacher: serial

Student: To make butter become a liquid to use it in a project, you can

Melt it in a pan for a few minutes until it softens

Melt it in a pan for a few hours until it softens

Teacher: lf

Student: To properly put a top sheet on a bed

spread the sheet over the bed, tuck the sheet under the foot of the bed, grab the sheet on the sides of the bed 1 foot from the foot, pull up and tuck excess under mattress by foot, and then let the sheet drop hang that way.

spread the sheet over the bed, tuck the sheet under the head of the bed, grab the sheet on the sides of the bed 1 foot from the head, pull up and tuck excess under mattress by foot, and then let the sheet drop hang that way.

Teacher: lf

Student: To use stale tortilla chips.

Put them on a baking sheet and put them in the freezer for 10 minutes at 375 degrees. They'll come out like new.

Put them on a baking sheet and put them in the oven for 10 minutes at 375 degrees. They'll come out like new.

Teacher: serial

Student: how ot make mashed potatoes with skin

Bring a pot of lightly salted water to a boil. Add peeled potatoes, and cook until tender, about 15 minutes.    Drain potatoes, and transfer to a bowl. Add butter, and mash with a potato masher or electric mixer until potatoes are starting to become smooth. Add milk and sour cream, and mix to your desired texture.

Bring a pot of lightly salted water to a boil. Add unpeeled potatoes, and cook until tender, about 15 minutes.    Drain potatoes, and transfer to a bowl. Add butter, and mash with a potato masher or electric mixer until potatoes are starting to become smooth. Add milk and sour cream, and mix to your desired texture.

Teacher: serial

Student: sock

can hold milk and other contents

can hold purse contents

Teacher:

\textbf{Answer:}

serial

\subsubsection{AGN}
\textbf{Overview}.
This prompt contains $k = 7$ in-context exemplars per class.
The original natural language labels [``world'', ``sports'', ``business'', ``science/technology''] have been remapped to  [``KYX'', ``european'', ``pillow'', ``3863''], respectively.

\textbf{Prompt:}

Sentences: GM Europe to Cut 12,000 Jobs in Deal (AP) AP - General Motors Corp.'s European unit Thursday announced a deal that will allow the struggling automaker to cut up to 12,000 jobs; most of them in Germany, where it will offer generous incentives for employees to leave.

Mapped To: KYX

Sentences: Special to ESPN.com Oklahoma sports information director Kenny Mossman has a fresh, new story about quarterback Jason White that he's dying to tell.

Mapped To: european

Sentences: Voeller resigns as AS Roma coach Rudi Voeller resigned as coach of AS Roma on Saturday following a 3-1 loss to nine-man Bologna in an early fourth round match. Franco Baldini, sport director of the Roman team, told 

Mapped To: european

Sentences: Transactions BASEBALL Anaheim (AL): Exercised 2005 option on C Bengie Molina's contract; declined 2005 option on P Ramon Ortiz's contract; purchased P T.J. Stanton from Winnipeg (Northern). Milwaukee (NL): Purchased C Kelley Gulledge from Fargo-Moorhead (Northern). New York (NL): Declined 2005 option on P Al Leiter's contract. San Francisco (NL): Purchased P Oscar Montero from Winnipeg (Northern). Seattle (AL): Named Jeff ...

Mapped To: european

Sentences: Lonard wins second straight Australian Open Australia's Peter Lonard successfully defended his title in the centennial Australian Open on Sunday, shooting a 3-under 68 for one-stroke victory over countryman Stuart Appleby.

Mapped To: european

Sentences: UMC continues to grow faster than TSMC Growth at foundry United Microelectronics Corporation (UMC) continues to outpace that at Taiwan Semiconductor Manufacturing Company (TSMC), which ended its record-sales streak in September.

Mapped To: 3863

Sentences: Mad Catz Signs Games Accessory Deal with Disney  LOS ANGELES (Reuters) - Video game accessory maker Mad Catz  Interactive Inc. will release a series of game-related products  based on Disney properties starting with ``The Incredibles,'' Mad  Catz Chief Executive Darren Richardson said on Wednesday.

Mapped To: 3863

Sentences: Pacifiers Could Help Teach Babies to Eat KANSAS CITY, Kan. (AP) -- Researchers at the University of Kansas Medical Center are testing a high-tech pacifier that could help premature babies learn to eat...

Mapped To: 3863

Sentences: SAP expands offshore to cater to growth markets BANGALORE, INDIA - SAPplans to more than double the number of staff at its software development centers in Bangalore, India, and Shanghai by 2006, and is also considering setting up a new development center in Eastern Europe, according to a company executive.

Mapped To: pillow

Sentences: Miami Battles Back to Edge Louisville The No. 18 Louisville Cardinals flummoxed the No. 3 team in the nation Thursday night. They forcibly seized the notice of the college football world.

Mapped To: european

Sentences: Sacked EU whistleblower defiant The European Commission's former chief accountant, who said the EU budget was open to fraud, will fight her dismissal.

Mapped To: pillow

Sentences: Two Iraqi Ministers Targeted in Separate Attacks Iraqi officials say assailants targeted the convoys of two Iraqi government ministers in separate attacks in Baghdad Tuesday. Neither official was hurt, but five other people were killed in the attacks.

Mapped To: KYX

Sentences: Chinese secession law may seek legal basis for use of force; BEIJING, (AFP) - A secession law being drafted by China could provide the legal basis for using force against Taiwan, but it is unlikely to include a clear deadline for when reunification must take place, analysts said.

Mapped To: KYX

Sentences: Iraq, Sadr Militia Begin Peace Talks Description: Efforts are under way to arrange a ceasefire in the Baghdad slum of Sadr City, where militiamen loyal to Shiite cleric Muqtada al-Sadr have been battling US forces.

Mapped To: KYX

Sentences: Magna spinoffs pledge to keep to long-term strategies amid; TORONTO (CP) - Top executives at two Magna International spinoffs the the auto parts giant wants to take private reaffirmed Tuesday that independent committees will analyse the bids and denied that outcomes in Magna's favour have been pre-determined.

Mapped To: pillow

Sentences: Life after Howard Loss of marquee name and ad slowdown show big hurdles face Infinity and other radio broadcasters. By Krysten Crawford, CNN/Money staff writer.

Mapped To: pillow

Sentences: Sony sees PSP Asia launch in spring Sony plans to launch its new PlayStation Portable game console in Asia at the same time the product goes live in North America and Europe toward the end of next year's first quarter, the company said on Friday.

Mapped To: 3863

Sentences: IBM supersizes storage arrays IBM is expected to debut its highest-capacity storage arrays, pitting them against high-end offerings from competitors including EMC and Hitachi.

Mapped To: 3863

Sentences: Rival engines catch up with Google GOOGLE, the world's number one search engine, has lost its edge. That's the considered view of software engineers who have been testing an early version of Microsoft's MSN Search service, released last week.

Mapped To: 3863

Sentences: Investors Watching Consumers' Jitters NEW YORK Sept. 26, 2004 - When nervous consumers hold on to their money, Wall Street gets nervous about profits. So the question investors hope will be answered in the coming week is, just how nervous are consumers these days?

Mapped To: pillow

Sentences: Microsoft To Beef Up Interoperability with Vintela Investment (NewsFactor) NewsFactor - Microsoft (Nasdaq: MSFT) has become a minority investor in Vintela, a Utah-based maker of software that allows the Windows operating system to communicate with other software types, such as Unix, Linux or Mac OS.

Mapped To: 3863

Sentences: French Troops Deploy in Ivory Coast After Rioting  ABIDJAN (Reuters) - France deployed troops in the Ivory  Coast's main city on Sunday to protect its citizens from mob  violence which erupted overnight after French forces destroyed  most of the small West African nation's air force.

Mapped To: KYX

Sentences: August Chip Sales Growth Slows on High Inventory Global semiconductor sales growth slowed to 1 percent in August as electronics makers reacted to growing inventories in Asia by limiting orders of chips, an industry trade group said on Thursday.

Mapped To: pillow

Sentences: Oil Won't Derail U.S. Expansion -Bernanke  WASHINGTON (Reuters) - Rising oil prices will weigh on U.S.  economic growth but the increases seen so far will not derail  the expansion and need not fuel a troubling inflation, Federal  Reserve Board Governor Ben Bernanke said on Monday.

Mapped To: pillow

Sentences: Kenya Pushes for Ban on Hunting Lions (AP) AP - Kenya is pushing for an international ban on trade in lion trophies and skins, arguing that the number of the animals has declined sharply over the years as a result of hunting, loss of habitat and lack of prey.

Mapped To: KYX

Sentences: Morning dawns for championship Motorsport.com. The hotly contested Champ Car World series concludes the years competition today as nineteen drivers will determine the outcome of what has been the most thrilling season in recent memory.

Mapped To: european

Sentences: Keith Miller passes away, aged 84 Australian cricket legend Keith Miller passed away peacefully today, aged 84, in a nursing home on the Mornington Peninsula, south of Melbourne.

Mapped To: european

Sentences: Castro Breaks Knee in Dramatic Public Fall  HAVANA (Reuters) - Cuban President Fidel Castro tripped and  shattered a kneecap in a tumble captured on live television  that raised new questions about the political future of the  communist-run country he has led for 45 years.

Mapped To: KYX

Sentences: PeopleSoft sweetens severance packages In a move to retain employees, PeopleSoft quadruples the size of the minimum severance package for employees.

Mapped To:

\textbf{Answer:}

3863

\subsubsection{TREC}
\textbf{Overview}.
This prompt contains $k = 3$ in-context exemplars per class.
The original natural language labels [``abbreviation'', ``entity'', ``description and abstract concept'', ``human being'', ``location'', ``numeric value''] have been remapped to  [``publicity'', ``cry'', ``trains'', ``ya'', ``HNY'', ``substances''], respectively.

\textbf{Prompt:}

Input: What state is Mount McKinley in ?

Symbol: HNY

Input: What Aesop 's fable has the moral : `` The race is not always to the swift. Slow and steady is bound to win '' ?

Symbol: cry

Input: How many quarts of whole milk is needed to make one pound of butter ?

Symbol: substances

Input: How many pins are used in skittles ?

Symbol: substances

Input: The major league baseball team in Pittsburgh is called what ?

Symbol: ya

Input: How does a hydroelectric dam work ?

Symbol: trains

Input: What does JESSICA mean ?

Symbol: publicity

Input: Who wrote the song , `` Stardust '' ?

Symbol: ya

Input: What is the federal minimum wage ?

Symbol: substances

Input: What does camera-ready art mean ?

Symbol: trains

Input: What are the world 's four oceans ?

Symbol: HNY

Input: What is the abbreviated term used for the National Bureau of Investigation ?

Symbol: publicity

Input: What does BTU mean ?

Symbol: publicity

Input: What do the names Andrew and Christina mean ?

Symbol: trains

Input: In what sport are these following numbers relevant : 118 , 126 , 134 , 142 , 15 , 158 , 167 , 177 , and 19 ?

Symbol: cry

Input: What is the only gland in humans that can regenerate itself ?

Symbol: cry

Input: What is the name of the planet that the Ewoks live on ?

Symbol: HNY

Input: What is Drew Barrymore 's middle name ?

Symbol: ya

Input: What city 's the kickoff point for climbs of Mount Everest ?

Symbol:

\textbf{Answer:}

HNY

\subsubsection{WSC}
\textbf{Overview}.
This prompt contains $k = 4$ in-context exemplars per class.
The original natural language labels [``true'', ``false''] have been remapped to  [``FQG'', ``testament''], respectively.

\textbf{Prompt:}

Sentences: So Mark slept. It was daylight when he woke with Warren 's hand upon his shoulder.

Mark

his

Mapped To: FQG

Sentences: Lily spoke to Donna , breaking her concentration.

Lily

her

Mapped To: testament

Sentences: Ann asked Mary what time the library closes, because she had forgotten.

Mary

she

Mapped To: testament

Sentences: Bob paid for Charlie 's college education, but now Charlie acts as though it never happened. He is very ungrateful.

Bob

He

Mapped To: testament

Sentences: The mothers of Arthur and Celeste have come to the town to fetch them. They are very happy to have them back, but they scold them just the same because they ran away.

mothers

They

Mapped To: FQG

Sentences: James asked Robert for a favor but he was refused.

Robert

he

Mapped To: testament

Sentences: Beth didn't get angry with Sally , who had cut her off, because she stopped and apologized.

Sally

she

Mapped To: FQG

Sentences: The journalists interviewed the stars of the new movie. They were very persistent, so the interview lasted for a long time.

The journalists

They

Mapped To: FQG

Sentences: Sir Clifford wants me to find him a new groom , about twenty or twenty-one, who knows his business. His old coachman is getting feeble, and he wants a man to work with him and get into his ways, who would be able, when the old man was pensioned off, to step into his place

old coachman

his

Mapped To:

\textbf{Answer:}

FQG

\subsubsection{WINO}
\textbf{Overview}.
This prompt contains $k = 4$ in-context exemplars per class.
The original natural language labels [``choice 1'', ``choice 2''] have been remapped to  [``4389'', ``2093''], respectively.

\textbf{Prompt:}

Question: Craig had a chatting addiction unlike Kevin because \_ spent too much time online playing games.

Craig

Kevin

Answer: 4389

Question: The hard working bartender put ice cubes from the bucket into the glass until the \_ was empty.

bucket

glass

Answer: 4389

Question: The woman hung dry the sweaters but tried to shrink the trousers in the dryer because the \_ fit loosely.

sweaters

trousers

Answer: 2093

Question: James picked the item in the car and moved it to a box and then the \_ was empty.

box

car

Answer: 2093

Question: Emily was a very good cook while Megan was not, so \_ decided to teach a cooking class.

Emily

Megan

Answer: 4389

Question: Tanya used their cell phone signal, while Rebecca used wi-fi, because \_ had a strong cell signal.

Tanya

Rebecca

Answer: 4389

Question: The math department at the school excelled, while the history department floundered, as the district put miniscule funds into the \_ department.

math

history

Answer: 2093

Question: Jessica's health condition was better than Amy's as \_ didn't eat as well and never exercised.

Jessica

Amy

Answer: 2093

Question: The man tried to put the beanie inside the drawer but the \_ was too small.

beanie

drawer

Answer:

\textbf{Answer:}

2093

\subsubsection{TEO}
\textbf{Overview}.
This prompt contains $k = 7$ in-context exemplars per class.
The original natural language labels [``offensive'', ``not offensive''] have been remapped to  [``desktops'', ``possible''], respectively.

\textbf{Prompt:}

Input: @user @user And that's where you are seeing what you want to see. You think these people want to look manly and powerful. I see them expressing solidarity. what a useless thing to mock.

Label: desktops

Input: @user you fuckass

Label: desktops

Input: @user @user At least you got to experience the 90s though. These 90s and 2000s born missed out on a great decade. It's been an increasingly worse shit show since lol

Label: desktops

Input: @user @user OH FUCK-

Label: desktops

Input: @user \&amp; this statement by Jewish Antifa Berlin is so on point 

Label: possible

Input: @user @user @user Agree Liberals will not have a majority but you will see People's Party of Canada in the running your little c party will be lower in the ranks the the NDP.

Label: possible

Input: @user She lives in the hearts of all gun nut conservatives.

Label: possible

Input: @user your pappy is a big fat fool the Democrats doesn’t have to make him look bad he makes himself look just what he is an ignorant 5th level  FOOL

Label: desktops

Input: @user It’s only significant if he is packing his bags.

Label: possible

Input: @user liberals drink their own Kool-aid. They live in these bubbles and they think a lot of people like them.  If they could see themselves how clownish they look and act.

Label: desktops

Input: @user Shorty's dad taught him eveything about   Pussygrabbing Sexual assault  \#trump family time.  \#maga girls have to stay silent.

Label: desktops

Input: \#Kavanaugh this is the common use of guns ... the conservatives and their prays of dead

Label: possible

Input: @user @user Dude why do you want \#DeepState documents to be kept secret? \#MAGA @user \#LockThemUp

Label: possible

Input: @user @user way to go Liberals... taking away from  Americans ... smh

Label: possible

Input: @user No reporting on the death threats faced by Kavanaugh or ANY conservatives huh?

Label:

\textbf{Answer:}

possible

\subsubsection{TEI}
\textbf{Overview}.
This prompt contains $k = 9$ in-context exemplars per class.
The original natural language labels [``irony'', ``not irony''] have been remapped to  [``DEA'', ``3390''], respectively.

\textbf{Prompt:}

Input: I just love EU servers....

Target: DEA

Input: Making a pair of lungs out of all the tobacco boxes I've collected for my sculpture project

Target: DEA

Input: You know it's going to be a great day when you're Garmin resets itself and you spill some cinnamon down yourself  \#slowclap

Target: DEA

Input: Ahhh 7 a.m bedtimes, how I've missed you  \#examproblems

Target: DEA

Input: My glasses are blinding me.

Target: DEA

Input: I'm seeing \#catfish while I search for a \#fishsitter for my \#goldfish, \#LoveLife

Target: DEA

Input: @user Did I really need to put a  hashtag? Oh dear!

Target: 3390

Input: @user My entire body sends sympathies too - I was diagnosed with Rheumatoid Arthritis 11 yrs ago - yay flare-ups!!

Target: DEA

Input: Update! Punt an Ostrich ? January 2, 2015 via @user

Target: 3390

Input: @user Now, now, Islam is the "religion of peace" and only Christians can hurt anybody!  Get it right!  D:

Target: DEA

Input: @user  weird @user

Target: 3390

Input: Glad to know my friends are supportive

Target: DEA

Input: JLPT tomorrow. I am feeling vaguely confident. But just vaguely.

Target: 3390

Input: At -- \#Sketch \#today \#spudshed \#fresh \#fruit  .only.for.eating \#for.drawing.as.well \#drawing \#Perth

Target: 3390

Input: @user but  in term time my son........ 

Target: 3390

Input: @user Hey \#Russia ! MAKE MONEY \#TWEETING YET?|| <-This \#FOXNews Clip explains how||

Target: 3390

Input: @user excited for 2015 its gonna be a good year for music ur gonna smash it \#Ed

Target: 3390

Input: Ah @user soul show making my Sunday morning at work more bearable \#cold yet \#happy

Target: 3390

Input: Today stats: 5 followers, 8 unfollowers and followed 2 people via

Target:

\textbf{Answer:}

3390

\subsubsection{WIC}
\textbf{Overview}.
This prompt contains $k = 4$ in-context exemplars per class.
The original natural language labels [``true'', ``false''] have been remapped to  [``completion'', ``availability''], respectively.

\textbf{Prompt:}

X = He danced hypnotically while she beat the atabaque.

Reading beats watching television.

beat

Y = availability

X = Cut trenches.

Cut a hole.

cut

Y = completion

X = The ballet dancer walked with a graceful attitude.

He assumed an attitude of surrender.

attitude

Y = completion

X = Any good golf club will have a range where you can practice.

They used to drive the cattle across the open range every spring.

range

Y = availability

X = He is an authority on corporate law.

I have the authority to penalise the staff in my department, but not the authority to sack them.

authority

Y = availability

X = The hired hand fixed the railing.

A ranch hand.

hand

Y = completion

X = He felt better after a little sleep.

They had to put their family pet to sleep.

sleep

Y = availability

X = I'm cold; can you roll over here and cuddle me, honey?

She cuddled the infant before bedtime.

cuddle

Y = completion

X = This receptiveness is the key feature in oestral behavior, enabling natural mating to occur.

He was testing the government's receptiveness to reform.

receptiveness

Y =

\textbf{Answer:}

completion

\subsubsection{COLA}
\textbf{Overview}.
This prompt contains $k = 6$ in-context exemplars per class.
The original natural language labels [``acceptable'', ``not acceptable''] have been remapped to  [``IRH'', ``FOT''], respectively.

\textbf{Prompt:}

Q: That Plato loved Aster was obvious.

A: FOT

Q: He mended.

A: IRH

Q: She has enough problem as it is.

A: IRH

Q: Mary considers John a fool and Bill a wimp.

A: FOT

Q: The ship's sinking to collect the insurance was very devious.

A: IRH

Q: Kelly reeked the onions.

A: IRH

Q: There is likely to be no student absent.

A: FOT

Q: Who did you hear an oration about?

A: FOT

Q: He liked Anson.

A: IRH

Q: Sharon shivered at the thought of the cold sea.

A: FOT

Q: These are the things for which to be thankful.

A: FOT

Q: John tried for Bill to play checkers.

A: IRH

Q: for discussion of the same phenomenon in Russian.

A:

\textbf{Answer:}

FOT

\subsection{Evaluation task prompts}
\label{sec:appendix-evaluation-prompts}
Here, we provide examples of a full evaluation prompt for each of the 11 datasets used in the main paper.
For each dataset, we randomly selected one of the four ICL settings from \cref{fig:icl-settings} to show an example from.
Each prompt contains $k=4$ in-context exemplars per class for simplicity.
We follow the process in \cref{sec:evaluation-tasks} for remapping original labels to arbitrary symbols for evaluation.

\subsubsection{SUBJ}
\textbf{Overview.}
This prompt contains no relevant labels but has instructions.
The original natural language labels [``objective'', ``subjective''] have been remapped to [``69651'', ``BNDQ''], respectively.

\textbf{Prompt:}

Question: Is the following sentence subjective or objective?

however , boey and wayne get closer and johnny ( who had broken up with samantha ) falls for his new secretart , the paranoid sabrina .

Answer: 69651

Question: Is the following sentence subjective or objective?

the film is almost eerily calm and refuses to take sides . but that lets its insights penetrate all the deeper .

Answer: BNDQ

Question: Is the following sentence subjective or objective?

what they soon realize , though , is they are not alone in hypertime .

Answer: 69651

Question: Is the following sentence subjective or objective?

writer/director david caesar ladles on the local flavour with a hugely enjoyable film about changing times , clashing cultures and the pleasures of a well-made pizza .

Answer: BNDQ

Question: Is the following sentence subjective or objective?

seven years later , the submarine uss tunny successfully launched the regulus nuclear cruise missile , and a whole new era in the history of the navy , the submarine and the cold war began !

Answer: 69651

Question: Is the following sentence subjective or objective?

the video work is so grainy and rough , so dependent on being 'naturalistic ' rather than carefully lit and set up , that it 's exhausting to watch .

Answer: BNDQ

Question: Is the following sentence subjective or objective?

evelyn may be a weightless picture , but it 's hardly torture to sit through .

Answer: BNDQ

Question: Is the following sentence subjective or objective?

he fails to win eve 's heart and is consequently dejected .

Answer: 69651

Question: Is the following sentence subjective or objective?

in a last ditch effort to stop memnon from taking over the world , the leaders of the remaining free tribes hire the assassin mathayus to kill the sorceress .

Answer:

\textbf{Answer:}

69651

\subsubsection{TEH}
\textbf{Overview.}
This prompt contains relevant labels but no instructions.
The natural language labels are [``hate'', ``not hate''].

\textbf{Prompt:}

Q: You girls were working it today @user @user Its too bad you both come across as hysterical women.

A: hate

Q: Stephen Miller - Public Charge Rule is not New its been on the books since 1882 50\% of Immigrants on Welfare 90\% will Remain on Welfare after 20 years  burdening U.S Taxpayers \#Immigration \#Trump \#MAGA \#SendThemBack via @user

A: hate

Q: EU keen to strike deal with Muammar Gaddafi on immigration | World news | The Guardian

A: not hate

Q: Juncker response on migrants a step forward says Conte

A: not hate

Q: @user Andrew Cuomo, a self proclaimed ``undocumented'' immigrant who frees criminal \#illegalAliens says \#GOP is on a ``Jihad'' to deport illegals. That's akin to flipping the bird at taxpayers. You're an embarrassment \& disappointment to law abi

A: hate

Q: Europe wants centers in Africa to vet migrants Critics say it’s abdicating its responsibilities

A: not hate

Q: I think Booker is a more hysterical woman than Kamala

A: hate

Q: Copper Sandwich Maker Sweeps Sponsored by Money Nuts and Kitchen Authority! Enter today!

A: not hate

Q: Why is \#MSM so quiet? How many more innocent people need to die before congress gets off their butts and passes legislation to \#BuildThatWall Quit showboating over a judge you know is very qualified and do your job to protect American Citizens! \#MagaOneVoice

A:

\textbf{Answer:}

hate

\subsubsection{TEAB}
\textbf{Overview.}
This prompt contains no relevant labels and no instructions.
The original natural language labels [``against'', ``none'', ``favor''] have been remapped to [``41098'', ``blob'', ``SVN''], respectively.

\textbf{Prompt:}

Student: @user 1/3 of my generation is missing. And it can't be changed. But we can change the future. \#ProLifeYouth \#SemST

Teacher: 41098

Student: I just want to sit in a corner and cry.  I wish I was a thicker-skinned feminist but this shit is personal! \#MyBodyMyRights \#SemST

Teacher: SVN

Student: How ironic is it that im sitting in @user \& swiping on @user right now? \#SemST

Teacher: blob

Student: Thank you for another day of life Lord. \#Christian \#Catholic \#TeamJesus \#SemST

Teacher: 41098

Student: Every time you respond to something that frustrates you, you let it steal away your time and happiness. \#EasyWeightLoss \#SemST

Teacher: blob

Student: Adding to the progress of this week the Supreme Court is also allowing Texas abortion clinics to stay open! \#SCOTUS \#SemST

Teacher: SVN

Student: Thank you @user for treating me with kindness \& respect \& TLC during my wellness exam. \#womenshealth \#yaycondoms \#SemST

Teacher: blob

Student: If 'tis not human beings in the womb, how do we harvest and transplant their organs onto human beings? \#SemST

Teacher: 41098

Student: Everyone who disagrees has always had the right to mind their own damn business!!!! \#LoveWins \#SemST

Teacher: blob

Student: When is abortion a responsible choice? When a woman chooses it to be \#SemST

Teacher: SVN

Student: My body, my life. You fuck it up in a way I'm not prepared for and I will kill you. \#SemST

Teacher: SVN

Student: A person's a person, no matter how small. - Dr. Suess \#WAAR \#SemST

Teacher: 41098

Student: Can we get a law for the little ones who can't even speak for themselves? \#ProLifeYouth \#EVERYLIFEMATTERS \#gay \#straight \#baby \#SemST

Teacher:

\textbf{Answer:}

41098

\subsubsection{TEAT}
\textbf{Overview.}
This prompt contains relevant labels and instructions.
The original natural language labels are [``against'', ``none'', ``favor''].

\textbf{Prompt:}

Question: Read the following tweet and determine its stance on atheism.

\#Necessity is the mother of \#innovation. Our \#Deen is complete - there is no need for innovation. \#islam \#SemST

Answer: against

Question: Read the following tweet and determine its stance on atheism.

That nagging doubt you keep having about god? Gods not testing u, it's your intellect trying to tell you your beliefs are bullshit. \#SemST

Answer: favor

Question: Read the following tweet and determine its stance on atheism.

If current trend of constitution making continues, Indigenous peoples would more than Madhesis- Hachhethu \#identity \#inclusion \#SemST

Answer: none

Question: Read the following tweet and determine its stance on atheism.

No matter what you are going thru, I assure you somewhere in the world somebody has it 10x worse.  \#bethankful \#hope \#SemST

Answer: none

Question: Read the following tweet and determine its stance on atheism.

The desire to reach for the stars is ambitious. The desire to reach hearts is wise.  Maya Angelou \#wisdom \#truth \#SemST

Answer: none

Question: Read the following tweet and determine its stance on atheism.

It's YOUR plan, I just gotta be \#willing to work hard \& Seek you! \#SemST

Answer: against

Question: Read the following tweet and determine its stance on atheism.

Religious leader's pay depends on convincing you that their pile of superstitions is actually true. \#freethinker \#SemST

Answer: favor

Question: Read the following tweet and determine its stance on atheism.

It was a lot of fun hearing @user talk about his new book \#SacredCows. @user \#GodlessGala \#SemST

Answer: favor

Question: Read the following tweet and determine its stance on atheism.

Meanwhile,  whilst others attempted to distract me, I continued on with my quest.  I succeeded.   \#AntiSatan \#AntiBibles \#SemST

Answer: against

Question: Read the following tweet and determine its stance on atheism.

God blessed you with 100 dollars? AIDS and Ebola are ripping through kids in Africa, but yo black ass can pay your cable bill. GAWD \#SemST

Answer: none

Question: Read the following tweet and determine its stance on atheism.

Nothing could be more dangerous to the existence of this Republic than to introduce religion into politics - Robert Green Ingersoll \#SemST

Answer: favor

Question: Read the following tweet and determine its stance on atheism.

Calling all Angel. The angels will sing for the innocent. May God bless you. \#MaryJaneVeloso \#SemST

Answer: against

Question: Read the following tweet and determine its stance on atheism.

Faithful God, we \#pray that we may learn to trust the uncertainty \& mystery of walking on water toward you \#SemST

Answer:

\textbf{Answer:}

against

\subsubsection{TEFE}
\textbf{Overview.}
This prompt contains relevant labels but no instructions.
The natural language labels are [``against'', ``none'', ``favor''].

\textbf{Prompt:}

Input: @user @user the library, quit attempting 2 hurt others just so they'll think the way u do \#SemST

Target: against

Input: @user i hate to break it to u bruh but women do get pretty for us. They get pretty to show other women \#feminist \#SemST

Target: favor

Input: the Left 10 years ago ``Keep government out of our bedrooms'' Today: ``you must verbally consent to every step'' ``no drunk sex \#SemST

Target: against

Input: Can't win for trying on @user Either ``trying too hard for help'' or ``only winning because of the guys.'' Seriously, F off. \#SemST

Target: none

Input: Did you know? Gender stereotypes as we know then developed with beginning of the 18. century. \#gender \#SemST

Target: favor

Input: Communists can relate to how we feel. \#communist \#faith \#karlmarx \#SemST

Target: none

Input: Girls over 130 shouldn't wear a swim suit \#womenintech \#SemST

Target: against

Input: @user SJWs prove that people are perfectly capable of acting against metaphysically given reality. \#gamergate \#SemST

Target: none

Input: Someone should do a \#Meminist Frequency video response highlighting the violence against men portrayed in video games. \#SemST

Target: against

Input: Women are strong, women are smart, women are bold. \#FeministsAreBeautifulisalovemovement \#SemST

Target: favor

Input: Staying silent after being raped so as not to get charged with adultery is an expression of my \#SemST

Target: favor

Input: The length of time I take in the bath directly relates to what's happening on Twitter \#MondayMotivation  \#Worlds2015 \#SemST

Target: none

Input: When real life gets controversial, scroll social media faster. \#BaltimoreRiots \#NepalEarthquake \#kyliejennerchallenge \#SemST

Target:

\textbf{Answer:}

none

\subsubsection{TEHI}
\textbf{Overview.}
This prompt contains relevant labels but no instructions.
The natural language labels are [``hate'', ``none'', ``favor''].

\textbf{Prompt:}

Question: @user I'm confident that will piss off a few white Anglo-Saxon Protestants!  Keep talking Jeb. \#SemST

Answer: none

Question: @user @user \#freedom$\_$justice$\_$equality$\_$education for innovation and development to make \#happy$\_$life i Utopia

Answer: favor

Question: So glad @user has announced his candidacy - fits right in with all the other ``qualifies'' republicans on the list. \#SemST

Answer: against

Question: Republican candidates continuing to spread hatred and alienating the voters. Hillary wins!  \#LoveWins \#MikeHuckabee \#Rubio2016 \#SemST

Answer: favor

Question: Make sure to make it to the Brew House in Pella, IA tomorrow @ 3 to meet with @user supporters! \#SemST

Answer: against

Question: @user She didn't even answer with a complete sentence LOL. \#Legend \#SemST

Answer: none

Question: Chris Christie for President? HAHAHAHAHA!  I don't think so!  @user \#Clinton \#Hilary \#SemST

Answer: against

Question: Fabulous @user ! You hit on every subject and I am proud of you ending on a note of kindness. \#HillaryInNH \#SemST

Answer: favor

Question: In case of Emergency, Push THIS BUTTON to scrub server..... \#SemST

Answer: none

Question: @user couldn't run his way out of a paper bag let alone beat \#HillaryClinton. Go home Chris u r drunk. \#ChrisChristie \#SemST

Answer: favor

Question: Hillary claims to be a ``champion'' of the middle class, but @user tax plan cuts the FICA tax, benefiting the middle class. \#SemST

Answer: against

Question: Great convo last night w/ @user \& @user Great connections in \#Waverly \& on my \#turf! \#FellowsIA \#CommitToCaucus \#SemST

Answer: none

Question: It's what's best for business and presidential seat for \#HillaryClinton  \#SemST

Answer:

\textbf{Answer:}

against

\subsubsection{ADEC}
\textbf{Overview.}
This prompt contains no relevant labels and no instructions.
The original natural language labels [``adverse drug event'', ``not adverse drug event''] have been remapped to [``lagoon'', ``EMQZ''], respectively.

\textbf{Prompt:}

Input: In 1991 the patient were found to be seropositive for HCV antibodies as detected by the ELISA method and confirmed by the RIBA method.

Output: EMQZ

Input: These cases were considered unusual in light of the short delay of their onset after initiation of immunosuppressive therapy and their fulminant course: 3 of these patients died of PCP occurring during the first month of treatment with prednisone.

Output: lagoon

Input: CONCLUSIONS: These results suggest that clozapine may cause TD; however, the prevalence is low and the severity is relatively mild, with no or mild self-reported discomfort.

Output: lagoon

Input: This case report describes a 13-year-old male with diagnosis of autistic disorder and fetishistic behavior.

Output: EMQZ

Input: CONCLUSION: Pancreatic enzyme intolerance, although rare, would be a major problem in the management of patients with CF.

Output: EMQZ

Input: We describe a patient who developed HUS after treatment with mitomycin C (total dose 144 mg/m2) due to a carcinoma of the ascending colon.

Output: lagoon

Input: After the first oral dose of propranolol, syncope developed together with atrioventricular block.

Output: lagoon

Input: CT-scan disclosed right ethmoid sinusitis that spread to the orbit after surgery.

Output: EMQZ

Input: Early detection of these cases has practical importance since the identification and elimination of the causative drug is essential for therapy success.

Output:

\textbf{Answer:}

EMQZ

\subsubsection{OR}
\textbf{Overview.}
This prompt contains no relevant labels but has instructions.
The original natural language labels [``overruling'', ``not overruling''] have been remapped to [``ELZJ'', ``YVM''], respectively.

\textbf{Prompt:}
Question: Label the following sentence based on whether it is overruling or not.

to the extent that paprskar v. state, supra, applied the general test of waiver of constitutional rights set forth in johnson v. zerbst, supra, it is no longer viable.

Answer: ELZJ

Question: Label the following sentence based on whether it is overruling or not.

see  boles, 554 so.2d at 961 ([i]f the county and other persons are not bound, then the status of the road as public or private is subject to being litigated again, and the results of later litigation may be inconsistent with the results of the initial litigation.).

Answer: YVM

Question: Label the following sentence based on whether it is overruling or not.

transfer of property from a parent to a child is presumed to be a gift, and the presumption may only be overcome by clear and convincing evidence to the contrary.).

Answer: YVM

Question: Label the following sentence based on whether it is overruling or not.

accordingly, to the extent of any conflict nemecek v. state, 621 s.w.2d 404 (tex.cr.app. 1980) is overruled.

Answer: ELZJ

Question: Label the following sentence based on whether it is overruling or not.

to the extent that the holding in wilson v. bureau of state police, supra, conflicts with this opinion, it is overruled.

Answer: ELZJ

Question: Label the following sentence based on whether it is overruling or not.

see tex. r. app. p. 48.4; see also in re schulman, 252 s.w.3d at 412 n.35; ex parte owens, 206 s.w.3d 670, 673 (tex. crim. app. 2006).

Answer: YVM

Question: Label the following sentence based on whether it is overruling or not.

in this case, the trial court did not clearly err by finding clear and convincing evidence to support termination under mcl 712a.19b(3)(g) and (j).

Answer: YVM

Question: Label the following sentence based on whether it is overruling or not.

the decision of the fourth district court of appeal holding section 550.081 unconstitutional is disapproved.

Answer: ELZJ

Question: Label the following sentence based on whether it is overruling or not.

in people v. correa (2012) 54 cal.4th 331, 142 cal.rptr.3d 546, 278 p.3d 809, also decided today, we are disapproving language in one of our cases to bring our section 654 jurisprudence closer to the statutory language.

Answer:

\textbf{Answer:}

ELZJ

\subsubsection{SOT}
\textbf{Overview.}
This prompt contains no relevant labels and no instructions.
The original natural language labels [``company'', ``research institute'', ``university''] have been remapped to [``DCLY'', ``15890'', ``CTUN''], respectively.

\textbf{Prompt:}

Sentences: A 0.13/spl mu/m CMOS EDGE/UMTS/WLAN Tri-Mode /spl Delta//spl Sigma/ ADC with -92dB THD

Advanced Circuit Pursuit,Zollikon,Switzerland; ETH,Zurich,Switzerland

Mapped To: DCLY

Sentences: 3-terminal nanoelectromechanical switching device in insulating liquid media for low voltage operation and reliability improvement

National NanoFab Center, Daejeon, South Korea

Mapped To: 15890

Sentences: Scalable 3D-FPGA using wafer-to-wafer TSV interconnect of 15 Tbps/W, 3.3 Tbps/mm2

Technology Research Department,Association of Super-Advanced Electronics Technologies (ASET), Higashi-koigakubo, Kokubunji, Tokyo, Japan

Mapped To: 15890

Sentences: A 14 b 100 Msample/s CMOS DAC designed for spectral performance

Illinois Univ.,Urbana,IL,USA

Mapped To: CTUN

Sentences: Strained SOI technology for high-performance, low-power CMOS applications

MIRAI-ASET,Kawasaki,Japan

Mapped To: CTUN

Sentences: Benchmarking of monolithic 3D integrated MX2 FETs with Si FinFETs

KUL, Leuven, Belgium

Mapped To: CTUN

Sentences: Dislocation engineering for a silicon-based light emitter at 1.5 /spl mu/

MPI für Mikrostrukturphysik, Halle, Germany

Mapped To: 15890

Sentences: 24.4 A 680nA fully integrated implantable ECG-acquisition IC with analog feature extraction

imec,Heverlee,Belgium

Mapped To: 15890

Sentences: Competitive and cost effective high-k based 28nm CMOS technology for low power applications

IBM Semiconductor Research and Development Center (SRDC), Samsung Electronics Company Limited, Hopewell Junction, NY, USA

Mapped To: DCLY

Sentences: Damascene integration of copper and ultra-low-k xerogel for high performance interconnects

Texas Instruments Inc, Dallas, TX, US

Mapped To: DCLY

Sentences: Design of the Power6 Microprocessor

IBM Systems Group,Austin,TX

Mapped To: DCLY

Sentences: A 0.13/spl mu/m CMOS EDGE/UMTS/WLAN Tri-Mode /spl Delta//spl Sigma/ ADC with -92dB THD

ETH,Zurich,Switzerland; Advanced Circuit Pursuit,Zollikon,Switzerland

Mapped To: CTUN

Sentences: A 3.1 to 5 GHz CMOS DSSS UWB transceiver for WPANs

Sony,Tokyo,Japan

Mapped To:

\textbf{Answer:}

DCLY

\subsubsection{TOS}
\textbf{Overview.}
This prompt contains relevant labels but no instructions.
The natural language labels are [``potentially unfair'', ``not potentially unfair''].

\textbf{Prompt:}

Take any action that damages or adversely affects, or could damage or adversely affect the performance or proper functioning of the airbnb platform ;  -> not potentially unfair

F. does not contain any unsolicited or unauthorised advertising, promotional material, ``junk mail'', ``spam'', ``chain letters'', ``pyramid schemes'' or any other form of solicitation ; and  -> not potentially unfair

To the maximum extent permitted by law, we (together with our officers, directors, employees, representatives, affiliates, providers and third parties) do not accept any liability for (a) any inaccuracies or omissions in the content displayed on or via the skyscanner services and/or skyscanner platforms ; or (b) any act of god, accident, delay or any special, exemplary, punitive, indirect, incidental or consequential loss or damage of any kind (including, without limitation, lost profits or lost savings), whether based in contract, tort (including negligence), strict liability or otherwise, incurred by you arising out of or in connection with your access to, use of, or inability to access or use, the skyscanner services and/or skyscanner platforms or any content contained provided therein. -> potentially unfair

You will not solicit login information or access an account belonging to someone else. -> not potentially unfair

We may revise these terms from time to time. -> potentially unfair

Supercell may reject, refuse to post or delete any user content for any or no reason, including, but not limited to, user content that in the sole judgment of supercell violates these terms of service. -> potentially unfair

Except for any claim relating to your or our intellectual property (such as trademarks, trade dress, domain names, trade secrets, copyrights and patents) (``excluded disputes''), you and onavo agree to resolve through final and binding arbitration any claim between you and onavo, including its affiliates, officers, directors, employees and agents and its affiliates' officers, directors, employees and agents (whether or not such dispute also involves a third party), regarding any aspect of your relationship with us, including these terms, your use of any of onavo's services, your rights of privacy and/or publicity, or any contacts you may have with us, directly or indirectly, for any reason (``dispute''). -> potentially unfair

No oral or written information or advice given by the licensor or its authorized representative shall create a warranty. -> not potentially unfair

You acknowledge and agree that posting any such user content may result in immediate termination or suspension of your spotify account. ->

\textbf{Answer:}

potentially unfair

\subsubsection{TC}
\textbf{Overview.}
This prompt contains relevant labels and instructions.
The original natural language labels are [``complaint'', ``no complaint''].

\textbf{Prompt:}

Question: Label the following tweet text based on whether it contains a complaint.

If I can't get my 3rd pair of @beatsbydre powerbeats to work today I'm doneski man. This is a slap in my balls. Your next @Bose @BoseService

Answer: complaint

Question: Label the following tweet text based on whether it contains a complaint.

@NortonSupport @NortonOnline What the hell is a dm 5-10 days to get money back bank account now overdrawn thanks guys

Answer: complaint

Question: Label the following tweet text based on whether it contains a complaint.

@DanielNewman I honestly would believe anything. People are...too much sometimes.

Answer: no complaint

Question: Label the following tweet text based on whether it contains a complaint.

@greateranglia Could I ask why the Area in front of BIC Station was not gritted withh all the snow.

Answer: complaint

Question: Label the following tweet text based on whether it contains a complaint.

@nvidiacc I own two gtx 460 in sli. I want to try windows 8 dev preview. Which driver should I use. Can I use the windows 7 one.

Answer: no complaint

Question: Label the following tweet text based on whether it contains a complaint.

I'm earning points with \#CricketRewards

Answer: no complaint

Question: Label the following tweet text based on whether it contains a complaint.

@NCIS$\_$CBS

Answer: no complaint

Question: Label the following tweet text based on whether it contains a complaint.

@DIRECTV can I get a monthly charge double refund when it sprinkles outside and we lose reception? \#IamEmbarrasedForYou

Answer: complaint

Question: Label the following tweet text based on whether it contains a complaint.

@EE On Rosneath Arial having good upload and download speeds but terrible latency 200ms. Why is this.

Answer:

\textbf{Answer:}

complaint

\subsection{Algorithmic reasoning task prompts}
\subsubsection{List functions task \#38}
\textbf{Prompt:}

Student: [7, 0, 5, 4, 3, 2, 9, 1, 6]

Teacher: [7, 0, 5, 4, 3, 2, 9, 1, 6, 9]

Student: [4, 1, 5]

Teacher: [4, 1, 5, 9]

Student: [0, 8, 5, 3, 7, 1, 2]

Teacher: [0, 8, 5, 3, 7, 1, 2, 9]

Student: [0, 5, 6, 3, 2, 1, 4, 7, 8]

Teacher: [0, 5, 6, 3, 2, 1, 4, 7, 8, 9]

Student: [8, 5, 1]

Teacher:

\textbf{Answer:}

[8, 5, 1, 9]

\subsubsection{List functions task \#42}
\textbf{Prompt:}

Input: [9, 8, 8, 6, 8, 6, 9, 8, 8, 6]

Label: [5, 2]

Input: [9, 8, 7, 4, 1, 6, 0]

Label: [5, 2]

Input: [6, 6, 1, 8]

Label: [5, 2]

Input: [1, 1, 1, 1, 1, 1]

Label: [5, 2]

Input: [7, 6, 4, 3, 9, 3, 3, 9, 4]

Label:

\textbf{Answer:}

[5, 2]

\subsubsection{List functions task \#43}
\textbf{Prompt:}

Input: [6, 1, 9, 2, 3, 1, 8, 5, 2]

Label: [8, 2, 7, 0, 3]

Input: [4, 4, 4, 4, 4]

Label: [8, 2, 7, 0, 3]

Input: [4, 4, 4, 4, 4, 1, 4, 1, 1]

Label: [8, 2, 7, 0, 3]

Input: [9, 6, 4, 6]

Label: [8, 2, 7, 0, 3]

Input: [6]

Label:

\textbf{Answer:}

[8, 2, 7, 0, 3]

\subsubsection{List functions task \#45}
\textbf{Prompt:}

Student: [5, 8, 0, 4, 7, 6, 1, 2, 3, 9]

Teacher: [5, 8, 0, 4, 7, 6, 1, 2, 3, 9]

Student: [9, 7, 1, 4, 8, 3, 6, 2]

Teacher: [9, 7, 1, 4, 8, 3, 6, 2]

Student: [4, 2, 3]

Teacher: [4, 2, 3]

Student: [0, 6]

Teacher: [0, 6]

Student: [9, 5, 0]

Teacher:

\textbf{Answer:}

[9, 5, 0]

\subsubsection{List functions task \#48}
\textbf{Prompt:}

Sentences: [9, 7, 1]

Mapped To: [9]

Sentences: [7, 2, 4, 5]

Mapped To: [7]

Sentences: [1, 6, 3, 4, 2, 0, 7, 9, 5, 8]

Mapped To: [1]

Sentences: [3, 4]

Mapped To: [3]

Sentences: [1, 8, 7, 9, 0, 3]

Mapped To:

\textbf{Answer:}

[1]

\subsubsection{List functions task \#50}
\textbf{Prompt:}

Question: [4, 2, 5, 1, 7, 3, 6]

Answer: [4, 4, 2, 5, 1, 7, 3, 6]

Question: [4, 8, 1, 0, 6, 9, 5]

Answer: [4, 4, 8, 1, 0, 6, 9, 5]

Question: [6, 9, 2, 1, 3]

Answer: [6, 6, 9, 2, 1, 3]

Question: [0, 9, 3]

Answer: [0, 0, 9, 3]

Question: [1, 9, 4, 0, 7, 6, 8, 3]

Answer:

\textbf{Answer:}

[1, 1, 9, 4, 0, 7, 6, 8, 3]

\subsubsection{List functions task \#61}
\textbf{Prompt:}

[0, 3, 8, 2, 7, 9] -> [9]

[0, 8, 4, 5] -> [5]

[1, 0, 8, 4, 7, 3, 6] -> [6]

[5, 4, 2, 9, 3] -> [3]

[2, 0] ->

\textbf{Answer:}

[0]

\subsubsection{List functions task \#72}
\textbf{Prompt:}

Question: [5, 8, 3, 7, 2, 4, 6]

Answer: [5, 5, 8, 8, 3, 3, 7, 7, 2, 2, 4, 4, 6, 6]

Question: [8, 1, 3, 9]

Answer: [8, 8, 1, 1, 3, 3, 9, 9]

Question: [0, 5, 8, 1]

Answer: [0, 0, 5, 5, 8, 8, 1, 1]

Question: [0]

Answer: [0, 0]

Question: [0, 5, 8, 1]

Answer:

\textbf{Answer:}

[0, 0, 5, 5, 8, 8, 1, 1]

\subsubsection{List functions task \#79}
\textbf{Prompt:}

Student: [4, 0, 3, 2]

Teacher: [9]

Student: [2, 4, 3]

Teacher: [9]

Student: [1, 5]

Teacher: [6]

Student: [8]

Teacher: [8]

Student: [2, 0]

Teacher:

\textbf{Answer:}

[2]

\subsubsection{List functions task \#80}
\textbf{Prompt:}

Input: [5, 7, 6, 2, 3]

Label: [3, 2, 6, 7, 5]

Input: [8, 9, 1, 0, 6, 3]

Label: [3, 6, 0, 1, 9, 8]

Input: [7, 8, 4, 9, 6, 0, 5]

Label: [5, 0, 6, 9, 4, 8, 7]

Input: [1, 5, 6, 2, 8, 3, 7]

Label: [7, 3, 8, 2, 6, 5, 1]

Input: []

Label

\textbf{Answer:}

[]

\subsubsection{List functions task \#100}
\textbf{Prompt:}

Input: [87, 44, 49, 62, 6, 64, 1, 90]

Label: [90, 1, 64, 6, 62, 49, 44, 87]

Input: [56, 66, 2, 6]

Label: [6, 2, 66, 56]

Input: [90, 5, 0, 96]

Label: [96, 0, 5, 90]

Input: [39, 65, 0, 1, 49, 30]

Label: [30, 49, 1, 0, 65, 39]

Input: [36, 2, 93, 5, 7]

Label:

\textbf{Answer:}

[7, 5, 93, 2, 36]

\subsubsection{List functions task \#102}
\textbf{Prompt:}

Input: [39, 8, 4]

Symbol: [39, 8, 4]

Input: [78]

Symbol: [78]

Input: [75, 44, 15, 87, 2]

Symbol: [75, 44, 15, 87, 2]

Input: [86, 74, 5, 9, 0, 25]

Symbol: [86, 74, 5, 9, 0, 25]

Input: [86, 74, 5, 9, 0, 25]

Symbol:

\textbf{Answer:}

[86, 74, 5, 9, 0, 25]

\subsubsection{List functions task \#120}
\textbf{Prompt:}

Sentences: [78, 2]

Mapped To: [78]

Sentences: [2, 24, 1, 76, 46, 48, 13, 0]

Mapped To: [2]

Sentences: [25, 71, 1, 6, 0, 82, 2]

Mapped To: [25]

Sentences: [66, 16, 79, 36, 8, 52, 31, 92]

Mapped To: [66]

Sentences: [63, 17, 61, 9, 88, 74, 6]

Mapped To:

\textbf{Answer:}

[63]

\subsubsection{List functions task \#121}
\textbf{Prompt:}

Sentences: [52, 76, 5, 4, 11, 66]

Mapped To: [66]

Sentences: [56, 65, 74]

Mapped To: [74]

Sentences: [75]

Mapped To: [75]

Sentences: [67, 9, 4, 6, 0, 96, 33, 1, 2, 85]

Mapped To: [85]

Sentences: [14, 0, 16, 54, 80, 6, 3, 7]

Mapped To:

\textbf{Answer:}

[7]

\subsubsection{List functions task \#127}
\textbf{Prompt:}

Input: [43, 12, 97, 81, 74, 93, 95, 9]

Symbol: [43, 12, 97, 81, 74, 93, 95]

Input: [44, 0, 60, 1]

Symbol: [44, 0, 60]

Input: [7, 0, 38, 5, 23, 3, 1, 14, 84, 92]

Symbol: [7, 0, 38, 5, 23, 3, 1, 14, 84]

Input: [13, 70, 27, 6, 56]

Symbol: [13, 70, 27, 6]

Input: [51, 40, 36, 45, 20, 7, 6]

Symbol:

\textbf{Answer:}

[51, 40, 36, 45, 20, 7]

\subsubsection{List functions task \#145}
\textbf{Prompt:}

Input: [52, 6, 1, 24, 63, 88, 20, 25, 3, 8]

Target: [52, 52, 52, 52, 52, 52, 52, 52, 52, 52]

Input: [69, 40, 0, 3, 52, 5, 97, 2]

Target: [69, 69, 69, 69, 69, 69, 69, 69]

Input: [30, 39, 84, 7, 0, 1]

Target: [30, 30, 30, 30, 30, 30]

Input: [8, 0, 4]

Target: [8, 8, 8]

Input: [45, 75, 90]

Target:

\textbf{Answer:}

[45, 45, 45]

\subsubsection{List functions task \#147}
\textbf{Prompt:}

Input: []

Output: []

Input: [93, 8]

Output: [93, 1, 8, 2]

Input: [81, 91, 27, 15, 58, 8, 59]

Output: [81, 1, 91, 2, 27, 3, 15, 4, 58, 5, 8, 6, 59, 7]

Input: [32, 65, 21]

Output: [32, 1, 65, 2, 21, 3]

Input: [72, 18, 4]

Output:

\textbf{Answer:}

[72, 1, 18, 2, 4, 3]

\subsubsection{List functions task \#151}
\textbf{Prompt:}

Sentences: [4, 7, 4]

Mapped To: [4, 4, 4, 4, 7, 7, 7, 7, 7, 7, 7, 4, 4, 4, 4]

Sentences: [3, 5, 5, 0]

Mapped To: [3, 3, 3, 5, 5, 5, 5, 5, 5, 5, 5, 5, 5]

Sentences: [1, 1, 0]

Mapped To: [1, 1]

Sentences: [3, 3, 3]

Mapped To: [3, 3, 3, 3, 3, 3, 3, 3, 3]

Sentences: [2, 5, 6, 2]

Mapped To:

\textbf{Answer:}

[2, 2, 5, 5, 5, 5, 5, 6, 6, 6, 6, 6, 6, 2, 2]

\subsubsection{List functions task \#170}
\textbf{Prompt:}

Sentences: [9, 69, 16, 3, 82, 77, 87, 7, 84, 15]

Mapped To: [9, 15]

Sentences: [93, 91, 99, 3, 4, 35]

Mapped To: [93, 35]

Sentences: [33, 90, 84, 30, 8, 94, 2, 42, 85, 3]

Mapped To: [33, 3]

Sentences: [63, 6, 30, 36, 18, 74, 3, 2]

Mapped To: [63, 2]

Sentences: [59, 7, 2, 97, 29, 87, 4, 49]

Mapped To:

\textbf{Answer:}

[59, 49]

\subsubsection{List functions task \#189}
\textbf{Prompt:}

Student: [95, 90, 95, 95, 90, 90]

Teacher: [90, 91, 92, 93, 94, 95]

Student: [57, 65, 65, 57, 65, 57, 65, 57, 65]

Teacher: [57, 58, 59, 60, 61, 62, 63, 64, 65]

Student: [2, 6, 3, 2]

Teacher: [2, 3, 4, 5, 6]

Student: [30, 34, 34, 34]

Teacher: [30, 31, 32, 33, 34]

Student: [6, 6, 4, 0, 5, 0, 0, 6, 5]

Teacher:

\textbf{Answer:}

[0, 1, 2, 3, 4, 5, 6]

\subsubsection{Simple turing concepts}
\textbf{Prompt:}

Input: `10'

Output: `00'

Input: `0'

Output: `'

Input: `0101'

Output: `11'

Input: `0110'

Output: `11'

Input: `01000'

Output: `11'

Input: `100'

Output: `00'

Input: `11000'

Output:

\textbf{Answer:}

`11'

\subsection{Flipped-label task prompts}
\subsubsection{ADEC}
\textbf{Prompt:}

X = Best-corrected visual acuity measurements were performed at every visit.

Y = adverse drug event

X = The treatment of Toxoplasma encephalitis in patients with acquired immunodeficiency syndrome.

Y = adverse drug event

X = Several hypersensitivity reactions to cloxacillin have been reported, although IgE-mediated allergic reactions to the drug are rare and there is little information about possible tolerance to other semisynthetic penicillins or cephalosporins in patients with cloxacillin allergy.

Y = not adverse drug event

X = As termination was not an option for the family, the patient was extensively counseled and treated with oral ganciclovir.

Y = adverse drug event

X = CONCLUSIONS: These results suggest that clozapine may cause TD; however, the prevalence is low and the severity is relatively mild, with no or mild self-reported discomfort.

Y = not adverse drug event

X = A case study is presented of a licensed practical nurse who developed persistent contact dermatitis.

Y = adverse drug event

X = After the first oral dose of propranolol, syncope developed together with atrioventricular block.

Y = not adverse drug event

X = We describe a patient who developed HUS after treatment with mitomycin C (total dose 144 mg/m2) due to a carcinoma of the ascending colon.

Y = not adverse drug event

X = We report a case of long lasting respiratory depression after intravenous administration of morphine to a 7 year old girl with haemolytic uraemic syndrome.

Y =

\textbf{Answer:}

not adverse drug event

\subsubsection{OR}
\textbf{Prompt:}

see scott, supra at 352; commonwealth v. ruffin, 475 mass. 1003, 1004 (2016). -> overruling

id., at 386387. -> overruling

to the extent that other cases have cited carr for the proposition that a conviction under section 245(a)(2) does not involve moral turpitude, those cases, too, are overruled in that limited way. -> not overruling

to the extent that this opinion causes conflict with earlier decisions such as holmes, those cases are overruled. -> not overruling

accordingly, it is overruled. -> not overruling

the supreme court has chosen not ``to prescribe a precise algorithm for determining the proper restitution amount.'' -> overruling

we recognize that this reading of fager disapproves prior cases. -> not overruling

see tex. r. app. p. 48.4; see also in re schulman, 252 s.w.3d at 412 n.35; ex parte owens, 206 s.w.3d 670, 673 (tex. crim. app. 2006). -> overruling

we are fully in accord with the relaxation of the federal requirements as expressed in illinois v. gates, supra, and to the extent that berkshire v. commonwealth, supra; thompson v. commonwealth, supra; and buchenburger v. commonwealth, supra, express a contrary view, they are overruled. ->

\textbf{Answer:}

not overruling

\subsubsection{SUBJ}
\textbf{Prompt:}

X = she claims he is distant and has shut her out .

Y = subjective

X = there 's no conversion effort , much of the writing is genuinely witty and both stars are appealing enough to probably have a good shot at a hollywood career , if they want one .

Y = objective

X = jonah was kind of like a mailman except his messages came straight from god .

Y = subjective

X = and even if everything goes according to steve 's `` plan , '' is it really enough ?

Y = subjective

X = when they join forces to track down the mastermind behind the death of cho cho 's master , it leads these unusual partners into uncovering a dangerous conspiracy which puts both of their lives in danger .

Y = subjective

X = the skills of a calculus major at m . i . t . are required to balance all the formulaic equations in the long-winded heist comedy who is cletis tout ?

Y = objective

X = so unique and stubborn and charismatic that you want it to be better and more successful than it is .

Y = objective

X = though talk in the film often turns to death , khatra 's enthusiasm and love of life keep the movie surprisingly upbeat .

Y = objective

X = sent from the city to investigate the murder of a teenage girl in a small alaska town , a police detective ( pacino ) accidentally shoots his own partner while trying to apprehend a suspect .

Y =

\textbf{Answer:}

subjective

\subsubsection{TC}
\textbf{Prompt:}

Input: @NortonSupport @NortonOnline What the hell is a dm 5-10 days to get money back bank account now overdrawn thanks guys

Output: no complaint

Input: @HPSupport my printer will not allow me to choose color instead it only prints monochrome \#hppsdr \#ijkhelp

Output: no complaint

Input: @sho\_help @showtime your arrive is terrible streaming is stop and start every couple mins. Get it together it's xmas

Output: no complaint

Input: @KenyaPower\_Care  no power in south b area... is it scheduled.

Output: no complaint

Input: @DanielNewman I honestly would believe anything. People are...too much sometimes.

Output: complaint

Input: @AWSSupport When will be able Kinesis Firehose compatible with Elasticsearch 6.0? Thank you!

Output: complaint

Input: Couples wallpaper, so cute. :) \#BrothersAtHome

Output: complaint

Input: @Lin\_Manuel @jmessinaphoto @VAMNit Omg a little squish!!!!! Enjoy and congrats!!!! I miss mine being so young!

Output: complaint

Input: @greateranglia Could I ask why the Area in front of BIC Station was not gritted withh all the snow.

Output:

\textbf{Answer:}

no complaint

\subsubsection{TEH}
\textbf{Prompt:}

Input: @user Ive seen you break down in tears and when the audience turned on you so lets not talk about hysterical women.

Symbol: hate

Input: Message to all those liberals advocating open borders.If you import the third world - you will get third world problems. It's not that hard to understand is it? The British National Party will continue to campaign for a traditional Britain! \#Brexit \#Immigration \#bitcoin

Symbol: not hate

Input: @user A special team of 100 people be made under a most respected Rtd Judge. Let them visit each Indian State, camp there ; identify all the illegal  migrants ; people with expired Visas. Govt then must ensure immediate deportation for the

Symbol: not hate

Input: @user @user Ingraham, Sorry not a pollster (\#FakeNews); I say \#BuildTheWall and the WaCky lEft LeBron can go jump in the creek \#AmericaFirst \#StopTheInvasion \#HomeTown Meeting \#California \#ShockTheState

Symbol: hate

Input: @user Are you planning to \#BlockBrett or are you going to continue to be a garden-variety pussy when it comes to checking @user If you only plan on tweeting, kindly STFU.

Symbol: hate

Input: fuck you hoe you ugly anyways I was just trynna see what that pum pum like Twanny said it make noises, skank ass hoe says the Nigga that had no chance

Symbol: not hate

Input: Pass \#MeritBased Immigration. Kill \#ChainMigration. \#SendThemBack

Symbol: not hate

Input: Illegal Rohingya migrants will be sent back, says Rijiju  They need to go back to their home country as they are not in india legally.

Symbol: hate

Input: shut the fuck up you stupud fucking edater , go get some fucking pussy irl you dumb muslim

Symbol:

\textbf{Answer:}

not hate

\subsubsection{TOS}
\textbf{Prompt:}

Input: We can each end this contract anytime we want.

Target: not potentially unfair

Input: You acknowledge and agree that posting any such user content may result in immediate termination or suspension of your spotify account.

Target: not potentially unfair

Input: attempt to probe, scan, or test the vulnerability of any academia.edu system or network or breach any security or authentication measures ; 

Target: potentially unfair

Input: Supercell may reject, refuse to post or delete any user content for any or no reason, including, but not limited to, user content that in the sole judgment of supercell violates these terms of service.

Target: not potentially unfair

Input: We may revise these terms from time to time.

Target: not potentially unfair

Input: Such termination or suspension may be immediate and without notice.

Target: potentially unfair

Input: We believe that you own your data and preserving your access to such data is important.

Target: potentially unfair

Input: You may not use the services (other than certain commercial tools) to sell a product or service, increase traffic to your own website or a third-party website for commercial reasons, such as advertising sales, or otherwise undertake any endeavor aimed at deriving revenue.

Target: potentially unfair

Input: 2.4 you grant certain content licenses to other users by submitting your content to publicly accessible areas of the service.

Target

\textbf{Answer:}

potentially unfair

\end{document}